\documentclass{article}

\usepackage{microtype}
\usepackage{graphicx}
\usepackage{booktabs} 

\usepackage[pagebackref=true,breaklinks,colorlinks,citecolor=blue,linkcolor=blue,urlcolor=blue,hypertexnames=false]{hyperref}

\usepackage[accepted]{icml2025}

\usepackage{amsmath}
\usepackage{amssymb}
\usepackage{mathtools}
\usepackage{amsthm}
\usepackage{colortbl}%
\newcommand{\myrowcolour}{\rowcolor[gray]{0.925}}

\usepackage[capitalize,noabbrev]{cleveref}

\usepackage{paralist}
\usepackage{subcaption}  
\usepackage{caption} 
\usepackage{mathptmx} 
\usepackage{enumitem}
\usepackage{algorithm}
\usepackage{colortbl} 
\usepackage{xcolor} 
\usepackage[most]{tcolorbox}

\theoremstyle{plain}

\theoremstyle{definition}

\theoremstyle{remark}

\usepackage[textsize=tiny]{todonotes}

\icmltitlerunning{UMC: Unified Resilient Controller for Legged Robots with Joint Malfunctions}

\begin{document}

\twocolumn[
\icmltitle{UMC: A Unified Approach for Resilient Control of Legged Robots Across
Masked Malfunction Training}



\icmlsetsymbol{equal}{*}

\begin{icmlauthorlist}
\icmlauthor{Yu Qiu}{scu}
\icmlauthor{Xin Lin}{scu}
\icmlauthor{Jingbo Wang}{shai}
\icmlauthor{Xiangtai Li}{ntu}
\icmlauthor{Lu Qi}{whu,insta360}
\icmlauthor{Ming-Hsuan Yang}{ucm}
\end{icmlauthorlist}

\icmlaffiliation{scu}{Sichuan University}
\icmlaffiliation{shai}{Shanghai AI Lab,}
\icmlaffiliation{ntu}{Nanyang Technological University}
\icmlaffiliation{insta360}{Insta360 Research}
\icmlaffiliation{whu}{Wuhan University}
\icmlaffiliation{ucm}{University of California, Merced}

\icmlcorrespondingauthor{Lu Qi}{qqlu1992@gmail.com}




\icmlkeywords{Embodied AI, Deep Learning, Robotics, ICML}

\vskip 0.3in
]

\newcommand{\ql}[1]{\textcolor{cyan}{#1}}
\newcommand{\cm}[1]{\textcolor{red}{#1}}
\newcommand{\lx}[1]{\textcolor{green}{#1}}
\newcommand{\yq}[1]{\textcolor{blue}{#1}}



\printAffiliationsAndNotice{} 

\begin{abstract}

Adaptation to unpredictable damages is crucial for autonomous legged robots, yet existing methods based on multi-policy or meta-learning frameworks face challenges like limited generalization and complex maintenance. To address this issue, we first analyze and summarize eight different types of damage scenarios, including sensor failures and joint malfunctions. Then, we propose a novel, model-free, two-stage training framework, \textbf{U}nified \textbf{M}alfunction \textbf{C}ontroller (UMC), which incorporates a masking mechanism to enhance damage resilience. Specifically, the model is initially trained with normal environments to ensure robust performance under standard conditions. In the second stage, we use masks to prevent the legged robot from relying on malfunctioning limbs, enabling adaptive gait and movement adjustments upon malfunction. Experimental results demonstrate that our approach improves the task completion capability by an average of 36\% for the transformer and 39\% for the MLP across three locomotion tasks. The source code and trained models will be made available to the public.


\end{abstract}

\section{Introduction}
\label{sec:intro}

\begin{figure*}[ht]
  \centering
  \includegraphics[width=0.95\linewidth]{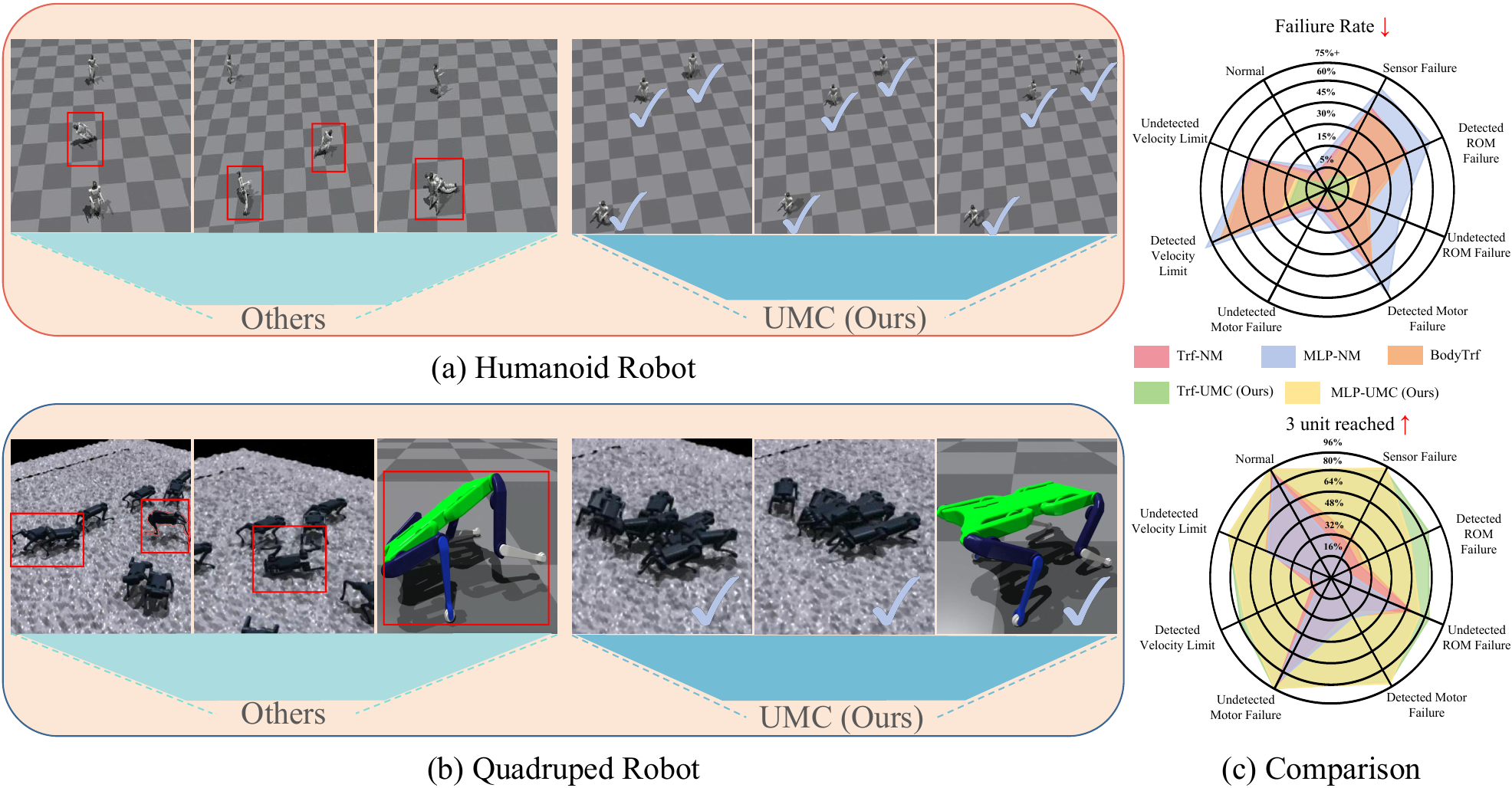}
  \caption{Qualitative and Quantitative Comparison of Our UMC Framework with Baselines and a SOTA Method. 
  `Trf' is the transformer, `NM' represents the baseline normal training (trained without damaged situations), and `UMC' is our method. `BodyTrf' is the abbreviation of BodyTransformer~\cite{sferrazza2024bodytransformerleveragingrobot}, also a baseline structure. `Failure Rate' refers to the proportion of robots that fall during their actions, and `3 Units Reached' refers to the percentage of robots that are still able to move a distance of 3 units after the added damage. (a) presents the qualitative comparison for the humanoid robot task, while (b) depicts it for the quadruped task. The last comparison set in (b) shows the qualitative comparison between our UMC framework and the SOTA method proposed in \cite{hou2024multitasklearningactivefaulttolerant}.}
  \label{fig:teaser}
  \vspace{-4mm}
\end{figure*}



Legged robots have witnessed significant advancement due to their flexibility and adaptability in various scenarios. 
%
Most current research focuses on the network design based on observational signals and self-states. 
However, they ignore robustness, particularly when their joints or limbs malfunction. 
This is a critical issue, as human intervention is impractical or even impossible in certain situations~\cite{Hutter2017ANYmalT, Bellicoso2018AdvancesIR, wensing2022optimizationbasedcontroldynamiclegged}. 
For example, in disaster recovery scenarios, a search-and-rescue robot navigating through rubble may experience joint failures caused by debris, making external assistance unsafe or impractical. 
Therefore, enhancing the performance of legged robots in such failures is essential for their effective deployment in real-world applications.



Some studies address the problem above by employing multi-policy and data augmentation strategies~\cite{kume2017mapbasedmultipolicyreinforcementlearning, YangGANARL, hou2024multitasklearningactivefaulttolerant} or leveraging meta-learning approaches~\cite{nagabandi2019learningadaptdynamicrealworld, raileanu2020fastadaptationpolicydynamicsvalue, guo2023decentralizedmotorskilllearning, 10598356} to enable robots to adapt to specific situations.
%
However, these methods raise concerns about their complex maintenance, generalization to various damaged conditions, and adaptability as well as performance in situations where conditions differ from the training environment. 
%
In Figure~\ref{fig:teaser}(b), we show that existing methods that are trained in specific damage settings exhibit many failures on both the humanoid and the quadruped robot in more open-world test settings. 
We formulate the problem of improving the robustness of diverse malfunctions in both data summarizing and pipeline designing. 
First, we systematically analyze multiple scenarios and categorize them into eight types, focusing on sensor-only failures, detectable joint damage, and undetectable joint damage. 
Each joint-damaged scenario includes tailored responses for specific impairments, such as restricted range of motion, reduced motor force, and limited linear velocity. 
Then, we propose a \textbf{U}nified \textbf{M}alfunction \textbf{C}ontroller (UMC), which is a simple but effective model-free approach employing a masking strategy to regularize the action network within two-stage training. 
%

Specifically, our UMC has a damage detection module and base structure that can generate joint-specific actions from original observation inputs representing each joint's position, velocity, and action information. 
The damage detection module receives the original observation input sequences and outputs two tensors, including a masked observation tensor that zeroes out the signals of the damaged joints from the original observation and a masking matrix among various joints.
The base structure includes a tokenizer, mask encoder, and detokenizer, where the tokenizer and detokenizer perform the transformation between the observation tensor and a sequence of action tokens.
%
%
%
The mask encoder is flexible to both transformer and MLP modules. It teaches the robot to act without damaged joints' information.
Based on the above architecture, UMC is first pre-trained on normal scenarios and then fine-tuned using collected damage scenarios, ensuring robust behaviour in both normal and damaged conditions.

Compared to baselines, our approach reduces the average fail rates by 30\% and 37\% with the base structure of the transformer and the MLP on three locomotion tasks, as shown in \cref{fig:teaser}(c). Furthermore, \cref{tab:sotacomp} demonstrates that our method achieves a 26.8\% improvement in overall task completion rate compared to a state-of-the-art (SOTA) approach in~\cite{hou2024multitasklearningactivefaulttolerant}. Those extensive experiments show the effectiveness and robustness of our UMC framework. 

The main contributions of our work are:

\begin{compactitem}
    \item We systematically classify eight distinct failure scenarios, focusing primarily on sensor-only failures, detectable joint damage, and undetectable joint damage. For each scenario involving joint damage, we provide a detailed analysis of the corresponding tailored responses to specific impairments, such as restricted range of motion, reduced motor force, and limited linear velocity.
    
    \item We propose a UMC framework, a two-stage training pipeline that integrates standard pretraining on normal data with fine-tuning on custom damage training environments. The UMC incorporates a masking mechanism to address joint malfunctions and is compatible with both transformer-based and MLP-based action networks.
    
    \item Extensive experiments across three tasks demonstrate the robustness of our method under all eight damage conditions. Furthermore, our approach does not require prior knowledge of joint or limb malfunctions during inference, ensuring its strong adaptability in real-world applications. 
\end{compactitem}




\vspace{-2mm}
\section{Related Work}
\label{sec:relworks}

\textbf{Reinforcement Learning in Legged Robots.}
In recent years, reinforcement learning (RL) has gained traction in legged robots' control and locomotion tasks~\cite{strudel2020learningcombineprimitiveskills, tang2020learningagilelocomotionadversarial}. Some deep-learning-based RL methods are proposed to improve quadrupedal robots' stability across diverse terrains through combined simulated and real-world training~\cite{A2022ReinforcementLB}. In this work, we utilize the Proximal Policy Optimization (PPO) algorithm provided by Legged Gym for RL-based control of legged robots. Our approach focuses on enhancing locomotion control by reconstructing the Actor model, improving performance in complex environments.

\begin{figure*}
  \centering
  \includegraphics[width=0.95\linewidth]{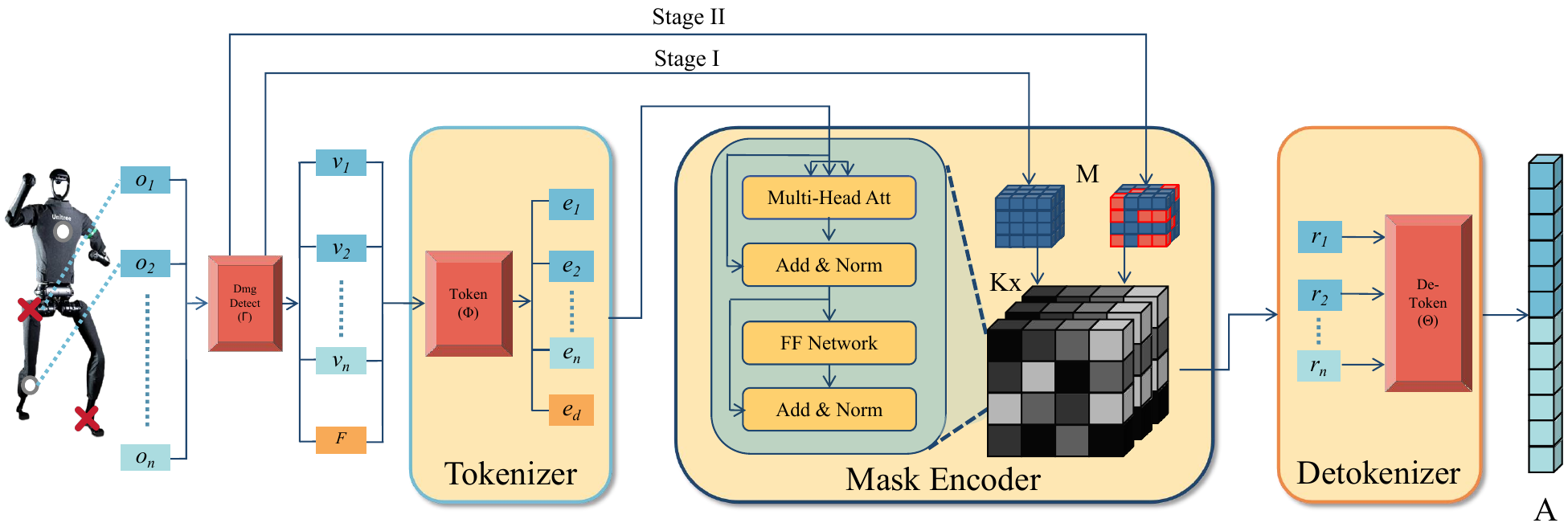}
  \vspace{-2mm}
  \caption{UMC system for transformer-based Actor-Model Architecture. \( K \) is the number of encoder layers. For more details of the architecture, please refer to \cref{sec:umcframework}.}
  \vspace{-5mm}
  \label{fig:trf_model}
\end{figure*}

\textbf{Self-recovering Robots.}
In recent years, self-recovering robots have attracted significant interest in robotics research~\cite{1044017, Guan2015FaultSF, 9249654}. As robotics technology matures, enabling legged robots to adapt to joint damage has become increasingly critical. However, few studies address this directly, and existing approaches often lack generalization, require excessive training data, or complex maintenance with conflicting strategies~\cite{kume2017mapbasedmultipolicyreinforcementlearning, nagabandi2019learningadaptdynamicrealworld, raileanu2020fastadaptationpolicydynamicsvalue, YangGANARL, ChenFADM, guo2023decentralizedmotorskilllearning}. Therefore, we aim to address as much damage as possible through one universal policy, providing insights for future research on this topic.

\textbf{Transformer Models in Robotics}
Transformers have gained sufficient popularity in various domains, including natural language processing~\cite{vaswani2023attentionneed}, computer vision~\cite{dosovitskiy2021imageworth16x16words,qi2022high}, and are now being explored in robotic control due to their ability to model sequential dependencies and capture complex long-range relationships in data. Recent studies have also demonstrated the effectiveness of transformer-based architectures in the robotics field~\cite{chen2021decisiontransformerreinforcementlearning, kurin2021bodycagerolemorphology,gupta2022metamorphlearninguniversalcontrollers,hong2022structureaware, radosavovic2024humanoidlocomotiontokenprediction,wan2024vint}, and our work is inspired by one such recent study called BodyTransformer~\cite{sferrazza2024bodytransformerleveragingrobot}. However, due to distinct focuses, BodyTransformer performs terribly under various damaged scenarios, while our methods could significantly resolve such problems.

\section{Method}
\label{sec:methods}

\begin{table}[t]
    \caption{Eight Damage Scenarios for Legged Robots. For `Sensor Status', `Damaged' means that the sensors cannot return correct observation readings, and `Functional' means that the sensors are well-functioned. `ROM' is the abbreviation of `range of motion'.}
    \vspace{-2mm}
    \centering
    \footnotesize
    \begin{tabular}{ccc}
        \toprule
        \textbf{Scenario} & \textbf{Sensor Status} & \textbf{Joint Damage Type} \\
        \midrule
        1 & Damaged & None \\
        2 & Damaged & ROM Restriction \\
        3 & Damaged & Reduced Motor Force \\
        4 & Damaged & Limited Linear Velocity \\
        5 & Functional & ROM Restriction \\
        6 & Functional & Reduced Motor Force \\
        7 & Functional & Limited Linear Velocity \\
        8 & Functional & None \\
        \bottomrule
    \end{tabular}
    \vspace{-6mm}
    \label{tab:damage_scenarios}
\end{table}

%
We aim to design a unified policy that enables the robot to complete tasks within various damage conditions.
We explore the robustness of legged robots by systematically analyzing various damage factors and proposing a unified malfunction controller (UMC) to address them.
First, we thoroughly summarize eight types of malfunctions given the reliability of sensors and joints. 
Sensors provide feedback on the robot's internal state through three key information inputs: position, velocity, and force of joint action. 
Sensor damage refers to blockage of only these inputs, whereas joint damage indicates a failure in the functionality of an actual joint.
On the other hand, damage related to the joint can be divided into a restricted range of motion, reduced motor force, and limited linear velocity. 
More details will be introduced in \cref{sec:mal}.
%

For the method, our proposed UMC is a model-free system that is highly robust to various damage factors within a unified framework. 
Specifically, our design adopts a two-stage training pipeline with a masking mechanism and is compatible with both the transformer and the MLP structure.
%
The masking mechanism used during training ensures that the network automatically ignores anomalous signals, allowing the trained policy to adapt to multiple types of damage. 
Meanwhile, the two-stage training pipeline retains the network’s ability to handle standard scenarios effectively.
%
%
%
%
%

In the following subsections, we first systematically analyze the various damage factors. 
Then, we will introduce the design of the UMC system, including the baseline framework, masking mechanism, and two-stage training pipeline.

\subsection{Malfunction Settings}
\label{sec:mal}

Compared to previous works that focused on self-diagnosis~\cite{Guan2015FaultSF,quamar2024reviewfaultdiagnosisfaulttolerant} or only tested limited damage types~\cite{YangGANARL}, we allow for as many different damage scenarios as possible, reflecting a wider range of real-world conditions, as shown in Table~\ref{tab:damage_scenarios}. 
%
For the sensor, we consider two kinds of statuses: `Damaged' and `Functional'.
The `Damaged' status indicates that the sensor cannot provide correct observation readings due to a malfunction and will only return a value of 0.
The `Functional' status indicates that the sensor operates correctly and can return accurate observation readings.

Furthermore, we have three categories of joint damage: range of motion restriction, reduced motor force, and limited linear velocity.
The range of motion restriction refers to the situation in which the joint's range of movement is limited due to external or internal factors. 
The reduced motor force occurs when the motor force output decreases due to wear or physical damage.
The limited linear velocity occurs when the joint speed is restricted, often triggered by overheating and thermal protection mechanisms.



For more details about our malfunction setting, please check our Appendix.


\subsection{UMC Framework}
\label{sec:umcframework}
Given the original joint observation inputs \( O = \{o_1, \dots, o_N\} \in 
\mathbb{R}^{N \times 3} \) that represent the position, velocity, and action information for the $N$ single degree-of-freedom (DOF) joints, we aim to output the next action instruction $A \in \mathbb{R}^{N \times 1}$ for each joint.
Then, $A$ is converted into a torque sequence which is applied as forces to each corresponding joint (DOF). 
This allows the robot to execute its next movement.
Since each joint has a single DOF, its position can be fully represented by a single scalar value, inherently making the position component one-dimensional.

Our UMC controller employs an actor-critic framework, a standard reinforcement learning strategy, for training.
%
As shown in \cref{fig:trf_model}, the actor model takes \( O \) as input and outputs \( A \), representing the robot's next action. 
The critic strategy, which has the same base architecture as the actor, evaluates action behaviour using proximal policy optimization (PPO). 
During inference, we only use the actor model to output a series of actions.
%
In our work, we focus on the actor model, leaving the critic strategy unchanged. This is because the critic needs to operate from a global perspective to accurately evaluate the actor's actions, so it will receive correct sensor signal values and does not require any mask-related modules.




\subsubsection{Actor Model}
\label{sec:actormodel}
For clarity, we consider the transformer structure as our base design for the UMC framework. We note in \cref{sec:MLPUMC} that UMC can be easily extended to the pure MLP network.

For the actor model, as shown in \cref{fig:trf_model}, we have a damage detection module and a base structure.
The damage detection module ensures that information from damaged joints does not interfere with that from functional joints.
The base structure consists of three main components: a tokenizer, a mask encoder, and a detokenizer. 
%
The tokenizer and detokenizer perform the transformation between the joint observation, a sequence of tokens and the action sequence, enabling seamless encoding and decoding processes.
Our core contribution lies in the design of a mask encoder along with mask strategy in two-stage training.
This design can capture dependencies and refine input representations using only the embeddings of well-functioning joints.
%

\noindent \textbf{Damage Detection Module.}
At first, \( O \) would be processed by \( \Gamma \), a damage detection module, to generate some information used in the base structure.
%
As shown in \cref{eq:dmgdetect}, \( \Gamma \) identifies joint failures in \( O \) and outputs two masking-related sequences.
\begin{equation}
V, M = \Gamma(O).
\label{eq:dmgdetect}
\end{equation}
The first output \( V  = \{v_1, \dots, v_N\} \in 
\mathbb{R}^{N \times 3} \) is transformed from \( O \) with its damaged joints being masked. 
Specifically, these damaged joints' observation inputs are zeroed out in \( O \) to effectively block its influence during all the following processes. 
For example, the $o_1$ in $O$ would be changed to zero vector and becomes $v_1$ if the first joint is damaged.
The second output \( M \in \mathbb{R}^{(N+1) \times (N+1)} \) is a masking matrix that encodes the failure information, which is added to the attention mechanism during training. 
Specifically, \( \Gamma \) sets all the positions corresponding to the damaged joints in \( M \) from zero to \( -\infty \). 
More details of \( M \) are delineated in \cref{sec:maskstrategy}.
%
%

\noindent \textbf{Base Structure.}
%
Our base structure follows a vanilla transformer design with a tokenizer, detokenizer, and mask encoder which consists of several stacked attention blocks.
%
%
Based on $V$, an additional sequence \( F \in \{-1, 1\}^{1 \times 3} \) is concatenated, forming the input of the tokenizer \( O'  \in \mathbb{R}^{(N+1) \times 3} \).
$F$ indicates whether potential malfunctions or irregularities in the robot's joints (not sensors) have been identified.
To ensure robustness, the dim of $F$ is three instead of one to prevent a single point of failure from causing complete system errors.
Once a joint malfunction is detected, the values in \( F\) would change from all -1 to all 1.
%

The \( O'\) is then processed by a tokenizer \( \Phi \), producing the joint embedding $E \in \mathbb{R}^{(N+1) \times D}$, where \( D\) is the embedding dimension.
\( \Phi \) applies projection and positional encoding operations for each joint, transforming \( O' \) into a format that corresponds to the input requirements of the mask encoder. 
Specifically, the projection operation consists of a set of linear layers, each of which maps an element in  \( O' \) into a higher-dimensional embedding space, while positional encoding uses a learnable embedding layer to encode each joint's position in the sequence.
Besides, this projection operation also ensures that the output embeddings corresponding to individual joints remain disentangled, preventing further interference between the joints and allowing flexible adaptation during masking.
\begin{equation}
E = \Phi(O').
\end{equation}

%

$E$ is then passed through a mask encoder $\Omega$. $\Omega$ consists of several stacked attention blocks where each block has a multi-head self-attention and feed-forward network module.
\begin{equation}
R = \Omega(E),
\end{equation}
where $R \in \mathbb{R}^{(N+1) \times D}$.
%

%

Finally, the action feature $R$ is processed by the detokenizer $\Theta$. $\Theta$ is a mapping layer designed to project \( R \) back to the action space. 
It consists of a set of linear layers, each corresponding to a specific joint. These layers independently map the feature representation of each joint in $R$ to the respective action outputs, yielding the final action sequence $A \in \mathbb{R}^{N \times 1}$. 
The number of the sequences is reduced from \( (N+1) \) to \( N \) because the additional dimension for the damage detection signals is only used for context during \( \Omega \) and is thus excluded from the final action-related process \(\Theta\).
\begin{equation}
A = \Theta(R).
\end{equation}
%
\subsubsection{Masking Strategy}
\label{sec:maskstrategy}

We add \( M \in \mathbb{R}^{(N+1) \times (N+1)} \) into the self-attention module of the attention block, where $N$ is the number of joint observation embeddings in \( E\), and the addition one refers to the damage detection embedding in \( E\).
%
This masking operation ensures that, after the softmax operation, the attention weights for the damaged joints become negligible, thereby excluding them from further contribution during the attention calculation. 
For example, the first attention block could be written as a formula below:
\begin{equation}
Output = \text{Softmax}\left(\frac{\mathbf{Q(E)} \mathbf{K(E)}^T}{\sqrt{d_k}} + M\right) \mathbf{V(E)},
\label{eq:attention}
\end{equation}
where \( \mathbf{Q(E)} \), \( \mathbf{K(E)} \), and \( \mathbf{V(E)} \) are the query, key, and value matrices derived from \( E \), and \( d_k \) is the dimensionality of the \( \mathbf{Q(E)} \) and \( \mathbf{K(E)} \).
The masking matrix \( M \) serves to block the positions corresponding to malfunctioning joints.

%

\noindent \textbf{Training Loss.}
The training loss consists of both actor and critic losses. Additionally, an entropy regularization term is included to promote exploration. This term encourages the agent to maintain a diverse set of actions and avoid premature convergence to suboptimal policies. Together, these components guide the optimization of both the policy and value functions.

The total loss function in PPO is defined as:
\begin{equation}
\mathbb{L} = \mathbb{L}_{\text{surrogate}} + \lambda_{1} \cdot \mathbb{L}_{\text{value}} + \lambda_{2} \cdot \mathbb{L}_{\text{entropy}},
\end{equation}
where $\lambda_{1}$ and $\lambda_{2}$ denote weight parameters. The $\mathbb{L}_{\text{surrogate}}$, $\mathbb{L}_{\text{value}}$ and $\mathbb{L}_{\text{entropy}}$ are the loss of policy surrogate, value function, and entropy regularization, respectively.

Please refer to our appendix for more details of those losses that are not the key points of our work.
\subsection{Two-Stage Training}

To enhance the robustness of legged robots in both normal and damaged conditions, we propose a two-stage training pipeline.
%
In the first stage, the network is pre-trained under normal conditions to establish a strong baseline performance.
In the second stage, it is fine-tuned across various conditions to adapt to diverse damage scenarios.
%
%
This approach ensures that the model excels in standard environments while maintaining the flexibility to handle unexpected challenges.
%

\noindent \textbf{Stage I.}
The objective is to train the model to perform tasks under undamaged conditions. This stage ensures that the robot acquires effective baseline capabilities in normal scenarios.

\noindent \textbf{Stage II.}
The model is fine-tuned to handle a range of joint and sensor damage conditions while retaining its baseline capabilities from Stage I. 
Specifically, we uniformly sample the four subcategories by default in the following:
%
\vspace{-3mm}
\begin{enumerate}[label=(\roman*)]
    \item \textbf{Normal conditions}: This subcategory includes scenario 8 in \cref{tab:damage_scenarios}, ensuring that the model retains its baseline performance.
    \item \textbf{Sensor-only damage}: This subcategory includes scenario 1 in \cref{tab:damage_scenarios}, which simulates sensor failures by blocking (zeroing out) the input of specific joints while leaving their performance unaffected. 
    %
    \item \textbf{Detectable joint damage}: This subcategory includes scenarios 2, 3, and 4 in \cref{tab:damage_scenarios}, which simulates partial limb damage by adding joint and sensor damage to certain joints and allowing the detection of damage information.
    %
    %
    \item \textbf{Undetectable joint damage}: This subcategory encompasses scenarios 6 and 7 as described in \cref{tab:damage_scenarios}, which simulates joint-specific damages. Notably, no maskings are applied in this subcategory and all sensors remain operational.
\end{enumerate}

\vspace{-3mm}
For detailed explanations and parameters of the four subcategories, please refer to the Appendix.

\subsection{Transition from Transformer to MLP Structure}
\label{sec:MLPUMC}

For the MLP structure, the main differences lie in part of the damage detection module \( \Gamma' \) and the base structure. 
This is because MLP does not obtain a self-attention module and does not need to handle structured input through a tokenizer.
In \( \Gamma' \), we retain only the first masking step, which involves zeroing out the observation inputs of the damaged joints, transforming $O$ into $O'$. Then, for the base structure, the input $O' \in \mathbb{R}^{(N+1) \times 3}$ is first flattened into $O'' \in \mathbb{R}^{3N+3}$. After that, the flattened input $O''$ is passed through multiple hidden layers of the MLP. Finally, the MLP processes the data and outputs the action sequence $A \in \mathbb{R}^{N \times 1}$. A detailed workflow diagram is provided in the Appendix.


\section{Experiments}
\label{sec:exp}
In this section, we begin by describing the experimental setup, followed by evaluation metrics. 
%
Next, we present both quantitative and qualitative comparison results with existing methods. 
Finally, we conduct extensive ablation studies to validate the effectiveness of the proposed model.

\subsection{Experimental Setup}
\label{sec:setup}


\noindent\textbf{Implementation Details.} All models are trained on a single Nvidia A6000 GPU and evaluated using PPO-based Reinforcement Learning~\cite{schulman2017proximalpolicyoptimizationalgorithms} for three different robot locomotion tasks, which are the A1-Walk task from ParkourGym~\cite{zhuang2023robot} and the H1 and G1 tasks from Unitree. 
For SOTA work comparison, we selected the Solo8 task~\cite{9015985}.
Among them, A1 and Solo8 are quadruped robots, while H1 and G1 are humanoid robots. 
All these locomotion tasks are performed within the IsaacGym environment~\cite{makoviychuk2021isaacgymhighperformance}, managed by the Legged Gym Repository~\cite{rudin2022learningwalkminutesusing}. 
We provide transformer-based and MLP-based UMC architectures. 
Please refer to the Appendix for more details on model configurations, malfunction limits, and other parameters. 

\noindent \textbf{Damage Settings During Inference.}
During inference, we apply three distinct damage settings for every task, all of which differ from those used during the training stage. 
First, for the rough terrain task A1, the robots operate on a terrain that is different from those encountered during training. 
%
Second, different joint combinations are randomly selected using various seeds to introduce damage, thereby preventing the model from relying on prior knowledge learned from the training set. 
Third, malfunctions are introduced at different times during inference to simulate more different robot gaits when suffering damage and different combinations of joint damage. 
%
For example, in one environment, a robot may lift one of its front legs, whereas in another, the same leg may point downward when its corresponding joints are damaged.
For more details, please refer to the Appendix.

\begin{table}[t!]
  \caption{Average Performance of Different Models on the A1 Task Across Eight Damage Conditions.}
  \centering
  \vspace{-2mm}
\renewcommand{\arraystretch}{1}
    \resizebox{\linewidth}{!}{
   \begin{tabular}{@{}ccccccc@{}}
    \toprule
    Methods & 1 unit $\uparrow$ & 2 unit $\uparrow$ & 3 unit $\uparrow$ & 4 unit $\uparrow$ & 5 unit $\uparrow$ & failed $\downarrow$  \\
    \midrule
    \myrowcolour%
    Trf-NM & 81\% & 64\% & 54\% & 46\% & 38\% & 7\% \\
    MLP-NM & 67\% & 54\% & 48\% & 43\% & 46\% & 23\%\\
    \myrowcolour%
    BodyTrf & 84\% & 68\% & 56\% & 45\% & 36\% & 10\%\\
    MLP-UMC (Ours) & \textcolor{blue}{93\%} & \textcolor{blue}{88\%} & \textcolor{blue}{83\%} & \textcolor{blue}{78\%} & \textcolor{blue}{66\%} & \textcolor{blue}{4\%}\\
    \myrowcolour%
    Trf-UMC (Ours) & \textcolor{red}{97\%} & \textcolor{red}{95\%} & \textcolor{red}{91\%} & \textcolor{red}{84\%} & \textcolor{red}{72\%} & \textcolor{red}{2\%}\\
    \bottomrule
  \end{tabular}}
  \vspace{-2mm}
  \label{tab:a1sum}
\end{table}

\begin{table}[t!]
  \caption{Average Performance of Different Models on the G1 Task Across Eight Damage Conditions.}
  \vspace{-3mm}
  \centering
\renewcommand{\arraystretch}{1}
    \resizebox{\linewidth}{!}{
   \begin{tabular}{@{}ccccccc@{}}
    \toprule
     Methods & 1 unit $\uparrow$ & 2 unit $\uparrow$ & 3 unit $\uparrow$ & 4 unit $\uparrow$ & 5 unit $\uparrow$ & failed $\downarrow$  \\
    \midrule
    \myrowcolour%
    Trf-NM & 44\% & 43\% & 42\% & 39\% & 35\% & 56\% \\
    MLP-NM & 33\% & 33\% & 31\% & 29\% & 25\% & 67\%\\
    \myrowcolour%
    BodyTrf & 52\% & 51\% & 49\% & 45\% & 40\% & 48\%\\
    MLP-UMC (Ours) & \textcolor{blue}{86\%} & \textcolor{blue}{85\%} & \textcolor{blue}{80\%} & \textcolor{blue}{73\%} & \textcolor{blue}{64\%} & \textcolor{blue}{14\%}\\
    \myrowcolour%
    Trf-UMC (Ours) & \textcolor{red}{91\%} & \textcolor{red}{90\%} & \textcolor{red}{85\%} & \textcolor{red}{79\%} & \textcolor{red}{70\%} & \textcolor{red}{9\%}\\
    \bottomrule
  \end{tabular}}
  \vspace{-3mm}
  \label{tab:g1sum}
\end{table}

\begin{table}[t!]
  \caption{Average Performance of Different Models on the H1 Task Across Eight Damage Conditions.}
  \centering
  \vspace{-2mm}
\renewcommand{\arraystretch}{1}
    \resizebox{\linewidth}{!}{
   \begin{tabular}{@{}ccccccc@{}}
    \toprule
     Methods & 1 unit $\uparrow$ & 2 unit $\uparrow$ & 3 unit $\uparrow$ & 4 unit $\uparrow$ & 5 unit $\uparrow$ & failed $\downarrow$  \\
    \midrule
    \myrowcolour%
    Trf-NM & 57\% & 56\% & 55\% & 51\% & 46\% & 43\% \\
    MLP-NM & 57\% & 57\% & 55\% & 52\% & 47\% & 43\%\\
    \myrowcolour%
    BodyTrf & 53\% & 53\% & 51\% & 48\% & 44\% & 47\%\\
    MLP-UMC (Ours)  & \textcolor{red}{97\%} & \textcolor{red}{97\%} & \textcolor{red}{94\%} & \textcolor{red}{88\%} & \textcolor{red}{80\%} & \textcolor{red}{3\%}\\
    \myrowcolour%
    Trf-UMC (Ours) & \textcolor{blue}{95\%} & \textcolor{blue}{94\%} & \textcolor{blue}{90\%} & \textcolor{blue}{84\%} & \textcolor{blue}{75\%} & \textcolor{blue}{5\%}\\
    \bottomrule
  \end{tabular}}
  \vspace{-4mm}
  \label{tab:h1sum}
\end{table}

\begin{table}[t!]
  \caption{Average Performance of Different Models on the Solo8 Task Across Eight Damage Conditions. `MT-FTC' is the method proposed in~\cite{hou2024multitasklearningactivefaulttolerant}.}
  \centering
  \vspace{-2mm}
\renewcommand{\arraystretch}{1}
    \resizebox{\linewidth}{!}{
   \begin{tabular}{@{}ccccccc@{}}
    \toprule
     Methods & 0.5 unit $\uparrow$ & 1 unit $\uparrow$ & 1.5 unit $\uparrow$ & 2 unit $\uparrow$ & 2.5 unit $\uparrow$ & failed $\downarrow$  \\
    \midrule
    \myrowcolour%
    MT-FTC & 39\% & 31\% & 30\% & 30\% & 29\% & 46\%\\
      Trf-UMC (Ours) & \textcolor{red}{73\%} & \textcolor{red}{67\%} & \textcolor{red}{60\%} & \textcolor{red}{52\%} & \textcolor{red}{41\%} & \textcolor{red}{12\%} \\
    \bottomrule
  \end{tabular}}
  \vspace{-6mm}
  \label{tab:sotacomp}
\end{table}

\begin{figure*}[t!]
  \centering
  \includegraphics[width=0.95\linewidth]{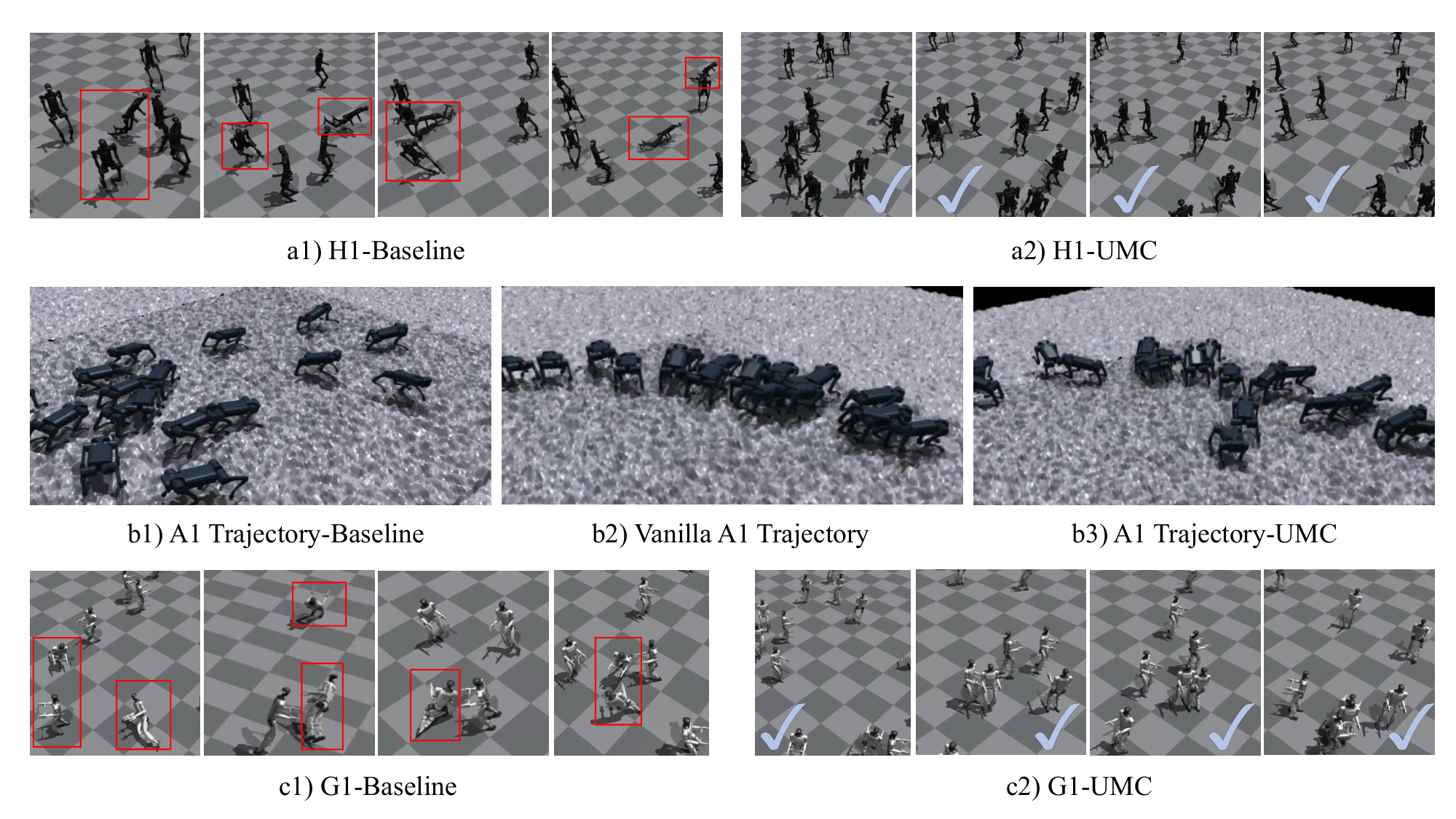}
  \vspace{-3mm}
  \caption{Qualitative Comparison Between Methods Under Damaged Scenarios. 
  \label{sec:res}
  %
  %
  %
  `Baseline' refers to robots trained using baseline methods, while `UMC' denotes robots trained with the UMC method. Figure `b2)' shows a snapshot of the original trajectory at a specific time point under undamaged conditions, while b1) and b3) are in damaged conditions.}
  \label{fig:qualitative}
  \vspace{-4mm}
\end{figure*}

\subsection{Evaluation Metrics}
\label{sec:evalmetric}


%

After legged robots walk certain steps under normal conditions, we apply damage to them and record the initial position. During the subsequent episodes, we record the following comprehensive metrics to validate the locomotion capabilities of legged robots.

%

%
Specifically, we evaluate whether the robots can move beyond the radii of 1, 2, 3, 4, and 5 units (0.5, 1, 1.5, 2, and 2.5 units for the Solo8 task) from their initial positions without falling.
If the robot can maintain its original trajectory despite the damage, this distance should correlate positively with time. 
Therefore, a greater distance travelled indicates a more effective policy, as it allows the robot to move further given the limited time.
Additionally, legged robots that exhibit any falling behaviour are excluded from the previous distance statistics and are instead counted in a separate metric labelled as `failed'. 

\begin{figure}[t!]
    \centering
    \vspace{-2mm}
    \begin{subfigure}[b]{0.2\textwidth}  
        \centering
        \includegraphics[width=\textwidth]{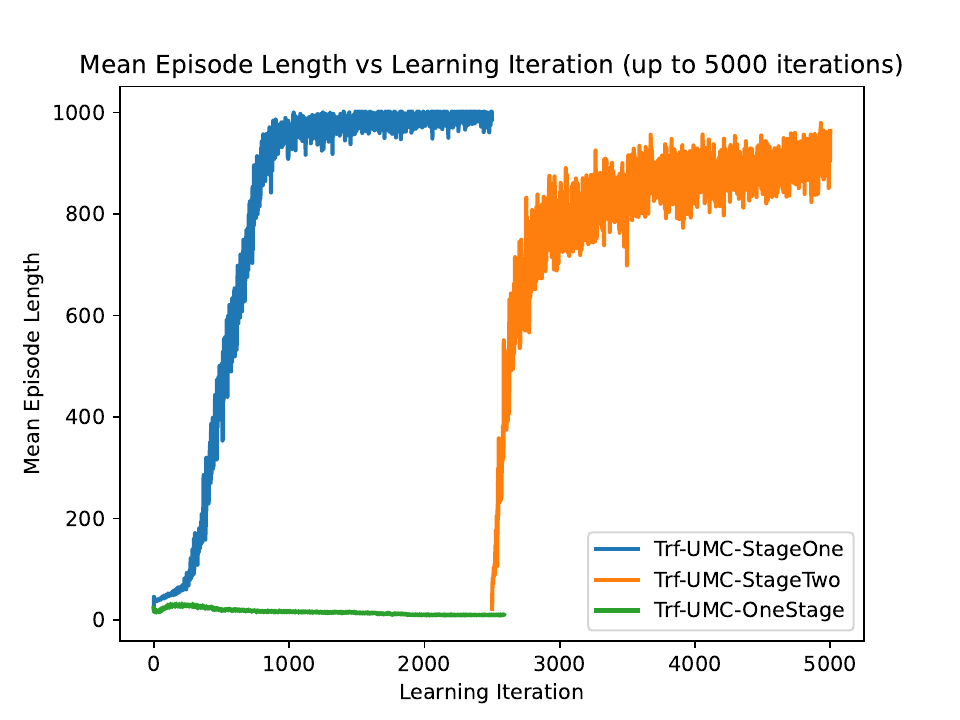}
        \caption{Episode \text{v.s.} Iteration}
    \end{subfigure}
    \begin{subfigure}[b]{0.2\textwidth}  
        \centering
        \includegraphics[width=\textwidth]{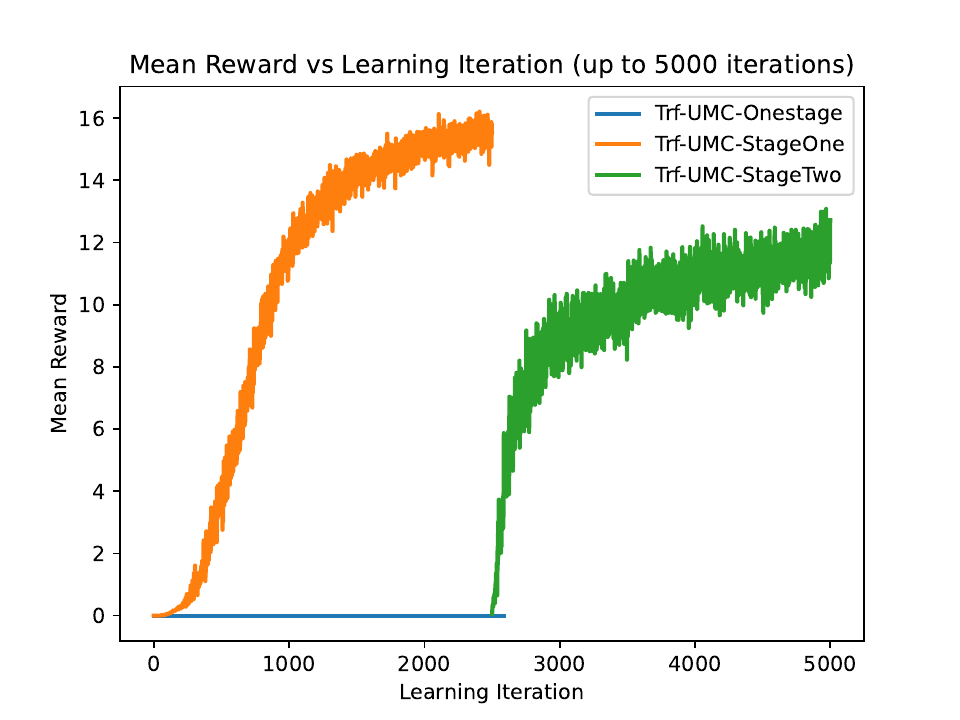}
        \caption{Reward \text{v.s.} Iteration}
    \end{subfigure}
    \caption{Comparison of One-Stage and Two-Stage Training in the G1 Task.}
    \label{fig:abl_twostage}
    \vspace{-7mm}
\end{figure}

\vspace{-2mm}
\subsection{Overall Results}

As shown in 
\cref{tab:a1sum}, \cref{tab:g1sum},  and \cref{tab:h1sum}, we present the average performance for each task with our metrics.
The averaged results for each model are computed by summing performance across eight damage scenarios and three inference settings, demonstrating the superiority of our UMC framework. More statistics are shown in the Appendix.

The UMC significantly reduces the number of fall cases on the H1, G1, and A1 tasks, and performs better against the BodyTransformer. 
Compared to the normally trained transformer, our baseline, the transformer-based UMC achieves an average reduction in failure rates across eight types of damage by 5\%, 38\%, and 47\% in tasks A1, H1 and G1, respectively. 
Similarly, MLP-based UMC demonstrates reductions of 19\%, 40\%, and 53\%, respectively. 
UMC prompts robots to rely more on their functional limbs when dealing with various failures, thereby effectively reducing the impact of damaged joints on their actions.
In real-world applications, especially for humanoid robots, falls during task execution could lead to substantial physical damage and result in high financial costs. 
Instead, our method effectively minimizes such risk of damage and associated costs. 

For the task completion performance of the transformer architecture, taking the A1-Walk task as an example, UMC improves the robot's achievement rates across the 1-unit to 5-unit metrics by 16\%, 31\%, 37\%, 38\%, and 34\%, respectively. 
For the MLP architecture, the robot also demonstrates improvements of 38\%, 38\%, 35\%, 33\%, and 29\% on the H1 task.
We attribute such improvements to our masking mechanism as it enables rapid adaptation to new types of damage without the need to switch to a new policy. 
Therefore, robots can respond to sudden damage more quickly and adjust their gait accordingly.

%
The \cref{fig:qualitative} indicate that UMC can handle various damage conditions and effectively maintain the intended trajectory, which demonstrates that UMC can reduce the impact of damages from another perspective.
Moreover, \cref{fig:teaser}(c) show that UMC retains and even slightly enhances the robot's performance under normal, undamaged conditions across three tasks.
This improvement comes from the design of our two-stage pipeline, which ensures that the trained robots maintain their performance under normal conditions. 

Additionally, we compared UMC with the method proposed in~\cite{hou2024multitasklearningactivefaulttolerant}, referred to as `MT-FTC', on the Solo8 task. The results in \cref{tab:sotacomp} indicate that UMC achieves a 26.8\% improvement in task completion performance and reduces the fall rate by 34\% compared to `MT-FTC'. These results demonstrate that UMC exhibits greater flexibility in adapting to various conditions compared to SOTA methods.

Furthermore, a comparison of \cref{tab:a1sum}, \cref{tab:g1sum}, and \cref{tab:h1sum} reveals that our baselines degrades less in humanoid robot tasks than in quadruped robot tasks as the metric increases from 1 to 5 units.
We attribute this to the structural differences between humanoid robots and quadruped ones like A1.
Unlike A1, which can easily remain upright and stable despite finding it difficult to move forward due to damage, humanoid robots face greater challenges in maintaining balance during movement.

\begin{table}[t!]
  \caption{Average Performance of Transformer-Based UMC with Different Stage-II Environment Settings. 
  The ratios correspond to four training scenarios in Stage II from left to right: `Normal', `Undamaged', `Sensor-only Damage', `Detectable Joint Damage', and `Undetectable Joint Damage'. 
  %
  }
  \vspace{-2mm}
  \label{sec:abl_ratio}
  \centering
\renewcommand{\arraystretch}{1}
    \resizebox{\linewidth}{!}{
   \begin{tabular}{@{}ccccccc@{}}
    \toprule
     Ratios & 1 unit $\uparrow$ & 2 unit $\uparrow$ & 3 unit $\uparrow$ & 4 unit $\uparrow$ & 5 unit $\uparrow$ & failed $\downarrow$  \\
    \midrule
    \myrowcolour%
    1:1:1:0 & 89\% & 84\% & 79\% & 73\% & 66\% & 10\% \\
    1:1:0:1 & 90\% & 83\% & 75\% & 65\% & 50\% & 2\%\\
    \myrowcolour%
    1:0:1:1 & 97\% & 94\% & 88\% & 80\% & 68\% & 2\%\\
    0:1:1:1 & 97\% & \textcolor{red}{95\%} & 90\% & 82\% & 69\% & \textcolor{red}{2\%}\\
    \myrowcolour%
    1:2:2:1 & 97\% & 94\% & 89\% & 81\% & 68\% & 2\%\\
    1:3:3:1 & 97\% & 94\% & 87\% & 77\% & 66\% & 2\%\\
    \myrowcolour%
    Default(1:1:1:1) & \textcolor{red}{97\%} & 94\% & \textcolor{red}{90\%} & \textcolor{red}{84\%} & \textcolor{red}{74\%} & 3\%\\
    \bottomrule
  \end{tabular}}
  \label{tab:abl_ratio}
  \vspace{-6mm}
\end{table}

\subsection{Ablation Study}

In this section, we focus on the overall ablation results, including mask strategy, training stages, sampling ratio and paradigms for each damage scenario. %
For more details and results, please refer to the Appendix.

\label{sec:ablation}


\noindent \textbf{Training Stage.}
We ablate our training stages in~\cref{fig:abl_twostage}. 
The blue and green curves represent the two training stages in our method, while the orange curve shows the one-stage training, training solely on our Stage II damaged environments. 
The curves show that the one-stage setting eventually fails to converge within 2500 iterations in the G1 task, whereas the two-stage approach proves effective across all three tasks. 
We attribute this to the introduction of an overly complex training set in the initial stage, which hindered the model's convergence and ultimately prevented the discovery of effective policies.
Thus, we selected the two-stage training process for our workflow.

\noindent \textbf{Sampling Ratio.}
We ablate the sampling ratio of the four training subcategories in Stage II. 
\cref{tab:abl_ratio} demonstrates that the default ratio of 1:1:1:1 achieves the best overall performance. 
The potential reason is two-fold. 
First, unlike conditions that exclude certain subcategories (e.g., 1:1:1:0), the model could learn all four damaged scenarios with the default ratio during Stage II. 
Second, compared to ratios that focus more heavily on detectable damage (e.g. 1:2:2:1), the default ratio achieves better balance and thus enables the model to learn to handle various types of damage more comprehensively.

\noindent \textbf{Masking Value.}
\label{sec:abl_maskvalue}
We ablate the masking value adopted in our masking mechanism, where the value indicates the observation of the damaged joint. 
\cref{tab:abl_maskvalue} shows that zero outperforms the two out-of-distribution values `-100' and `100'.
We attribute this to out-of-distribution values exerting greater influence on the model’s decision-making.
For example, if an action in the observation input is set to 100, though out of range, it still carries some information that the model can analyze. 
And the impact of such information is greater than that of the default value of 0. 
Additionally, excessively large values may result in disproportionate rewards or penalties, further affecting the model's performance.

\noindent \textbf{Paradigms.}
We ablate the foundational paradigms of UMC. In addition to the stage-based approach, we tested a curriculum-based learning strategy, progressing through increasing levels of difficulty: no damage, undetectable joint damage, sensor-only damage, and detectable joint damage. As shown in \cref{tab:abl_paradigm}, the curriculum-based approach performed slightly worse than the stage-based method, likely because its progressive focus on higher-difficulty tasks reduces its adaptability to lower-difficulty scenarios, resulting in suboptimal overall performance.

\begin{table}[t!]
  \caption{Average Performance of Transformer-Based UMC with Different Masking Values.}
  \vspace{-2mm}
  \centering
\renewcommand{\arraystretch}{1}
    \resizebox{\linewidth}{!}{
   \begin{tabular}{@{}ccccccc@{}}
    \toprule
     Values & 1 unit $\uparrow$ & 2 unit $\uparrow$ & 3 unit $\uparrow$ & 4 unit $\uparrow$ & 5 unit $\uparrow$ & failed $\downarrow$  \\
    \midrule
    \myrowcolour%
    100 & 96\% & 93\% & 87\% & 79\% & 67\% & 3\% \\
    -100 & 95\% & 91\% & 86\% & 79\% & 68\% & 4\%\\
    \myrowcolour%
    Default(0) & \textcolor{red}{97\%} & \textcolor{red}{94\%} & \textcolor{red}{90\%} & \textcolor{red}{84\%} & \textcolor{red}{74\%} & \textcolor{red}{3\%}\\
    \bottomrule
  \end{tabular}}
  \label{tab:abl_maskvalue}
  \vspace{-3mm}
\end{table}

\begin{table}[t!]
  \caption{Average Performance of Transformer-Based UMC with Different Paradigms.}
  \vspace{-2mm}
  \centering
\renewcommand{\arraystretch}{1}
    \resizebox{\linewidth}{!}{
   \begin{tabular}{@{}ccccccc@{}}
    \toprule
     Paradigms & 1 unit $\uparrow$ & 2 unit $\uparrow$ & 3 unit $\uparrow$ & 4 unit $\uparrow$ & 5 unit $\uparrow$ & failed $\downarrow$  \\
    \midrule
    \myrowcolour%
    Curriculum-Based & 92\% & 88\% & 84\% & 79\% & 67\% & 6\% \\
    Stage-Based & \textcolor{red}{97\%} & \textcolor{red}{94\%} & \textcolor{red}{90\%} & \textcolor{red}{84\%} & \textcolor{red}{74\%} & \textcolor{red}{3\%}\\
    \bottomrule
  \end{tabular}}
  \label{tab:abl_paradigm}
  \vspace{-6mm}
\end{table}


\section{Conclusion}
\label{sec:con}
In this paper, we present UMC, a unified, model-free framework that substantially improves the resilience of legged robots facing various failure scenarios, including sensor malfunctions and joint issues such as restricted motion, weakened motor, or limited velocity. 
Our approach UMC adopts two training stages that enable fast adaption to damaged conditions while allowing the robots to perform well in undamaged normal states. 
Specifically, UMC incorporates a masking strategy, isolating faulty joints, allowing the robot to compensate by emphasizing unaffected limbs and adapting dynamically to diverse damage patterns.
Experimental results validate that our UMC consistently improves both transformer and MLP architectures across different robot models and tasks, markedly reducing failure rates and improving task success under variable damage conditions, further improving the adaptability of robotic systems in challenging environments.


\section*{Impact Statement}
This paper presents work whose objective is to advance the field of fault tolerance in robotics. 
There are many potential societal consequences of our work, none of which we feel should be specifically highlighted here.

\nocite{langley00}

\bibliography{example_paper}
\bibliographystyle{icml2025}

\newpage 
\appendix
\onecolumn
\renewcommand{\thesection}{\Alph{section}} 

This Appendix provides additional details and empirical results to demonstrate the benefits of our proposed method. The contents of this Appendix are structured as follows:

\begin{itemize}
    \item \textbf{UMC Implementation Details}: Detailed parameters of the UMC structures. 
    \item \textbf{Details of the Stage II Training Environment}: Explanations of the design of the Stage-II training environment.
    \item \textbf{Malfunction and Experiment Settings}: Detailed parameters of the malfunction and other settings during training and inference.
    \item \textbf{More Experiment Results}: More experiment results that are not shown and illustrated in the body of the paper.
    \item \textbf{More Ablation Results}: More details and ablation experiment results that are not shown and illustrated in the ablation parts.
    \item \textbf{Loss}: Detailed Introduction of the loss functions adopted in our work.

\end{itemize}

\section{UMC Implementation Details}

In this section, as shown in \cref{tab:model_param}, we provide detailed experimental parameters of our UMC structure to facilitate reproducibility and related operations.

\begin{table}[h]
  \caption{Detailed Parameters of the transformer-based and MLP-based Actor Model.}
  \centering
   \begin{tabular}{@{}l|cc@{}}
    \toprule
    Parameter & MLP & Transformers \\
    \midrule
    Stage-One Epochs  & 2500 & 2500 \\
    Stage-Two Epochs & 2500 & 2500 \\
    Encoder Layers & 4 & 4 \\
    Embedding Input Size & N/A & 120 \\
    Feedforward Size & [256,512,256,256] & 128 \\
    Attention Heads & N/A & 2 \\
    \midrule
    Total Parameters & 345,100 & 366,164 \\
    \bottomrule
  \end{tabular}

  \label{tab:model_param}
\end{table}

\begin{figure*}
  \centering
  \includegraphics[width=1.0\linewidth]{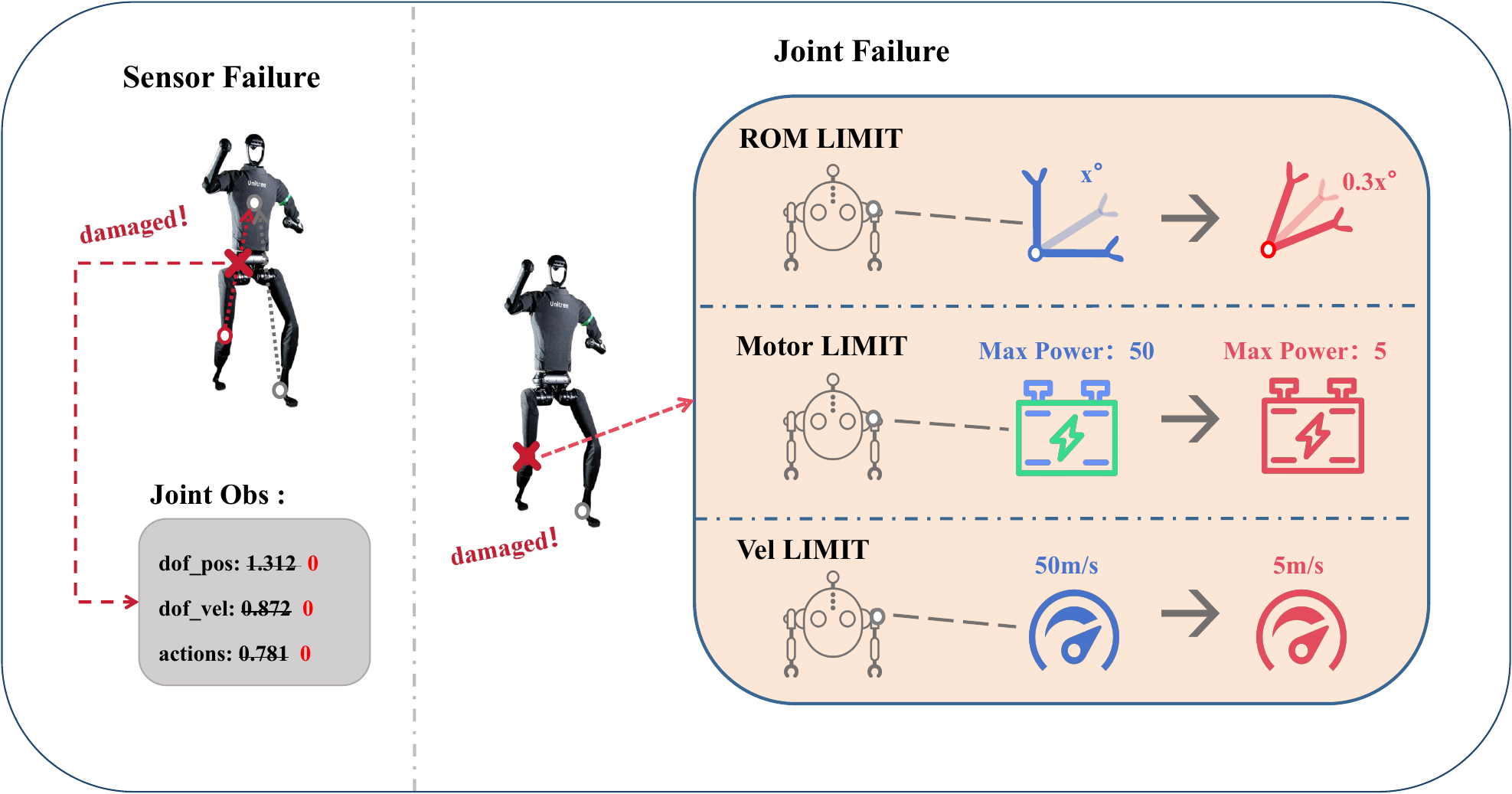}
  \caption{Demonstration of different damage conditions.}
  \label{fig:malfunction_demonstration}
\end{figure*}

\begin{figure}
  \centering
  \includegraphics[width=1.0\linewidth]{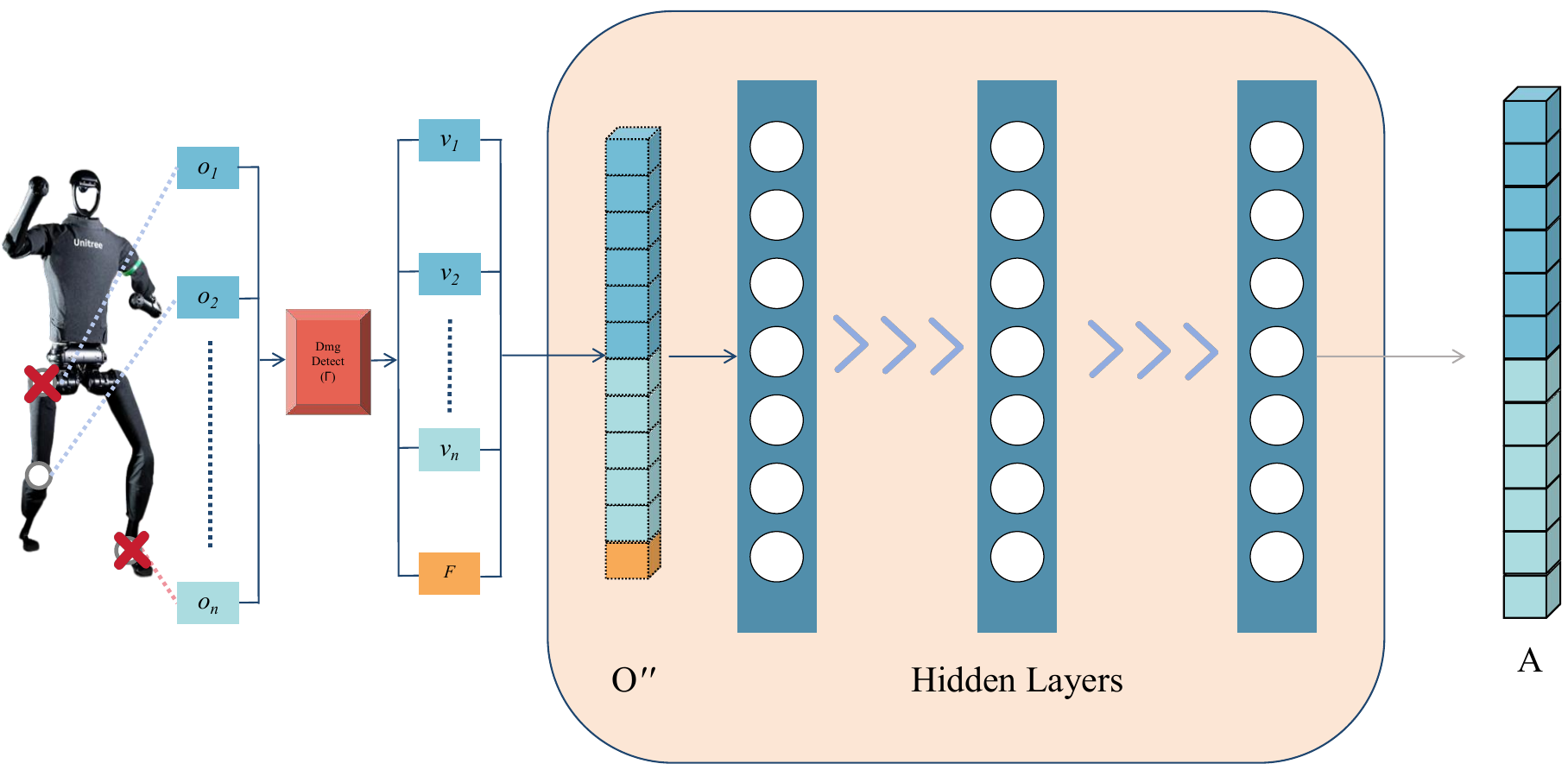}
  \caption{UMC Framework for MLP-based Actor-Model Architecture.}
  \label{fig:mlp_structure}
\end{figure}

\begin{table}[h]
    \caption{Malfunction Settings for Training and Inference in the A1-Walk Task.}
  \centering
  \begin{subtable}[t]{\linewidth}
  \caption{Training Parameters.}
    \centering
    \begin{tabular}{@{}l|c@{}}
      \toprule
      Parameter & Values \\
      \midrule
      Num Envs & 7400  \\
      Random Damage Range & [2,4]  \\
      ROM Limit & Random 30\% \\
      Motor Limit & 5 \\
      Velocity Limit & 3 \\
      Track Width & 1.6 \\
      Track Block Length & 2.0 \\
      Border Size & 8 \\
      Perlin Noise Seed & 1 \\
      Random Malfunction Seed & 42 \\
       Episode Length & 1000 \\
      Malfunction Timing & N/A \\
      \bottomrule
    \end{tabular}

  \end{subtable}
  \\[0.5cm] 
  \begin{subtable}[t]{\linewidth}

    \caption{Inference Parameters.}
    \centering
    \begin{tabular}{@{}l|c@{}}
      \toprule
      Parameter & Values \\
      \midrule
      Num Envs & 4000  \\
      Random Damage Range & [4,5]  \\
      ROM Limit & Random 10\% \\
      Motor Limit & 8 \\
      Velocity Limit & 3 \\
      Track Width & 6.0 \\
      Track Block Length & 6.0 \\
      Border Size & 4 \\
      Perlin Noise Seed & [100, 25, 75] \\
      Random Malfunction Seed & [1, 800, 50] \\
      Malfunction Timing & [75, 100, 125] \\
       Episode Length & 750 \\
      \bottomrule
    \end{tabular}
    
  \end{subtable}

  \label{tab:a1_mal}
\end{table}

\begin{table}[h]
 \caption{Malfunction Settings for Training and Inference in the H1 Task.}
  \centering
  \begin{subtable}[t]{\linewidth}
  \caption{Training Parameters.}
      \centering
    \begin{tabular}{@{}l|c@{}}
      \toprule
      Parameter & Values \\
      \midrule
      Num Envs & 10000  \\
      Random Damage Range & [2,4]  \\
      ROM Limit & Random 30\% \\
      Motor Limit & 10 \\
      Velocity Limit & 5 \\
      Random Malfunction Seed & 42 \\
      Malfunction Timing & N/A \\
       Episode Length & 1000 \\

      \bottomrule
    \end{tabular}
    
  \end{subtable}
  \\[0.5cm] 
  \begin{subtable}[t]{\linewidth}
  \caption{Inference Parameters.}
 \centering
    \begin{tabular}{@{}l|c@{}}
      \toprule
      Parameter & Values \\
      \midrule
      Num Envs & 8192  \\
      Random Damage Range & [2,3]  \\
      ROM Limit & Random 30\% \\
      Motor Limit & 8 \\
      Velocity Limit & 3 \\
      Random Malfunction Seed & [1, 50, 75] \\
      Malfunction Timing & [75, 100, 125] \\
        Episode Length & 750 \\
      \bottomrule
    \end{tabular}
    
  \end{subtable}

  \label{tab:h1_mal}
\end{table}

\begin{table}[h]
  \caption{Malfunction Settings for Training and Inference in the Unitree-G1 Task.}
  \centering
  \begin{subtable}[t]{\linewidth}
  \caption{Training Parameters.}
    \centering
    \begin{tabular}{@{}l|c@{}}
      \toprule
      Parameter & Values \\
      \midrule
      Num Envs & 8192  \\
      Random Damage Range & [2,4]  \\
      ROM Limit & Random 30\% \\
      Motor Limit for Hip Joints & 8 \\
      Motor Limit for Knee Joints & 13 \\
      Motor Limit for Ankle Joints & 4 \\
      Velocity Limit & 3 \\
      Random Malfunction Seed & 42 \\
      Episode Length & 1000 \\
      Malfunction Timing & N/A \\
      \bottomrule
    \end{tabular}
    
  \end{subtable}
  \\[0.5cm] 
  \begin{subtable}[t]{\linewidth}
  \caption{Inference Parameters.}
    \centering
    \begin{tabular}{@{}l|c@{}}
      \toprule
      Parameter & Values \\
      \midrule
      Num Envs & 10000  \\
      Random Damage Range & [2,4]  \\
      ROM Limit & Random 30\% \\
      Motor Limit for All Joints & 5 \\
      Velocity Limit & 3 \\
      Random Malfunction Seed & [1, 50, 75] \\
      Malfunction Timing & [75, 100, 125] \\
          Episode Length & 750 \\
      \bottomrule
    \end{tabular}
    
  \end{subtable}

  \label{tab:g1_mal}
\end{table}

\begin{table}[h]
  \caption{Malfunction Settings for Training and Inference in the Solo8 Task.}
  \centering
  \begin{subtable}[t]{\linewidth}
  \caption{Training Parameters.}
    \centering
    \begin{tabular}{@{}l|c@{}}
      \toprule
      Parameter & Values \\
      \midrule
      Num Envs & 4096  \\
      Random Damage Range & [1,3]  \\
      ROM Limit & Random 30\% \\
      Motor Limit & 0.75 \\
      Velocity Limit & 5 \\
      Random Malfunction Seed & 42 \\
      Malfunction Timing & N/A \\
      Episode Length & 1000 \\
      \bottomrule
    \end{tabular}
    
  \end{subtable}
  \\[0.5cm] 
  \begin{subtable}[t]{\linewidth}
  \caption{Inference Parameters.}
    \centering
    \begin{tabular}{@{}l|c@{}}
      \toprule
      Parameter & Values \\
      \midrule
      Num Envs & 4096  \\
      Random Damage Range & [2,4]  \\
      ROM Limit & Random 30\% \\
      Motor Limit for All Joints & 5 \\
      Velocity Limit & 3 \\
      Random Malfunction Seed & 50 \\
      Malfunction Timing & 100 \\
      Episode Length & 1000 \\
      \bottomrule
    \end{tabular}
    
  \end{subtable}

  \label{tab:solo8}
\end{table}

\section{Details of the Stage II Training Environment}
To enable our model to adapt to a wide range of damage scenarios, we thoughtfully designed four subcategories in the training environment used during Stage II. Explanations of how we design these scenarios will be illustrated in the following:

\begin{itemize}
    \item \textbf{Normal Scenarios}: We add this scenario in the second training stage to ensure that the model could retain the ability to perform well under no-damaged scenarios.

    \item \textbf{Senor-only Damage Scenarios}: This scenario represents cases where the joint functions normally but fails to return data to the central controller due to sensor damage. For such cases, the observation input of the affected joint would be reset back to a default value of zero due to no detected signals, while joints could perform as usual.

    \item \textbf{Detectable Joint-Damage Scenarios}: In this scenario, all three types of damage are applied together to randomly selected joints for each environment. We also damage the corresponding sensor as joint damage is likely to cause sensor failure in real-world conditions. And even if the sensors are currently functional, their data may be unreliable, and their continued functionality cannot be guaranteed. Thus, we assume a worst-case scenario where, whenever a joint's damage is detected, its corresponding sensor also will not work.

    \item \textbf{Undetectable Joint-Damage Scenarios}: The objective of this subcategory is to improve system robustness in situations where failures cannot be directly detected. In this subcategory, two types of joint damage are applied simultaneously to each affected joint: reduced motor force and limited linear velocity. `Undetectable' means these damages are undetectable to the damage detection module, so no masking is applied. Moreover, in such instances, since the damage is not detected, we assume the sensors to be capable of transmitting accurate data, or otherwise, anomalies or damage could be inferred from the sensor input given the maturity of the existing damage detection methods~\cite{quamar2024reviewfaultdiagnosisfaulttolerant}.
    
\end{itemize}

\section{Malfunction and Experiment Settings}

In this section, we use detailed statistics and \cref{fig:malfunction_demonstration} to illustrate our damage design further. We conducted three sets of tests for each task with different damage conditions. The training and inference parameters are provided in \cref{tab:a1_mal}, \cref{tab:h1_mal}, \cref{tab:g1_mal} and \cref{tab:solo8}, where:

\textit{Malfunction Timing} refers to the specific episode we apply malfunctions to the robots. \textit{ROM Limit} indicates the range of motion for each joint in the environment is restricted to a certain percentage of its original range. \textit{Motor Limit} specifies that the motor strength for each joint is capped at a certain value. \textit{Velocity Limit} means the maximum speed of joint movement is a certain value.
\textit{Random Damage Range} denotes the number of randomly selected joints damaged in each environment. \textit{Random Malfunction Seed} refers to the seed we use when randomly selecting which joints to be damaged for each environment. \textit{Perlin Noise Seed}, \textit{Track Width}, \textit{Border Size} and \textit{Track Block Length} emphasize that we test our methods on different terrain settings in \cref{tab:a1_mal}.

During inference, each damage scenario is tested separately. Also, in each scenario, the malfunction limits (ROM, Motor and Velocity) are applied to the joints under three malfunction setting groups (the timing to add malfunction, different damage range, etc.). This approach ensures that the robot's limbs encounter a wide range of states, enhancing the robustness and rigour of the process. The rationale is that the difficulty of overcoming obstacles and completing tasks significantly depends on the robot's posture. For example, a malfunction occurring when a limb is fully extended to support the robot's weight presents a greater challenge compared to when the limb is retracted during a recovery phase. Therefore, we eventually set up various inference groups with different damage settings to generate as many postures as possible.

\section{More Experiment Results}

\begin{figure*}[ht]
  \centering
  \begin{subfigure}{0.49\linewidth}
    \centering
    \includegraphics[width=\linewidth]{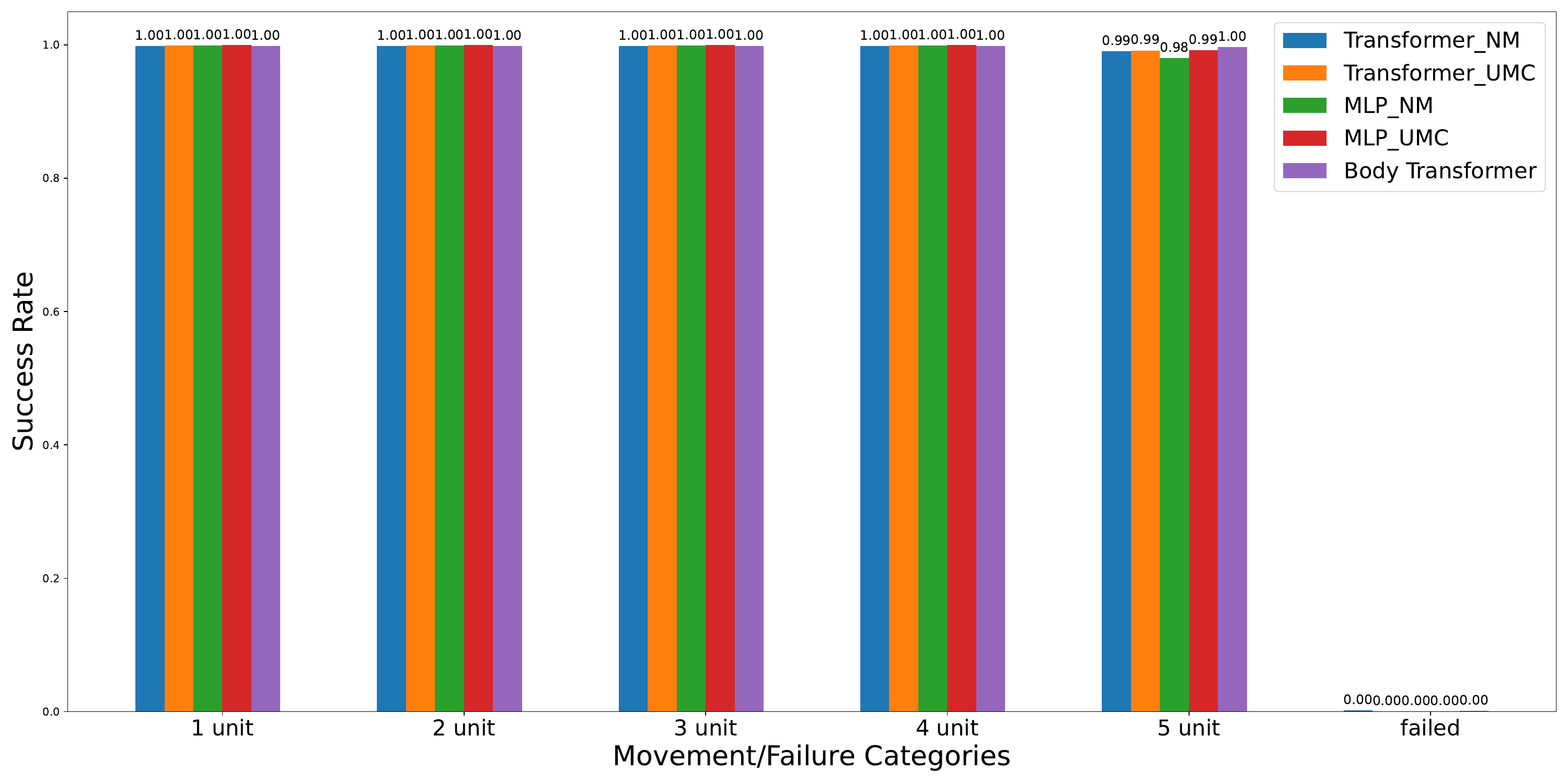}
    \caption{Normal Condition}
    \label{fig:a1_norm}
  \end{subfigure}
  \hfill
  \begin{subfigure}{0.49\linewidth}
    \centering
    \includegraphics[width=\linewidth]{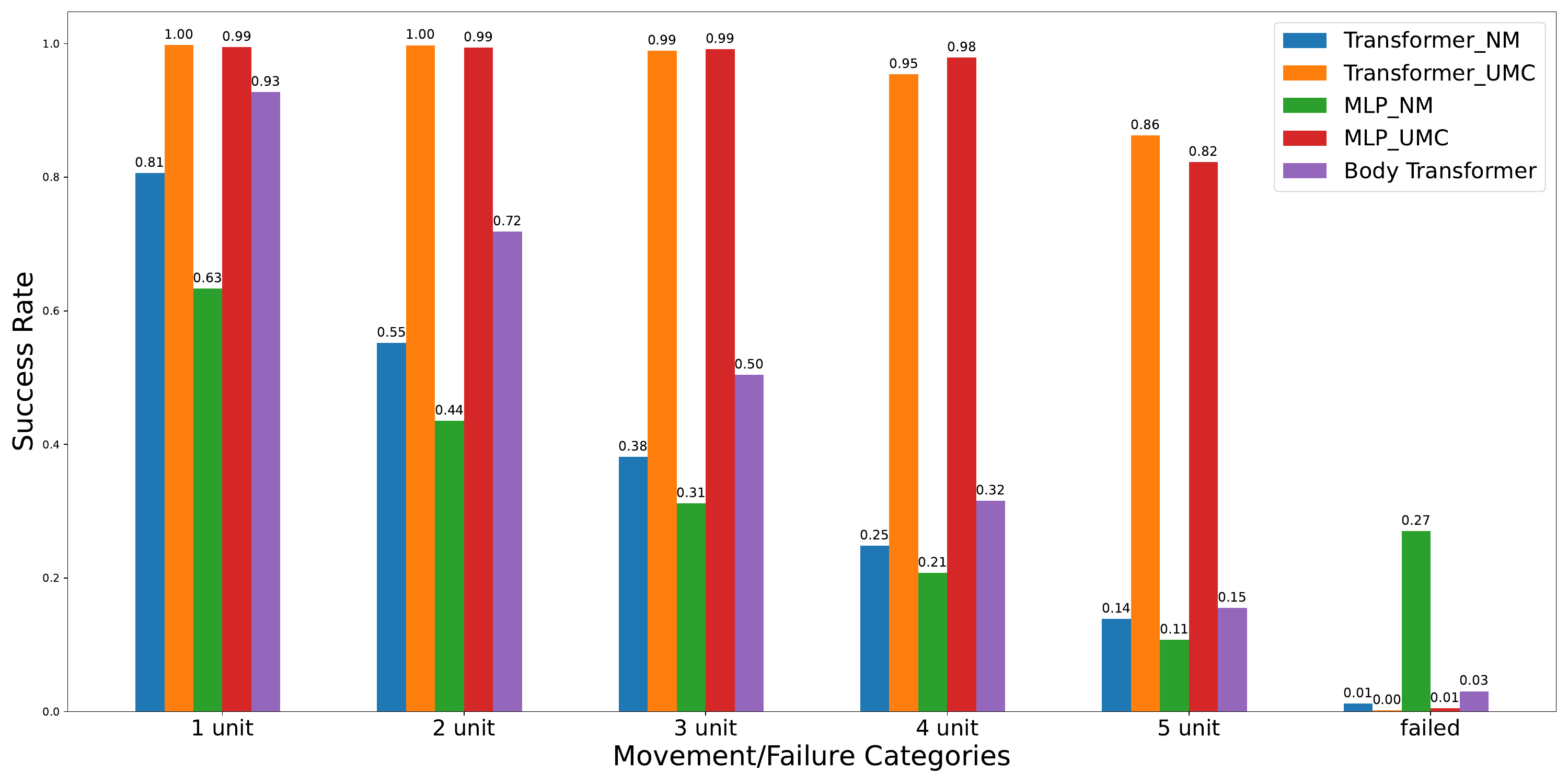}
    \caption{Sensor-Damaged Condition}
    \label{fig:a1_obslimit}
  \end{subfigure}
  \hfill
  \begin{subfigure}{0.49\linewidth}
    \centering
    \includegraphics[width=\linewidth]{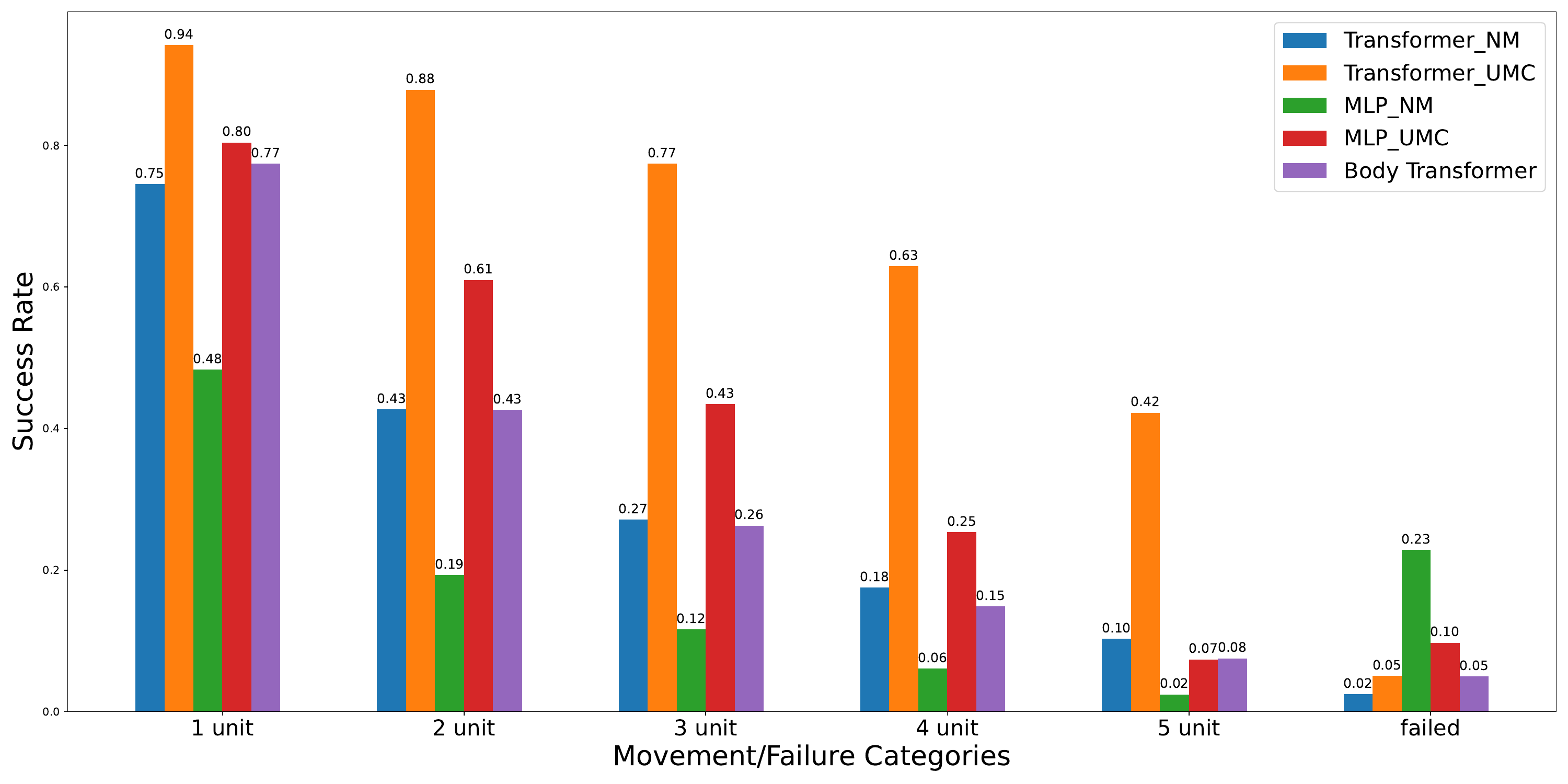}
    \caption{Detected ROM-Limit Condition}
    \label{fig:a1_rom_det}
  \end{subfigure}
  \hfill
  \begin{subfigure}{0.49\linewidth}
    \centering
    \includegraphics[width=\linewidth]{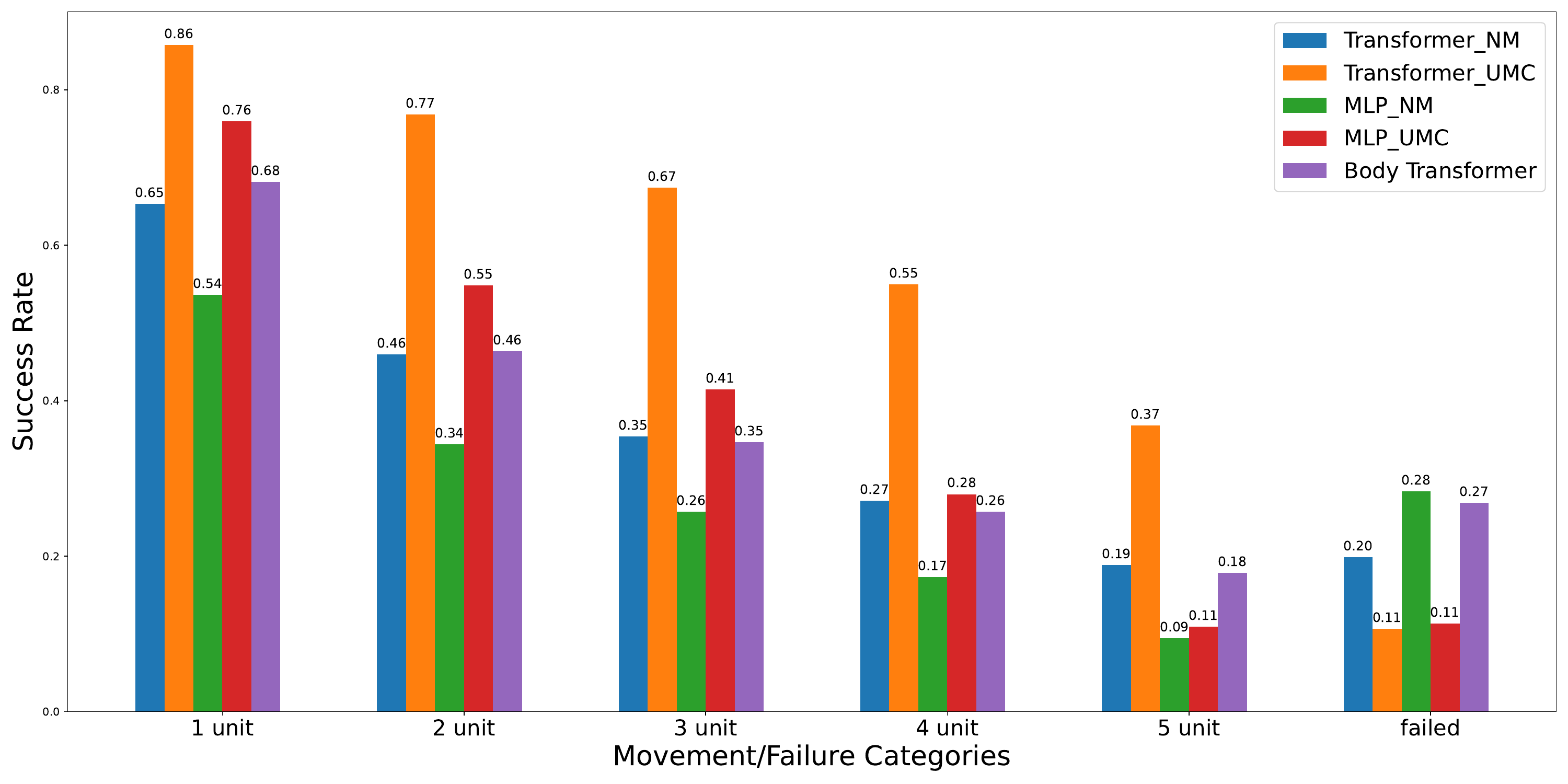}
    \caption{Undetected ROM-Limit Condition}
    \label{fig:a1_rom_und}
  \end{subfigure}
  \hfill
  \begin{subfigure}{0.49\linewidth}
    \centering
    \includegraphics[width=\linewidth]{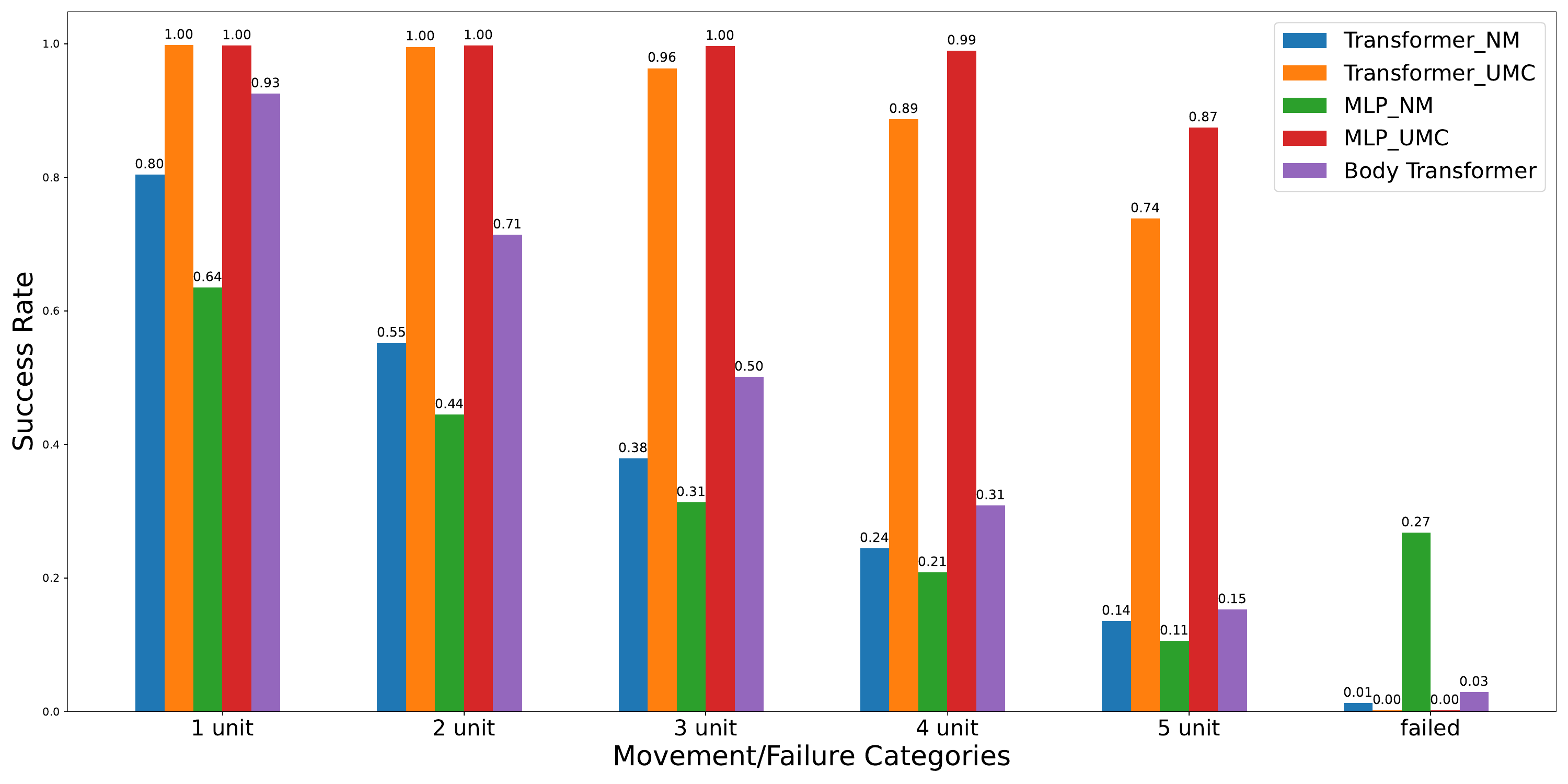}
    \caption{Detected Motor-Limit Condition}
    \label{fig:a1_mot_det}
  \end{subfigure}
  \hfill
  \begin{subfigure}{0.49\linewidth}
    \centering
    \includegraphics[width=\linewidth]{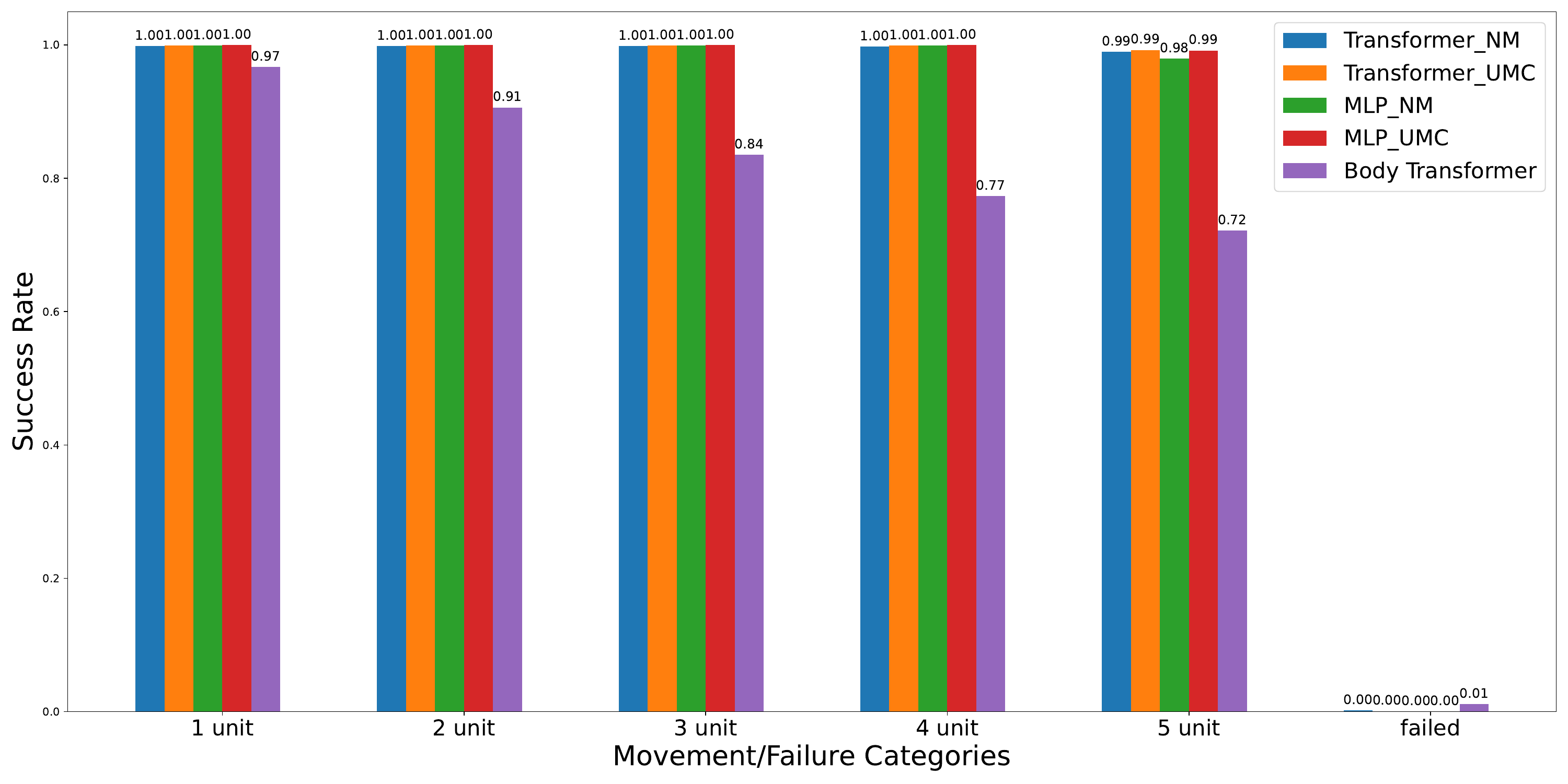}
    \caption{Undetected Motor-Limit Condition}
    \label{fig:a1_mot_und}
  \end{subfigure}
  \hfill
  \begin{subfigure}{0.49\linewidth}
    \centering
    \includegraphics[width=\linewidth]{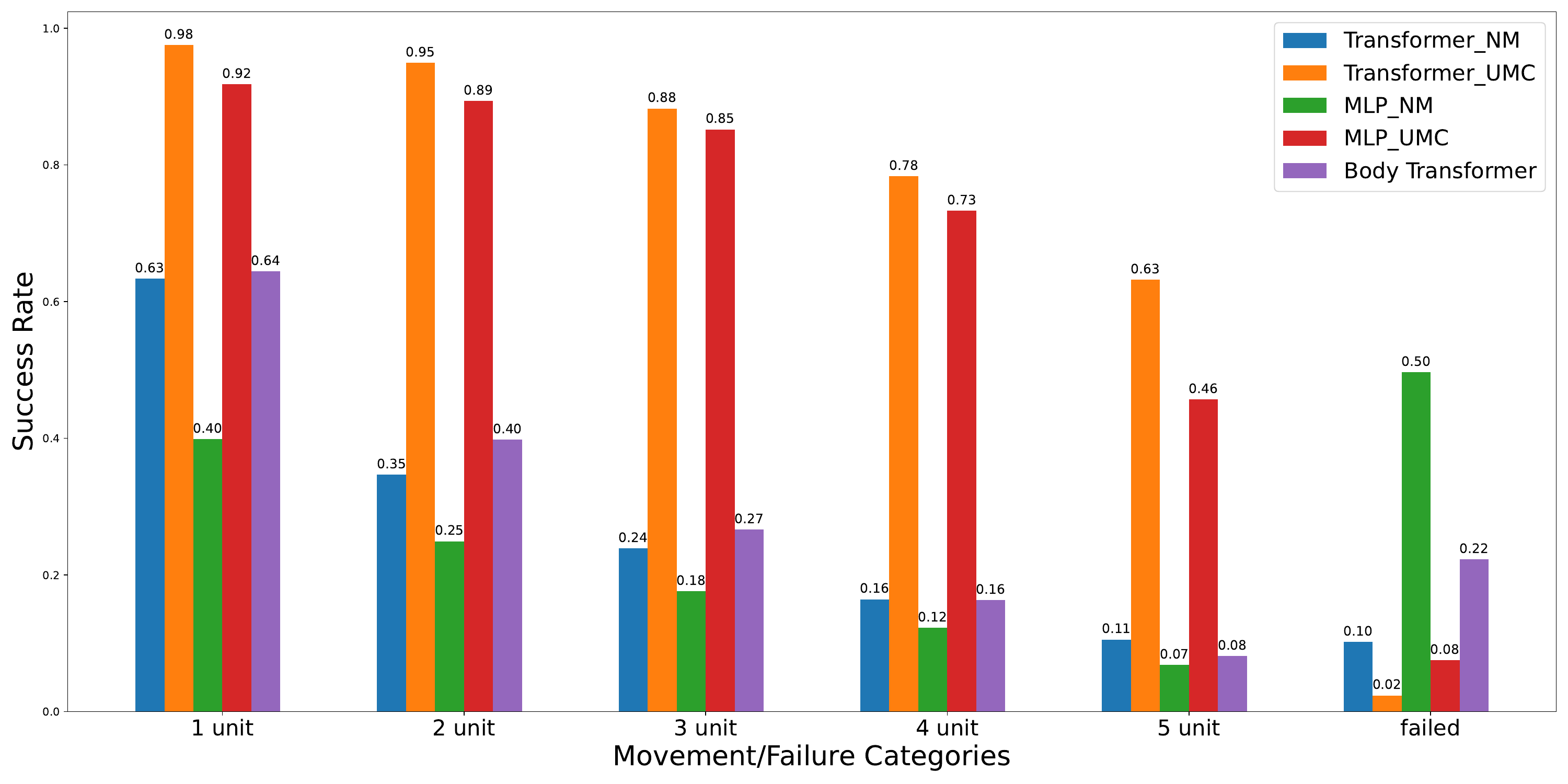}
    \caption{Detected Velocity-Limit Condition}
    \label{fig:a1_vel_det}
  \end{subfigure}
  \hfill
  \begin{subfigure}{0.49\linewidth}
    \centering
    \includegraphics[width=\linewidth]{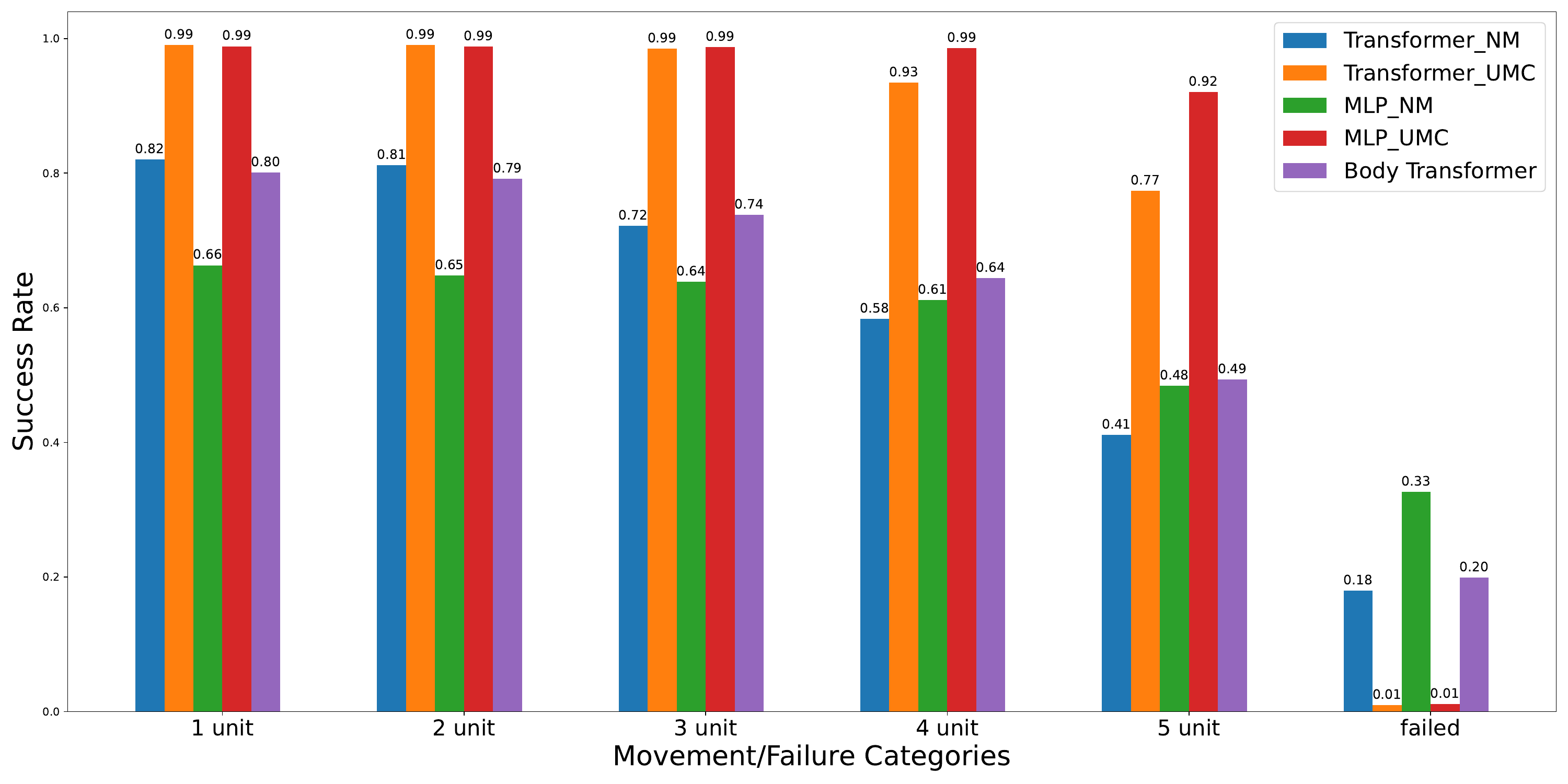}
    \caption{Undetected Velocity-Limit Condition}
    \label{fig:a1_vel_und}
  \end{subfigure}
  \hfill
  \caption{Performance of Five Methods Under Different Damage Conditions in the A1-Walk Task.}
  \label{fig:a1_detail_results}
\end{figure*}

\begin{figure*}[ht]
  \centering
  \begin{subfigure}{0.49\linewidth}
    \centering
    \includegraphics[width=\linewidth]{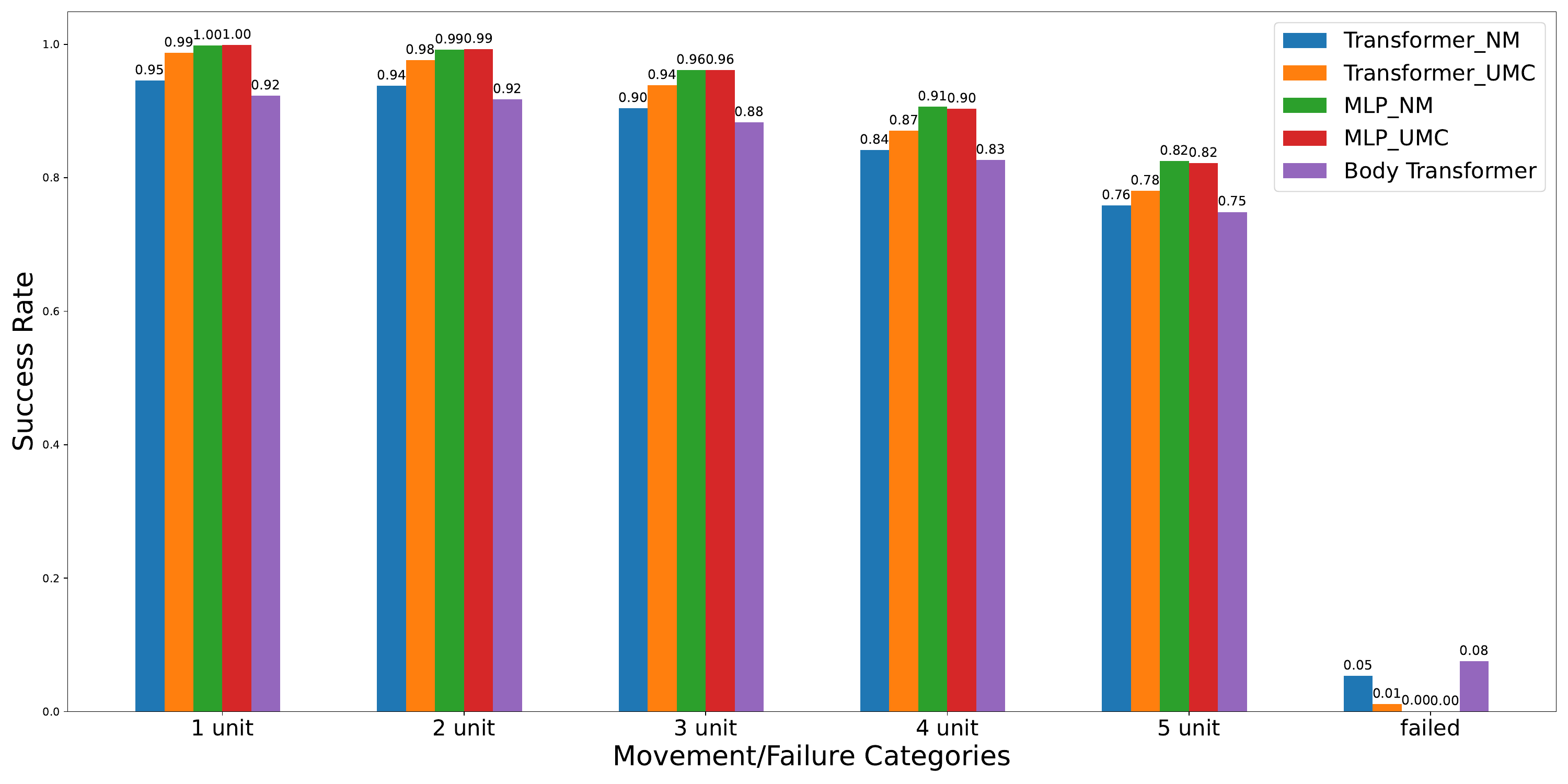}
    \caption{Normal Condition}
    \label{fig:h1_norm}
  \end{subfigure}
  \hfill
  \begin{subfigure}{0.49\linewidth}
    \centering
    \includegraphics[width=\linewidth]{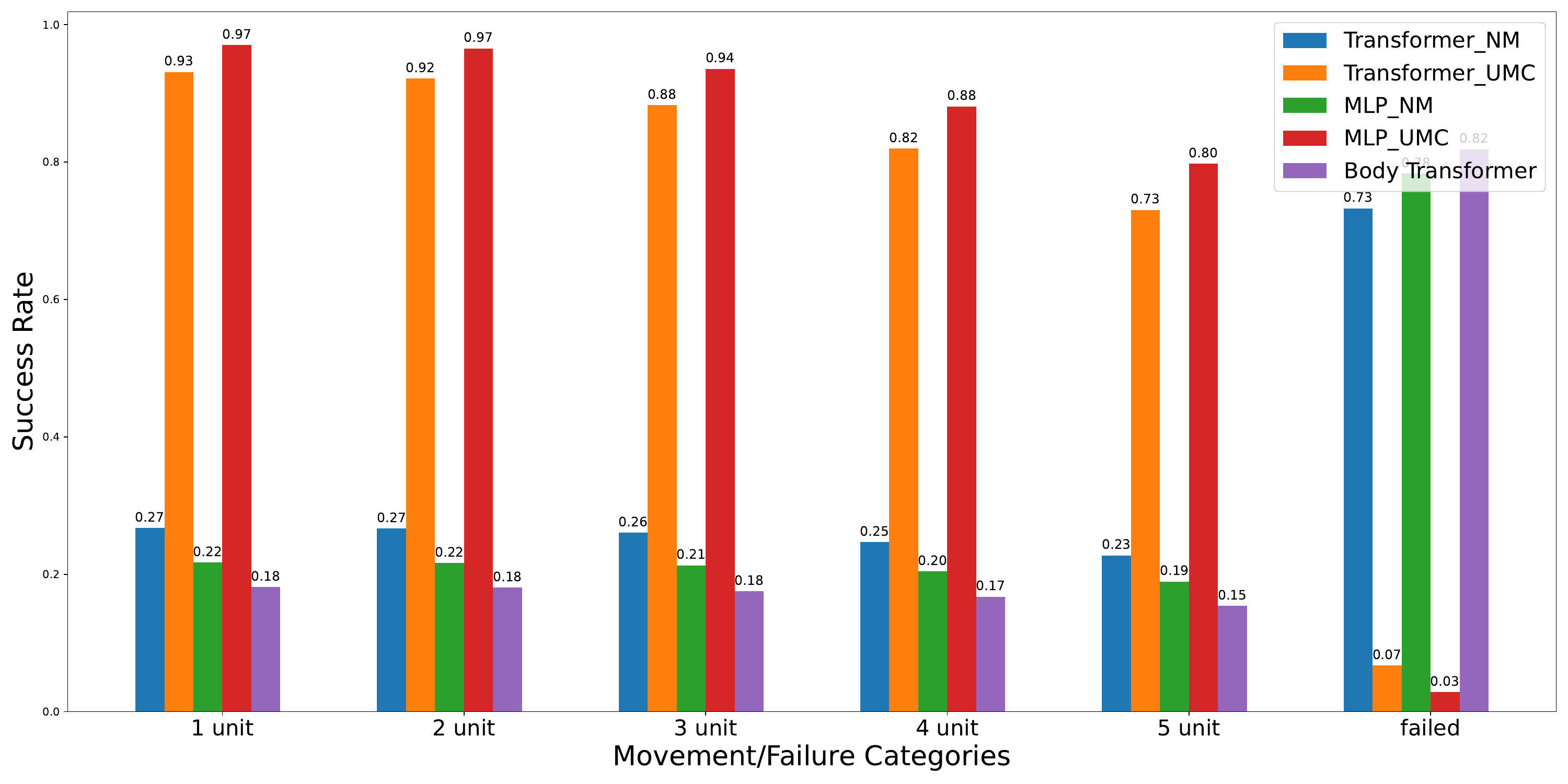}
    \caption{Sensor-Damaged Condition}
    \label{fig:h1_obslimit}
  \end{subfigure}
  \hfill
  \begin{subfigure}{0.49\linewidth}
    \centering
    \includegraphics[width=\linewidth]{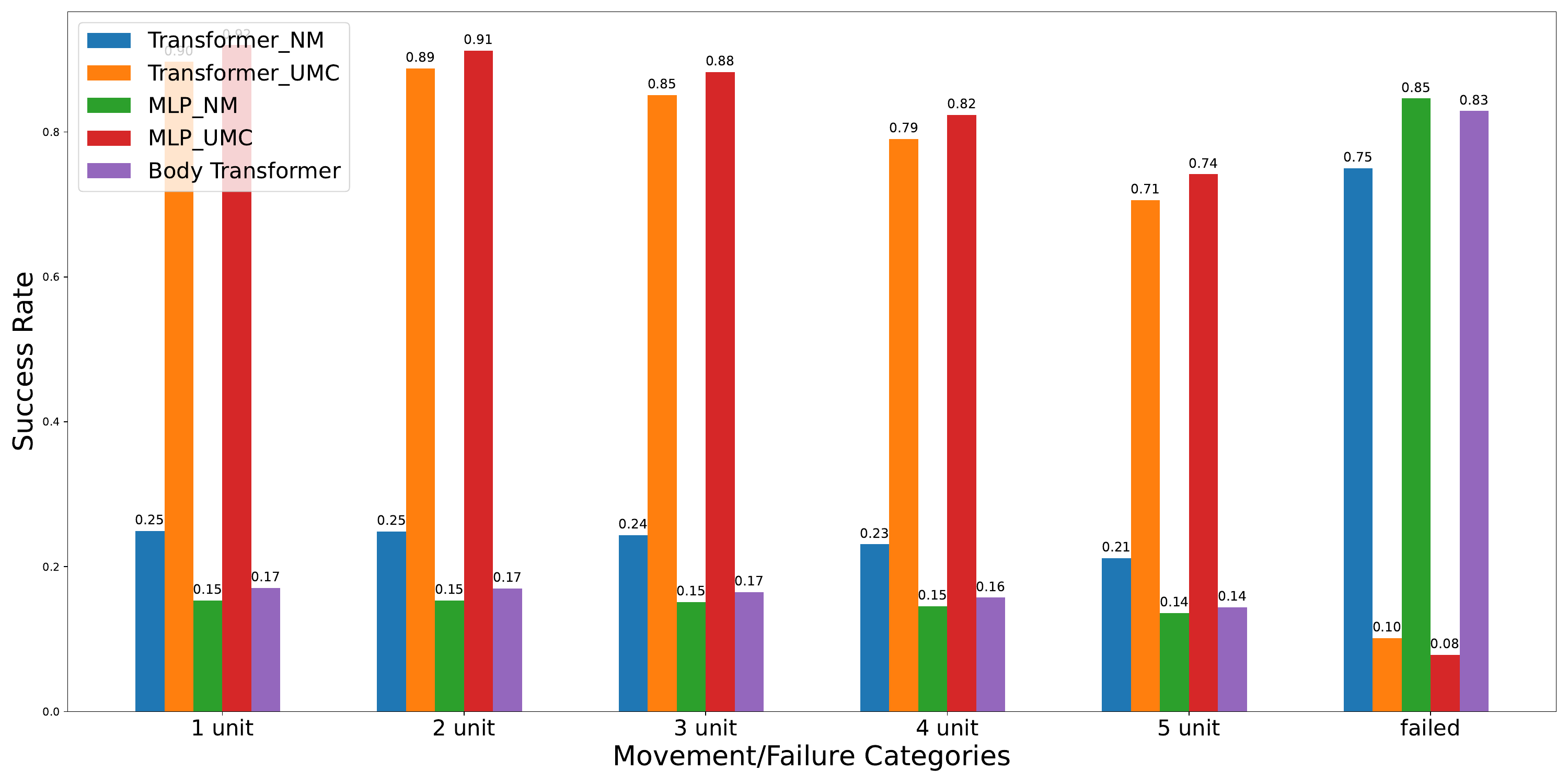}
    \caption{Detected ROM-Limit Condition}
    \label{fig:h1_rom_det}
  \end{subfigure}
  \hfill
  \begin{subfigure}{0.49\linewidth}
    \centering
    \includegraphics[width=\linewidth]{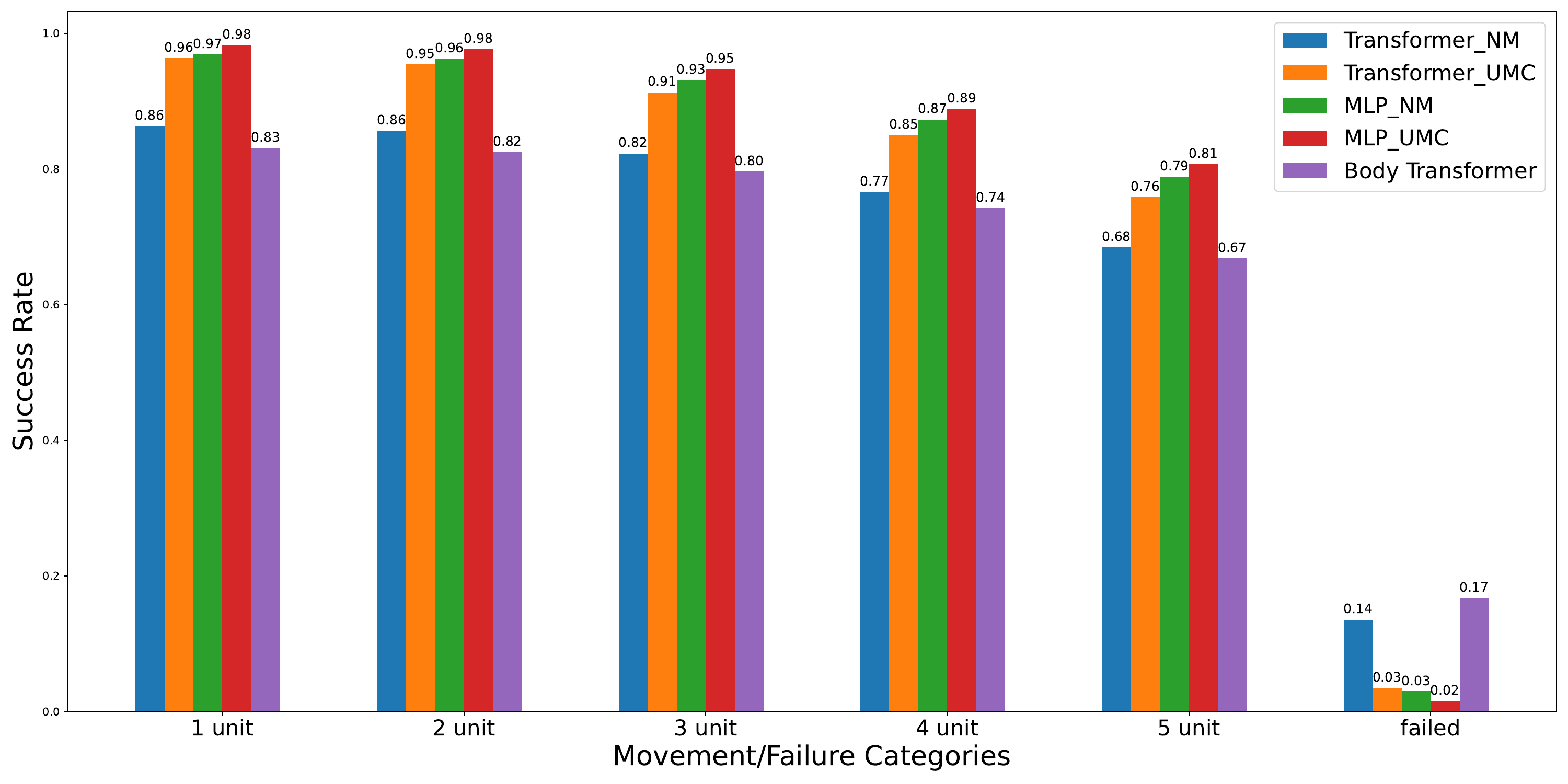}
    \caption{Undetected ROM-Limit Condition}
    \label{fig:h1_rom_und}
  \end{subfigure}
  \hfill
  \begin{subfigure}{0.49\linewidth}
    \centering
    \includegraphics[width=\linewidth]{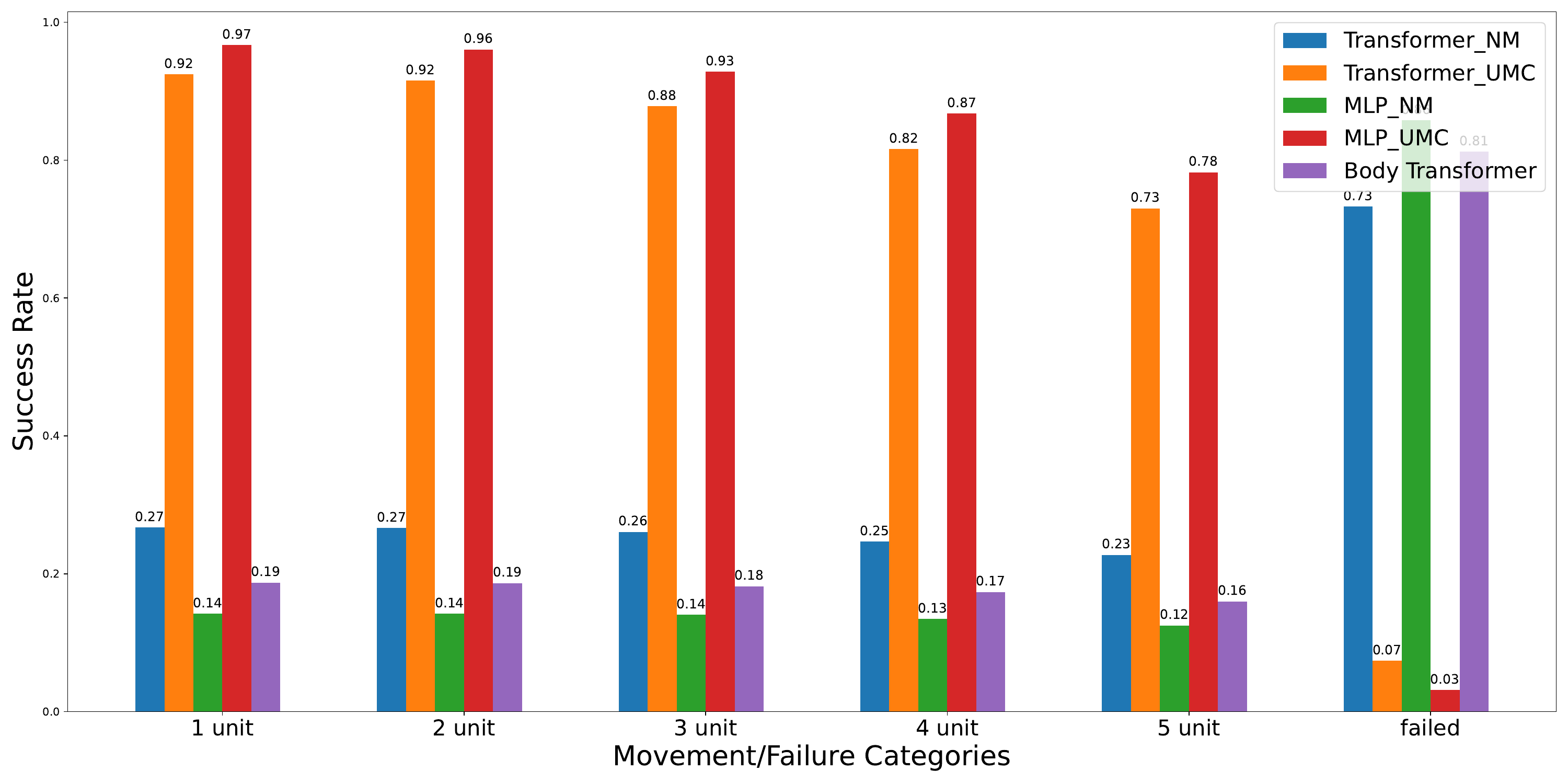}
    \caption{Detected Motor-Limit Condition}
    \label{fig:h1_mot_det}
  \end{subfigure}
  \hfill
  \begin{subfigure}{0.49\linewidth}
    \centering
    \includegraphics[width=\linewidth]{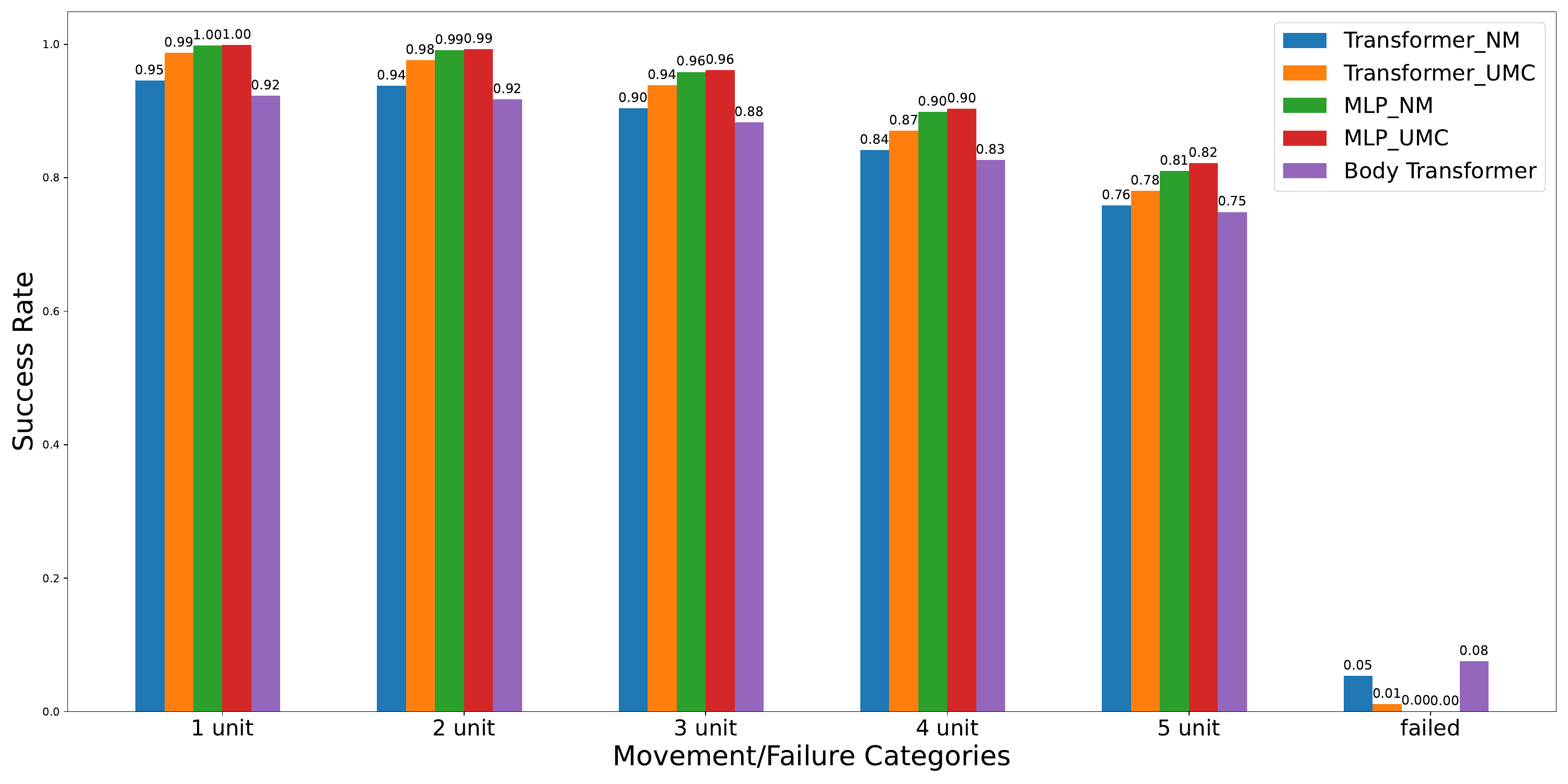}
    \caption{Undetected Motor-Limit Condition}
    \label{fig:h1_mot_und}
  \end{subfigure}
  \hfill
  \begin{subfigure}{0.49\linewidth}
    \centering
    \includegraphics[width=\linewidth]{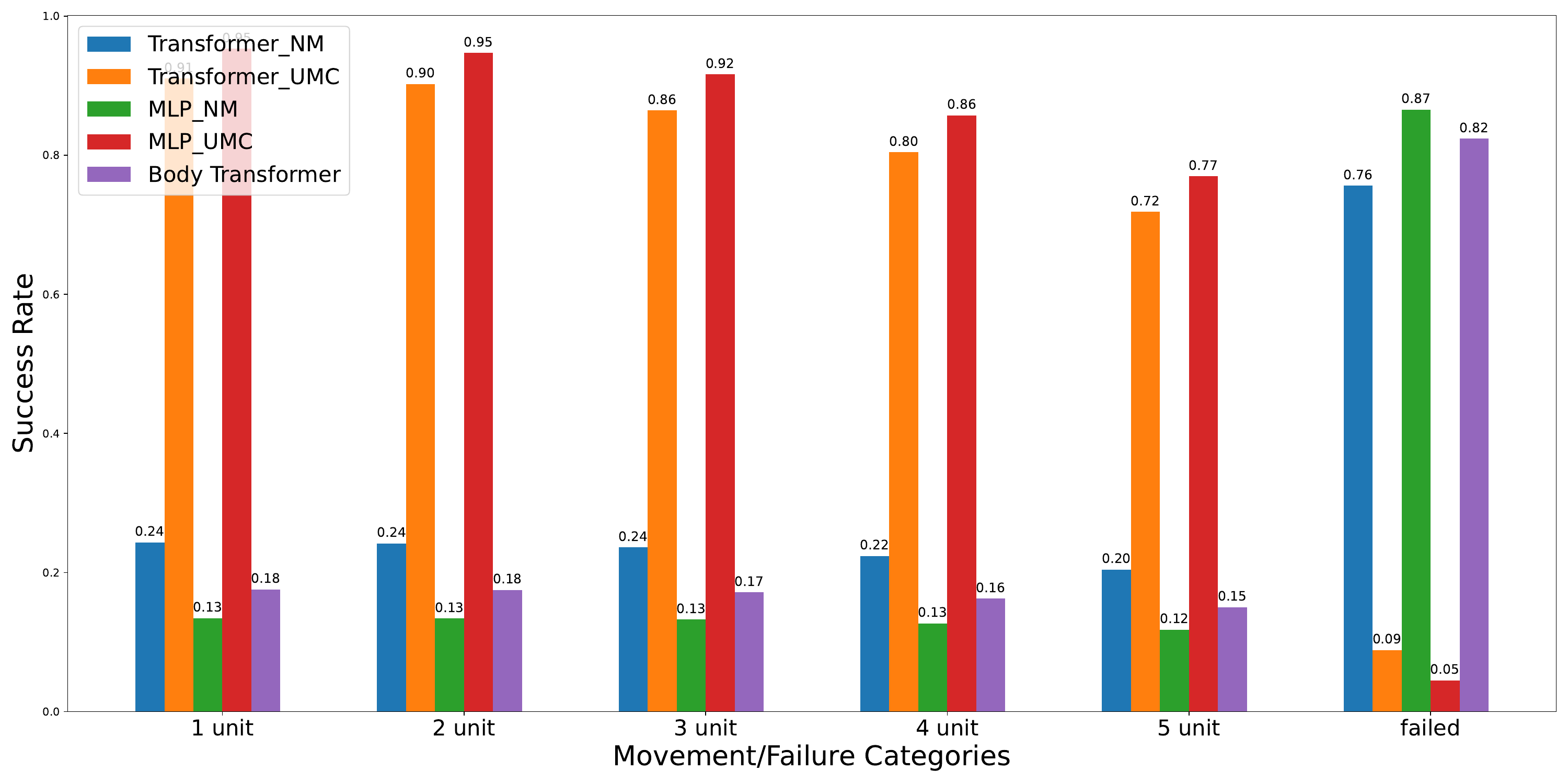}
    \caption{Detected Velocity-Limit Condition}
    \label{fig:h1_vel_det}
  \end{subfigure}
  \hfill
  \begin{subfigure}{0.49\linewidth}
    \centering
    \includegraphics[width=\linewidth]{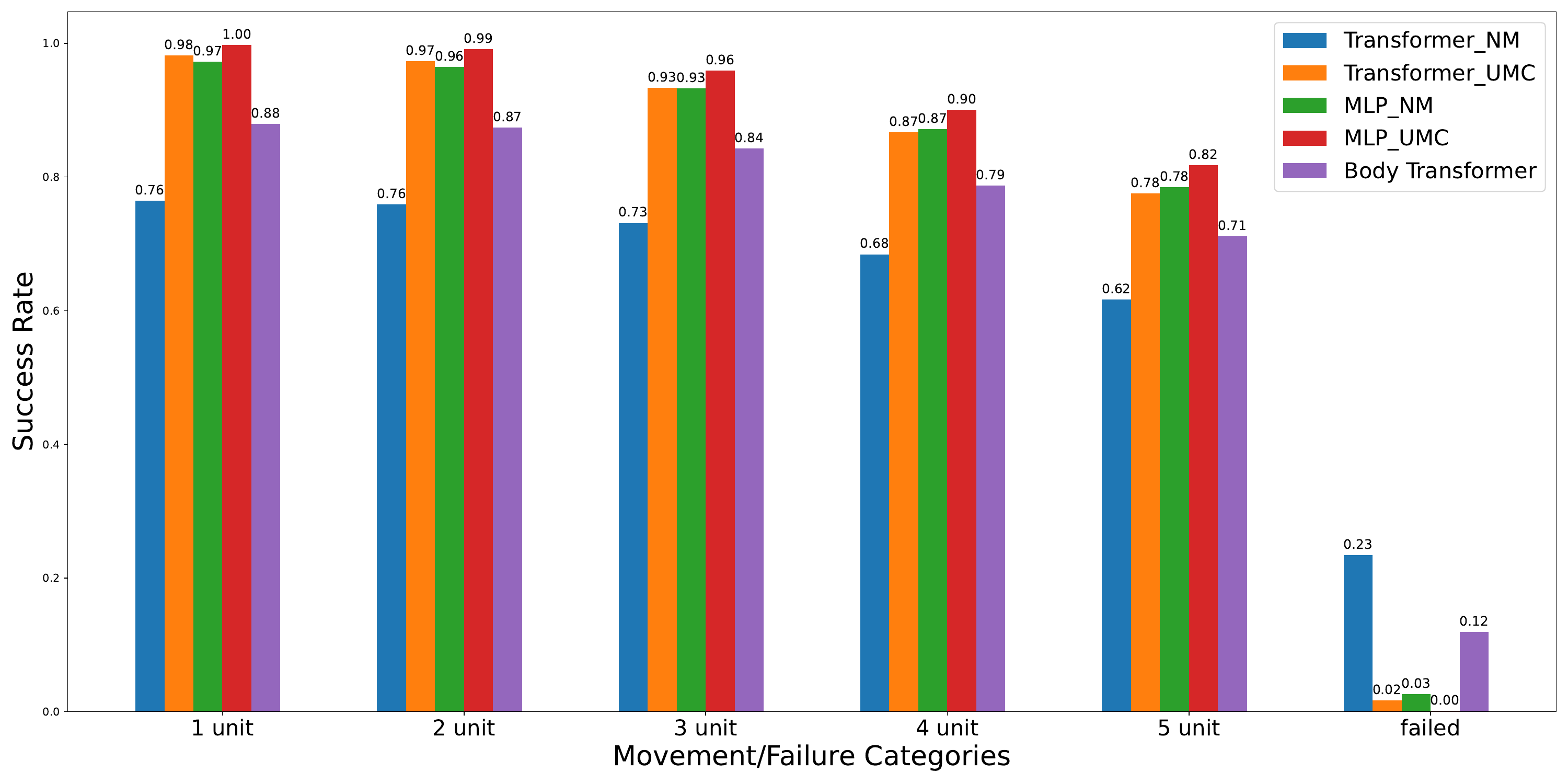}
    \caption{Undetected Velocity-Limit Condition}
    \label{fig:h1_vel_und}
  \end{subfigure}
  \hfill
  \caption{Performance of Five Methods Under Different Damage Conditions in the Unitree-H1 Task.}
  \label{fig:h1_detail_results}
\end{figure*}

\begin{figure*}[ht]
  \centering
  \begin{subfigure}{0.49\linewidth}
    \centering
    \includegraphics[width=\linewidth]{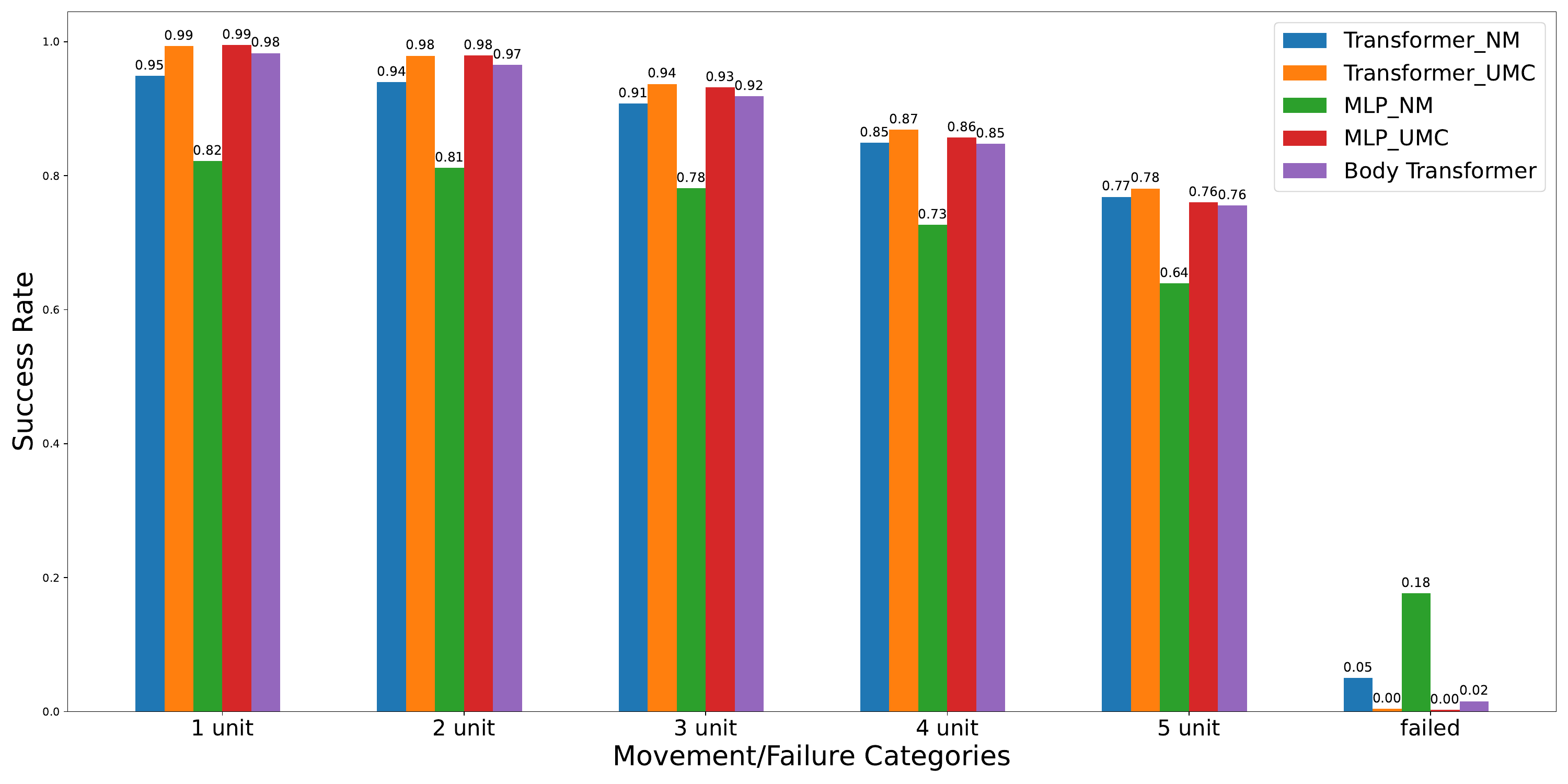}
    \caption{Normal Condition}
    \label{fig:g1_norm}
  \end{subfigure}
  \hfill
  \begin{subfigure}{0.49\linewidth}
    \centering
    \includegraphics[width=\linewidth]{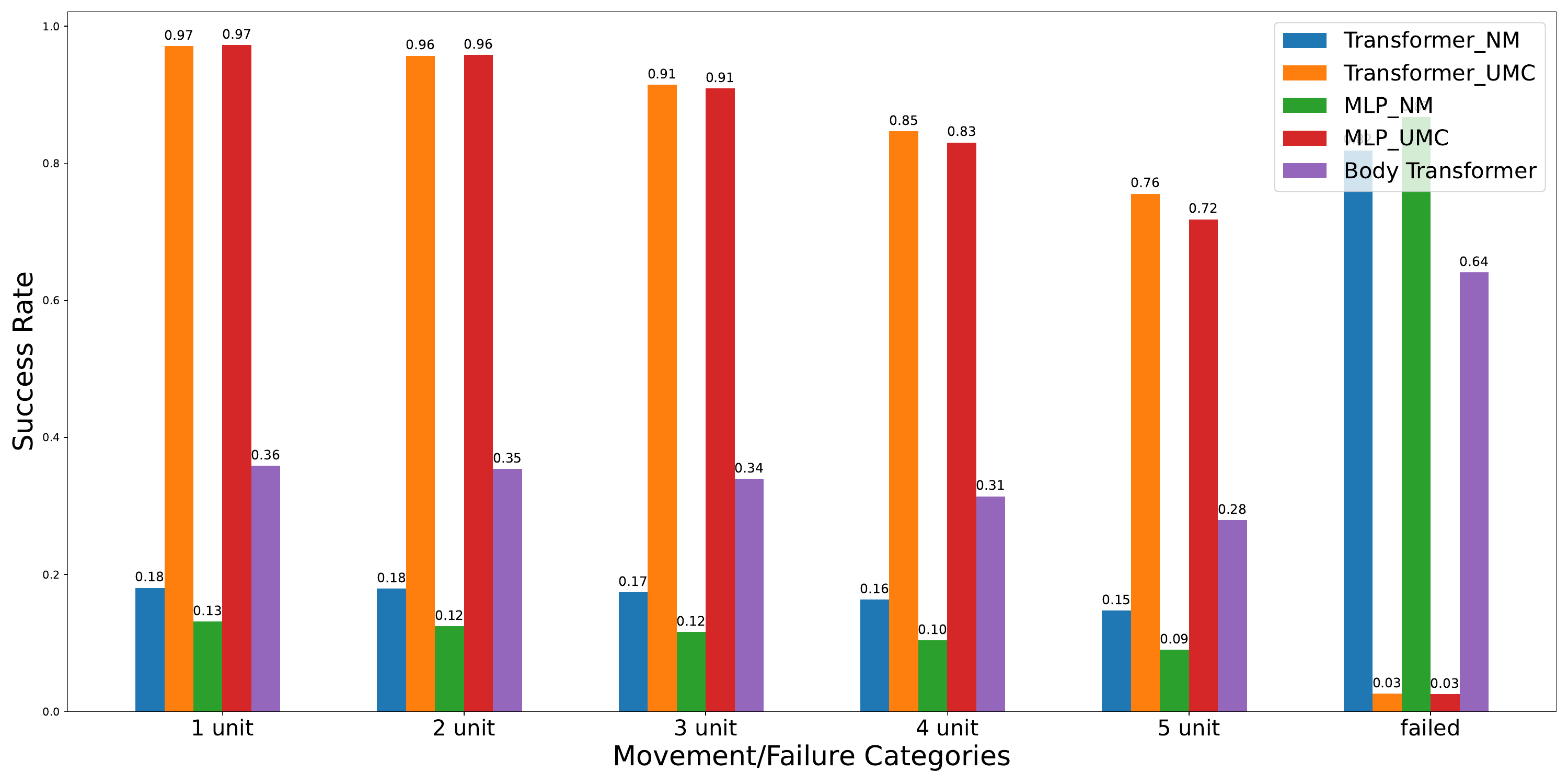}
    \caption{Sensor-Damaged Condition}
    \label{fig:g1_obslimit}
  \end{subfigure}
  \hfill
  \begin{subfigure}{0.49\linewidth}
    \centering
    \includegraphics[width=\linewidth]{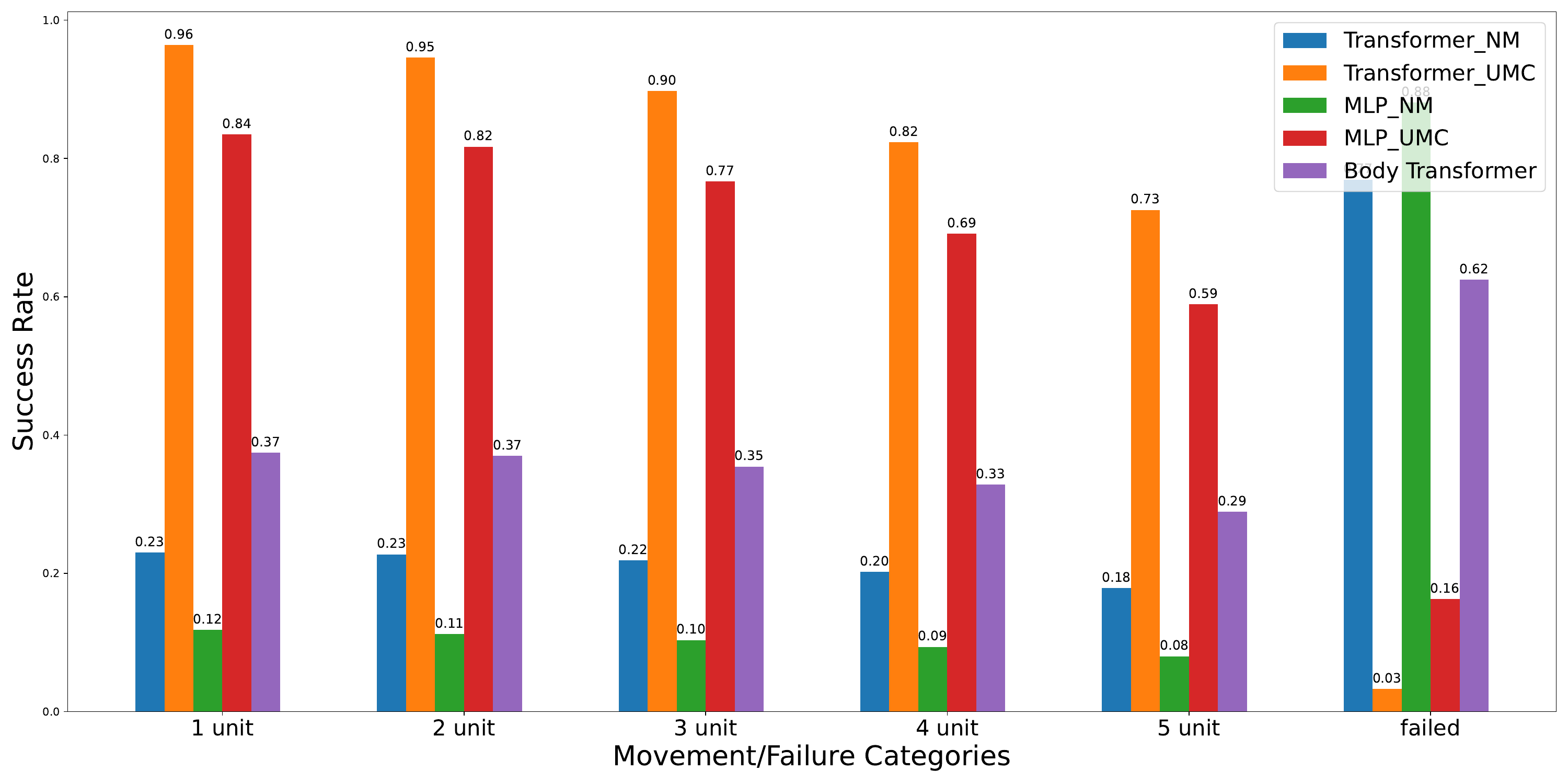}
    \caption{Detected ROM-Limit Condition}
    \label{fig:g1_rom_det}
  \end{subfigure}
  \hfill
  \begin{subfigure}{0.49\linewidth}
    \centering
    \includegraphics[width=\linewidth]{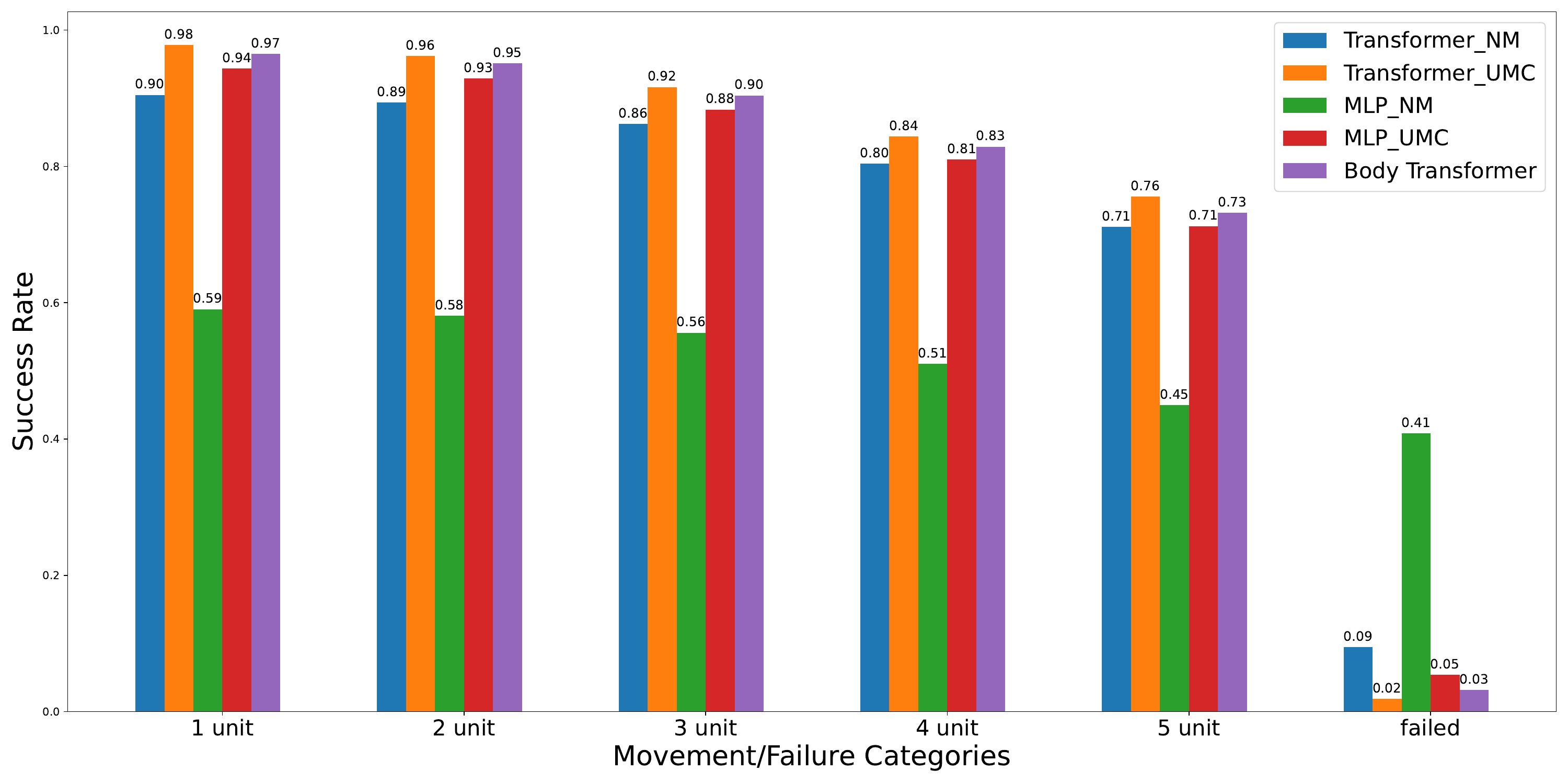}
    \caption{Undetected ROM-Limit Condition}
    \label{fig:g1_rom_und}
  \end{subfigure}
  \hfill
  \begin{subfigure}{0.49\linewidth}
    \centering
    \includegraphics[width=\linewidth]{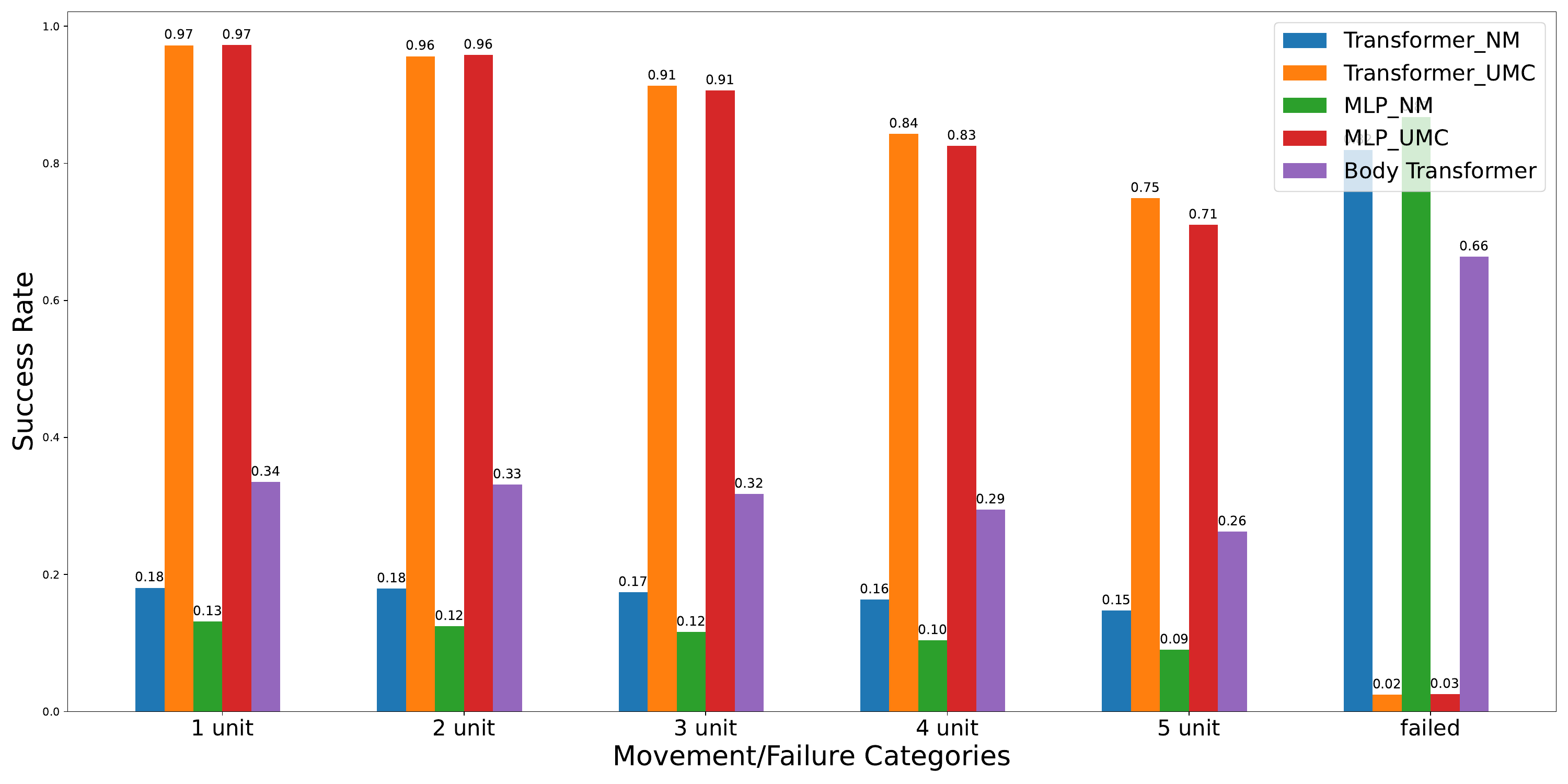}
    \caption{Detected Motor-Limit Condition}
    \label{fig:g1_mot_det}
  \end{subfigure}
  \hfill
  \begin{subfigure}{0.49\linewidth}
    \centering
    \includegraphics[width=\linewidth]{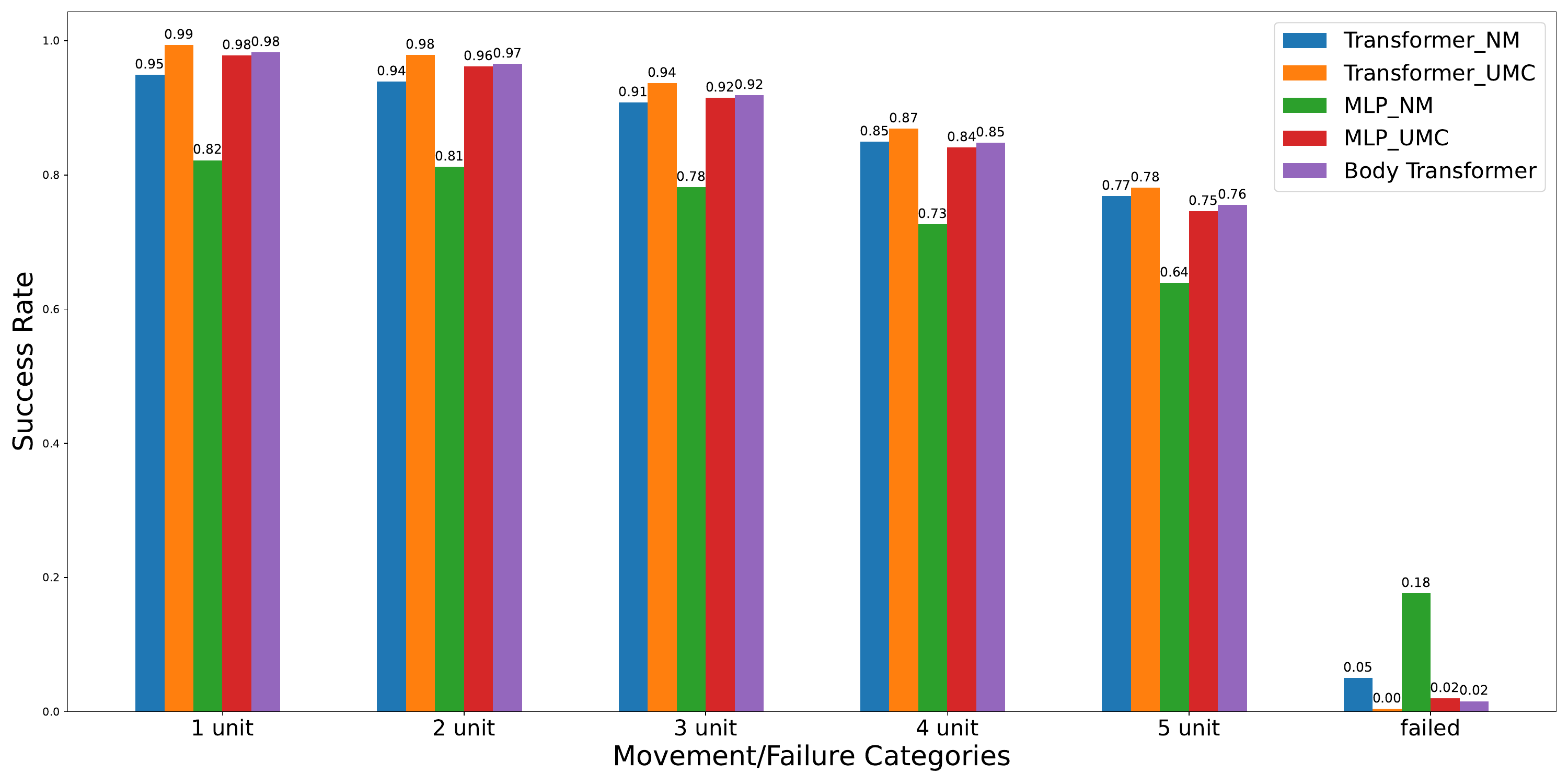}
    \caption{Undetected Motor-Limit Condition}
    \label{fig:g1_mot_und}
  \end{subfigure}
  \hfill
  \begin{subfigure}{0.49\linewidth}
    \centering
    \includegraphics[width=\linewidth]{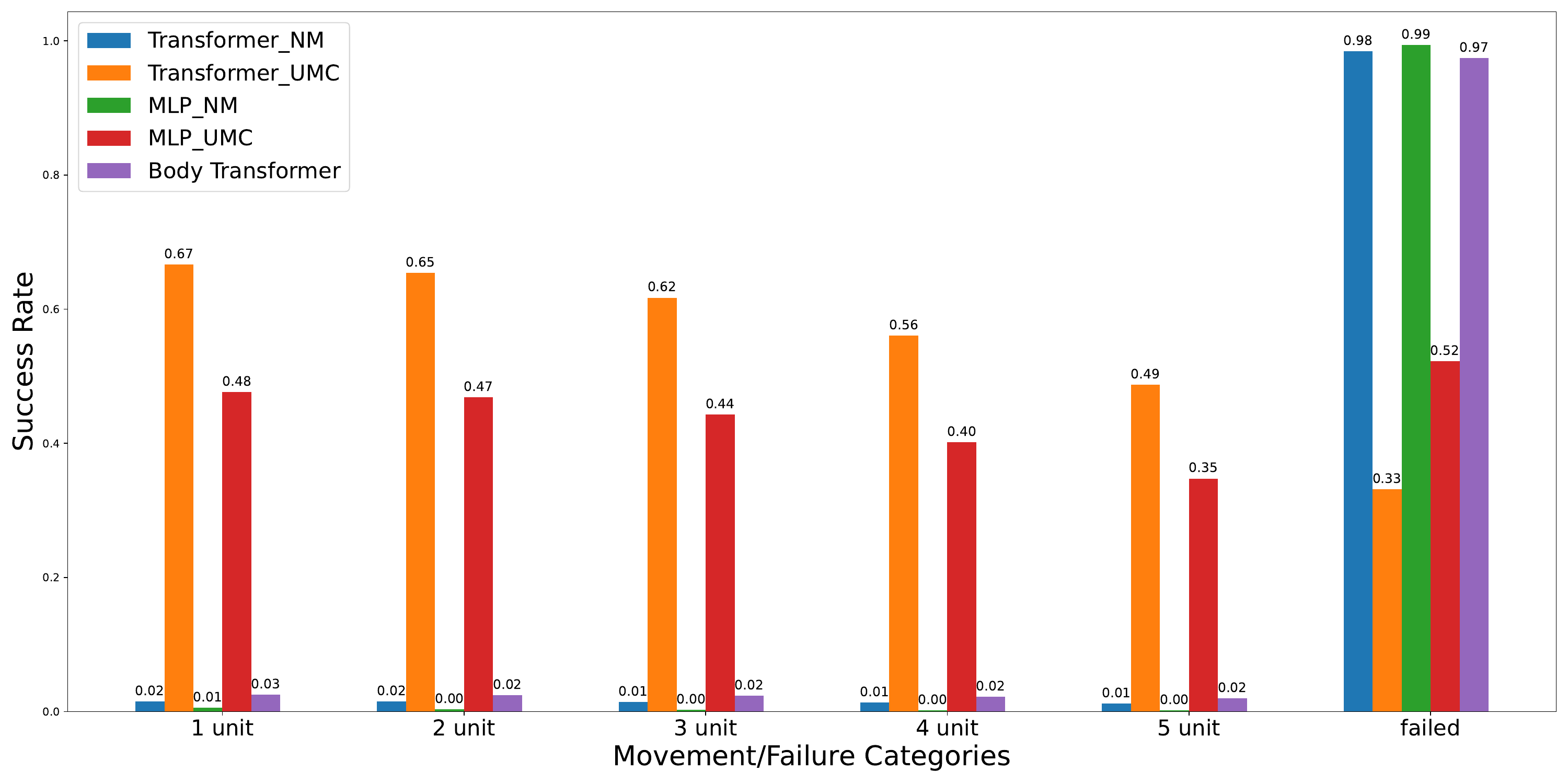}
    \caption{Detected Velocity-Limit Condition}
    \label{fig:g1_vel_det}
  \end{subfigure}
  \hfill
  \begin{subfigure}{0.49\linewidth}
    \centering
    \includegraphics[width=\linewidth]{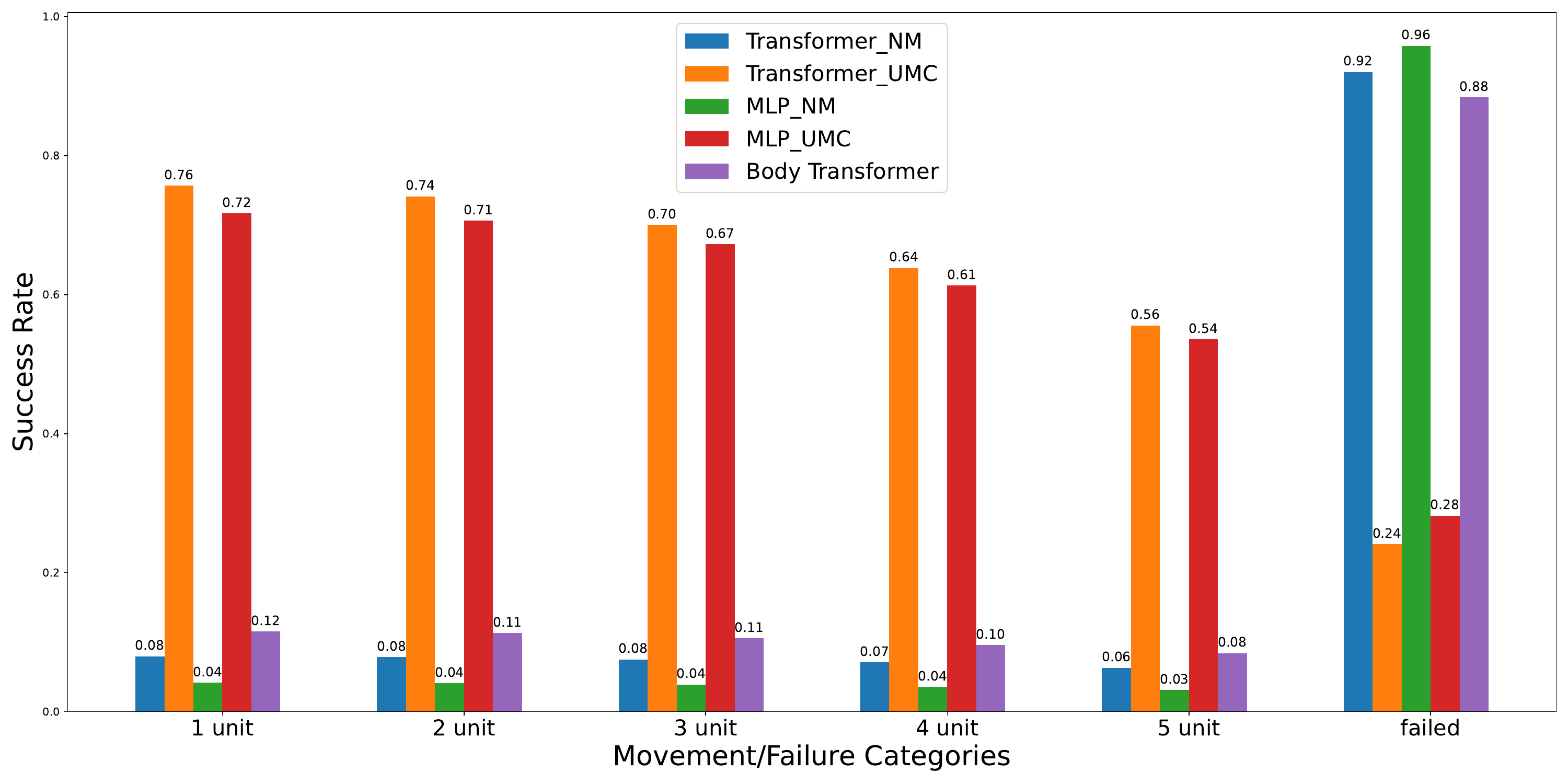}
    \caption{Undetected Velocity-Limit Condition}
    \label{fig:g1_vel_und}
  \end{subfigure}
  \hfill
  \caption{Performance of Five Methods Under Different Damage Conditions in the Unitree-G1 Task.}
  \label{fig:g1_detail_results}
\end{figure*}

\begin{figure*}[ht]
  \centering
  \begin{subfigure}{0.49\linewidth}
    \centering
    \includegraphics[width=\linewidth]{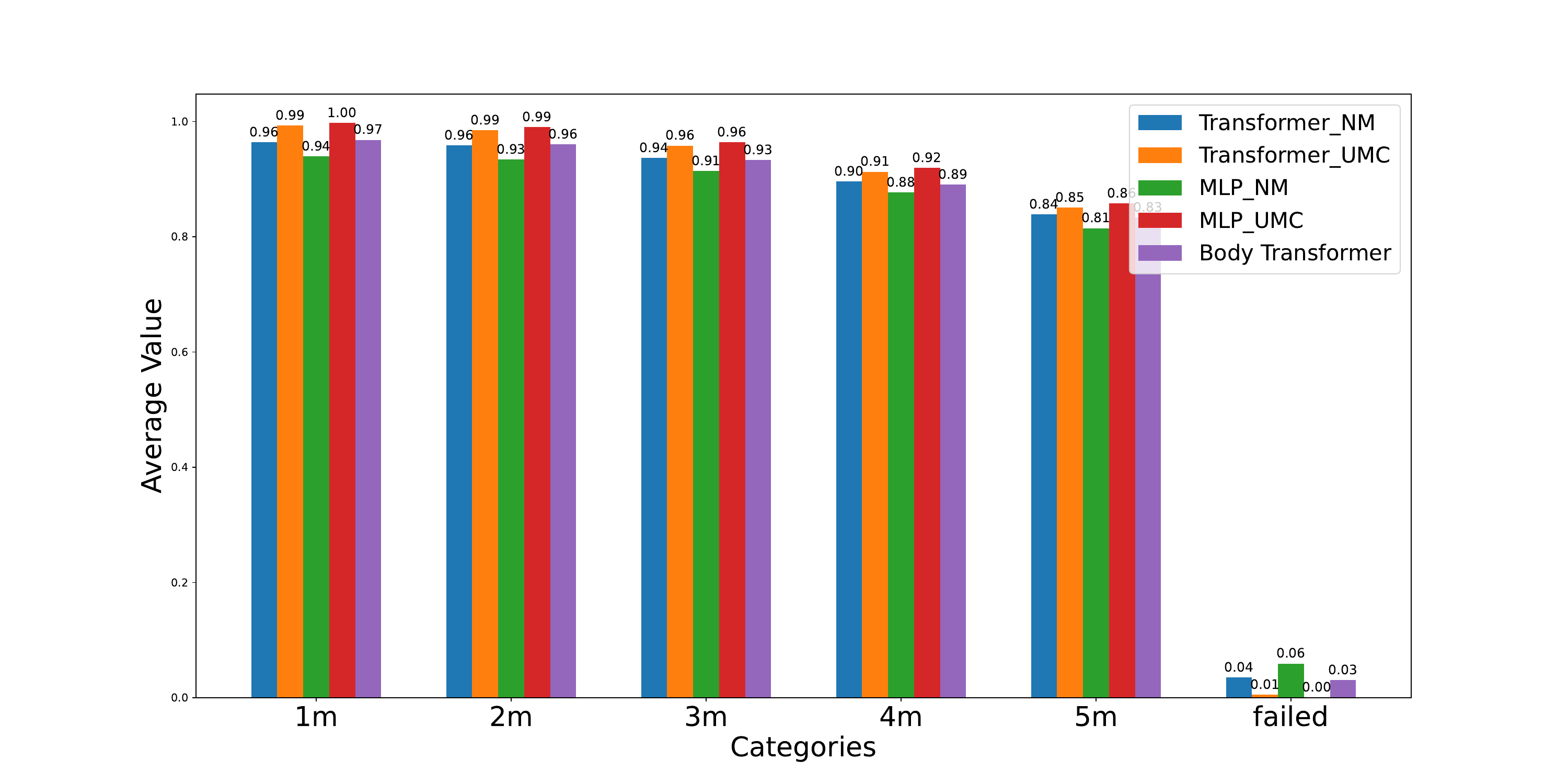}
    \caption{Normal Condition}
    \label{fig:avg_norm}
  \end{subfigure}
  \hfill
  \begin{subfigure}{0.49\linewidth}
    \centering
    \includegraphics[width=\linewidth]{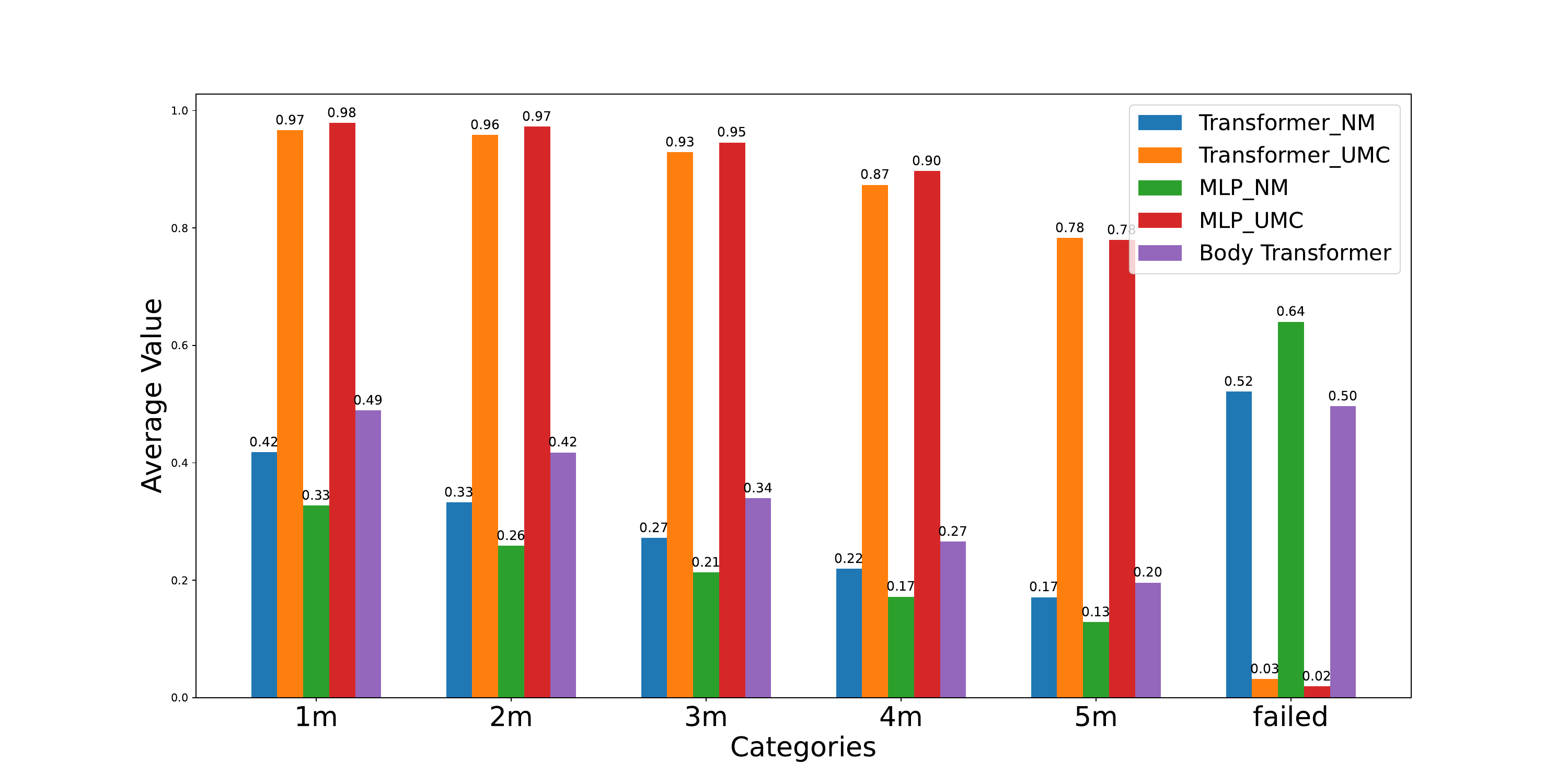}
    \caption{Sensor-Damaged Condition}
    \label{fig:avg_obslimit}
  \end{subfigure}
  \hfill
  \begin{subfigure}{0.49\linewidth}
    \centering
    \includegraphics[width=\linewidth]{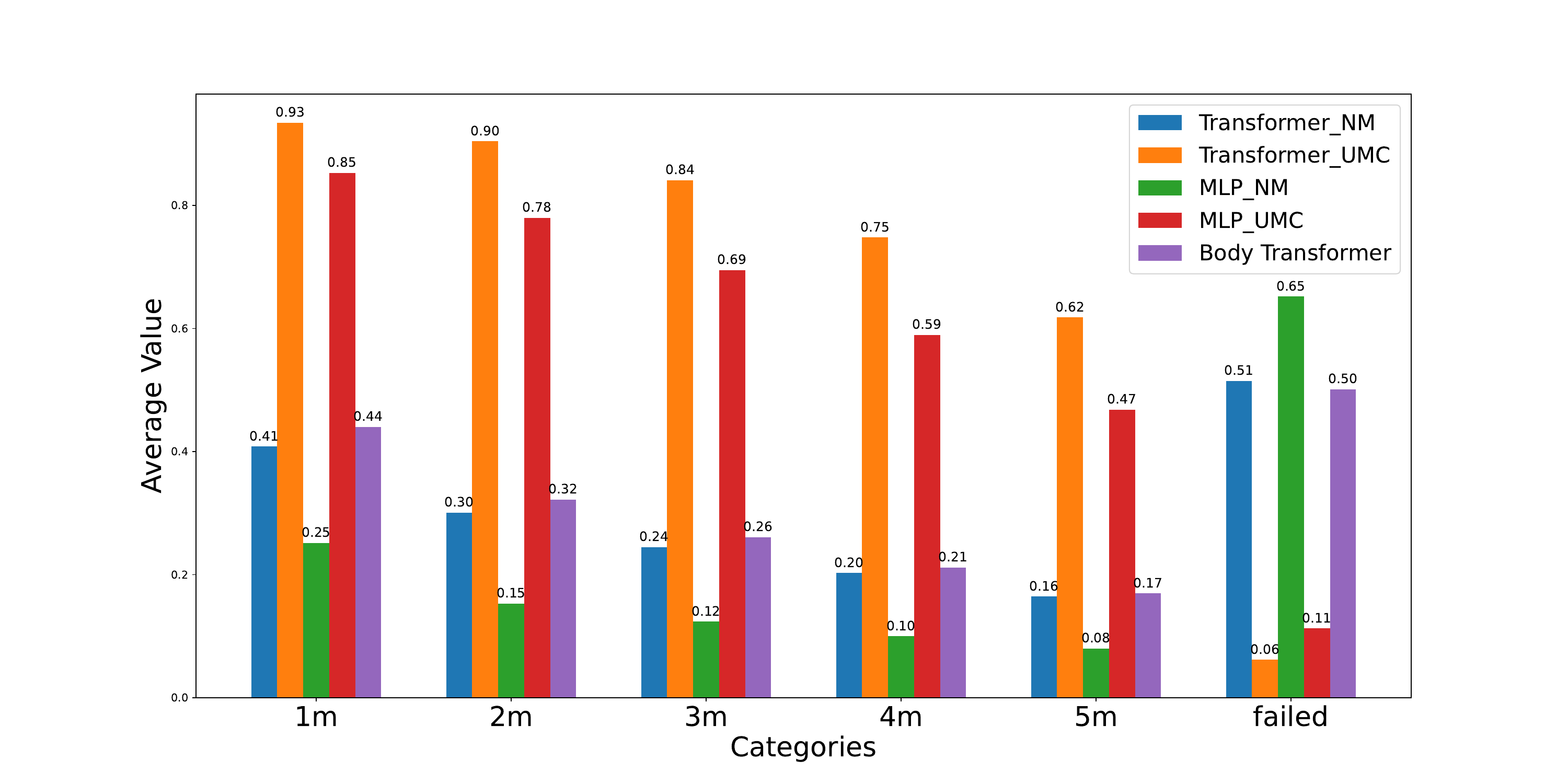}
    \caption{Detected ROM-Limit Condition}
    \label{fig:avg_rom_det}
  \end{subfigure}
  \hfill
  \begin{subfigure}{0.49\linewidth}
    \centering
    \includegraphics[width=\linewidth]{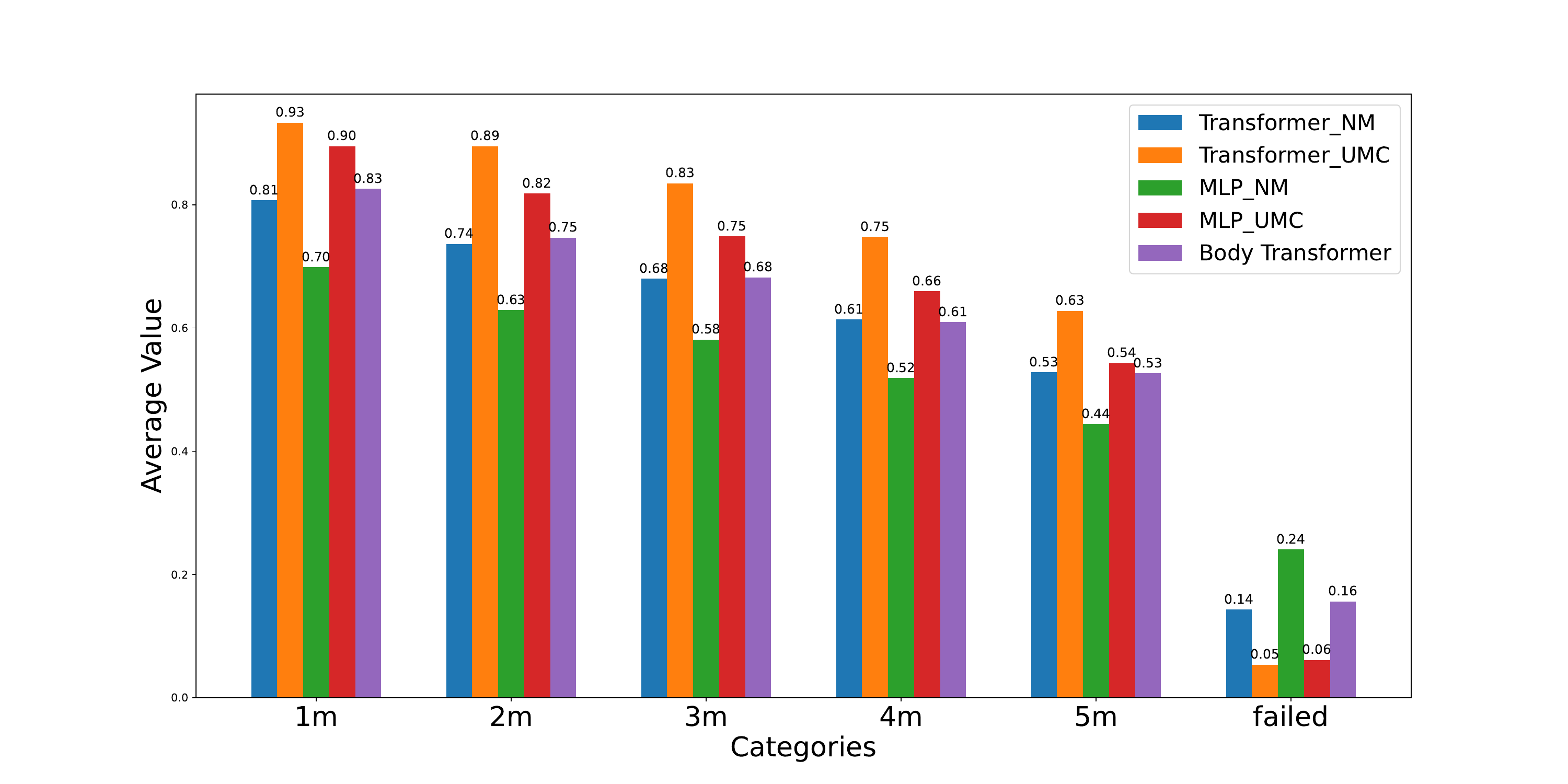}
    \caption{Undetected ROM-Limit Condition}
    \label{fig:avg_rom_und}
  \end{subfigure}
  \hfill
  \begin{subfigure}{0.49\linewidth}
    \centering
    \includegraphics[width=\linewidth]{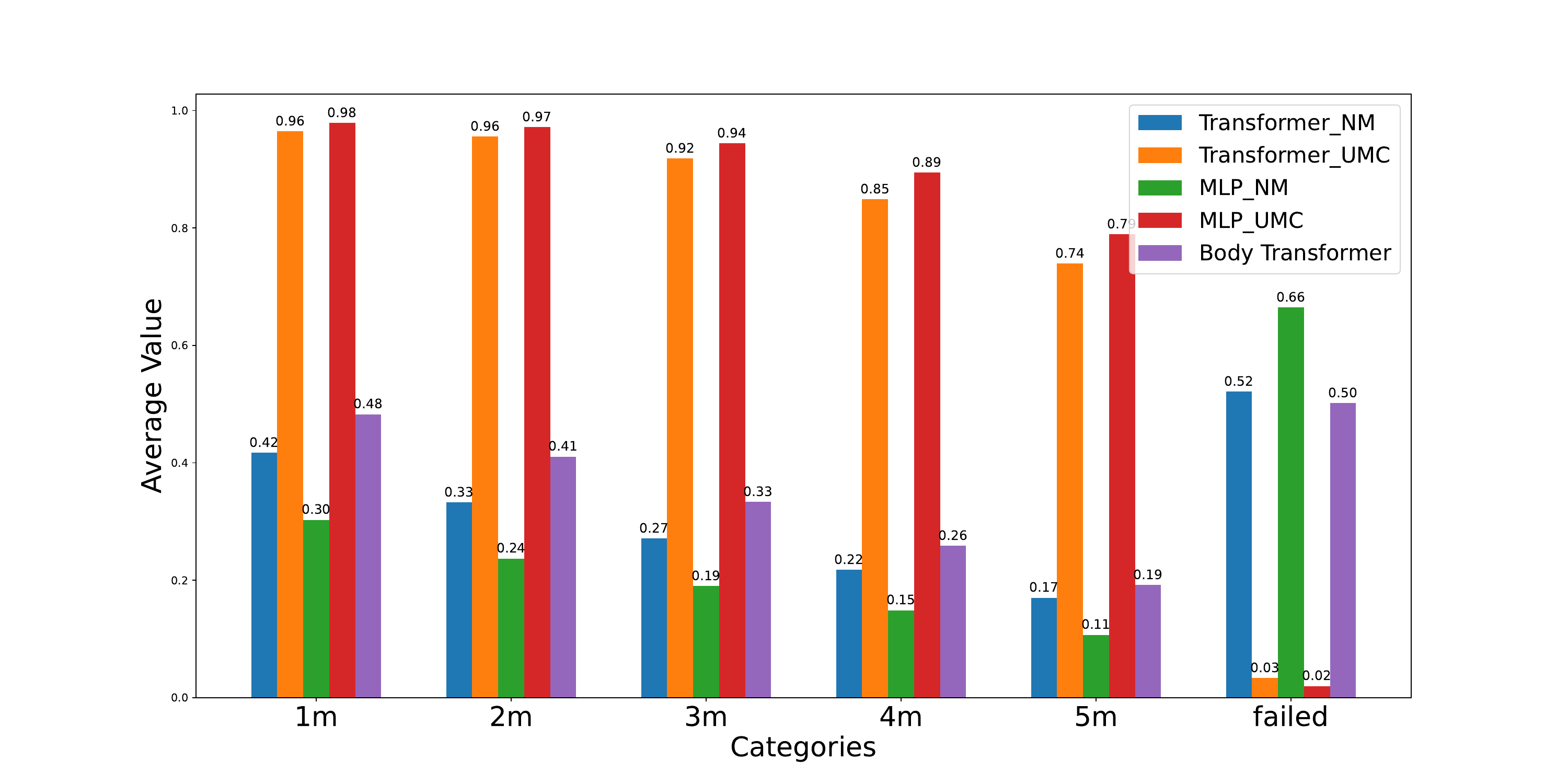}
    \caption{Detected Motor-Limit Condition}
    \label{fig:avg_mot_det}
  \end{subfigure}
  \hfill
  \begin{subfigure}{0.49\linewidth}
    \centering
    \includegraphics[width=\linewidth]{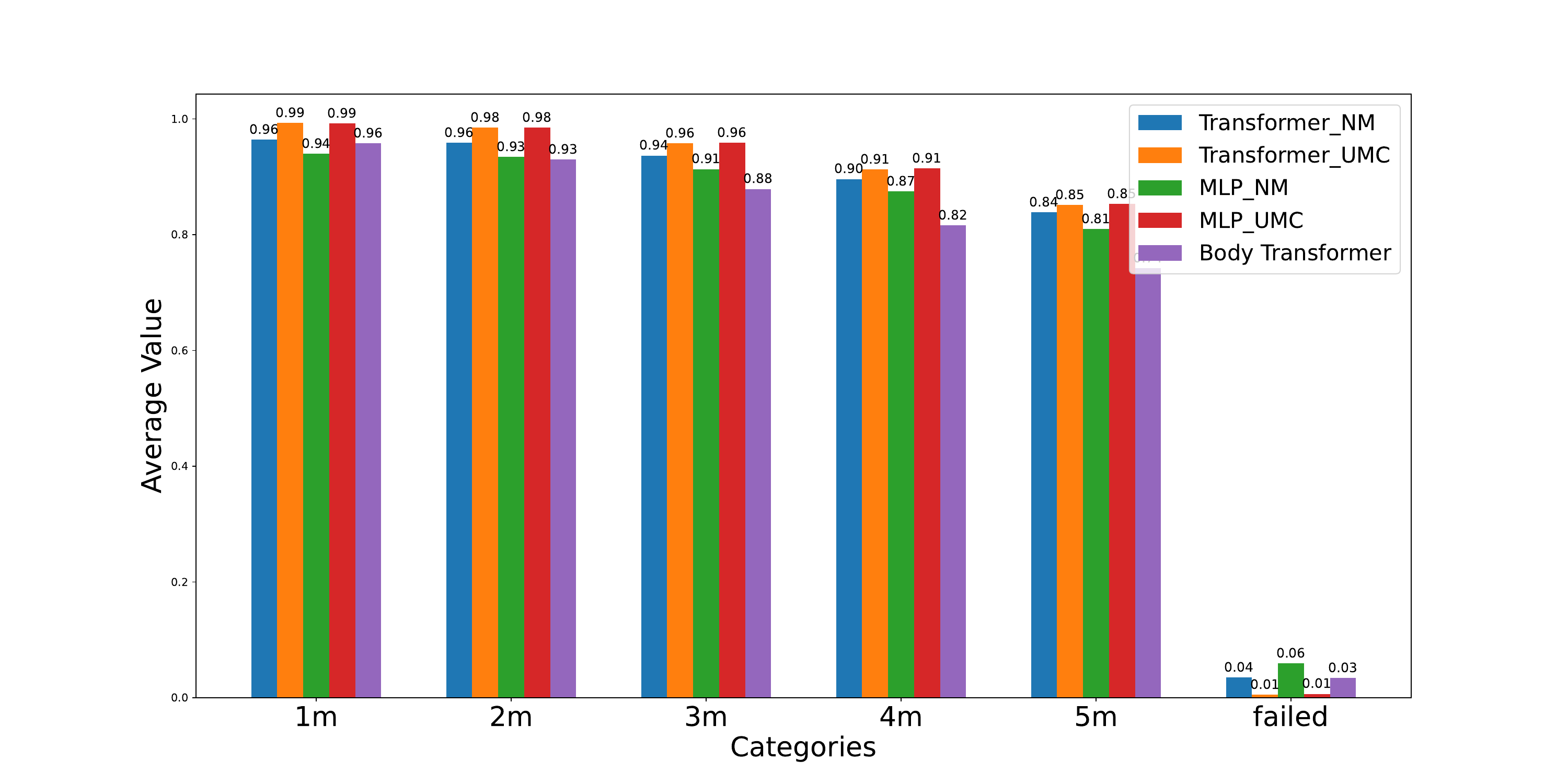}
    \caption{Undetected Motor-Limit Condition}
    \label{fig:avg_mot_und}
  \end{subfigure}
  \hfill
  \begin{subfigure}{0.49\linewidth}
    \centering
    \includegraphics[width=\linewidth]{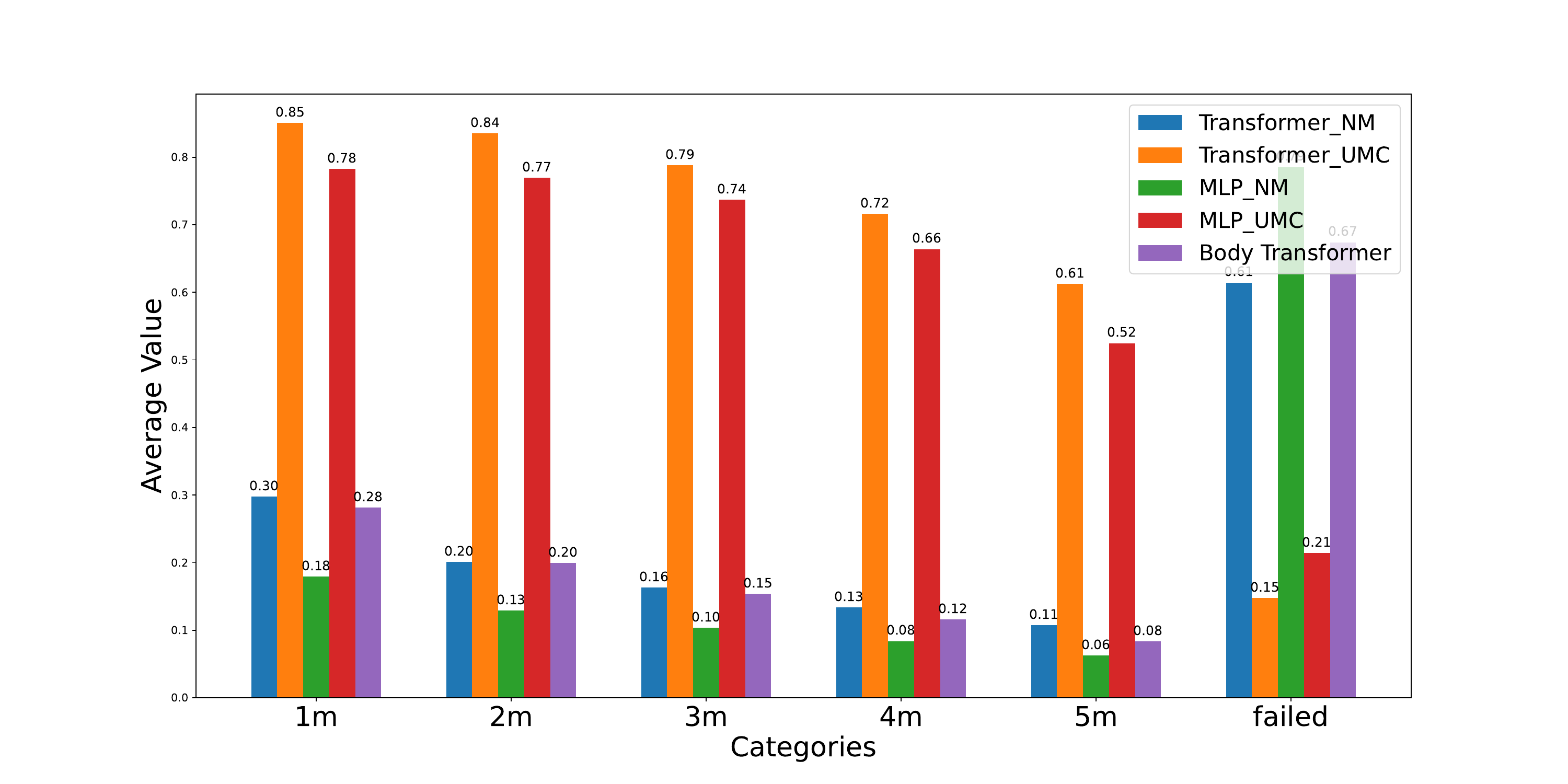}
    \caption{Detected Velocity-Limit Condition}
    \label{fig:avg_vel_det}
  \end{subfigure}
  \hfill
  \begin{subfigure}{0.49\linewidth}
    \centering
    \includegraphics[width=\linewidth]{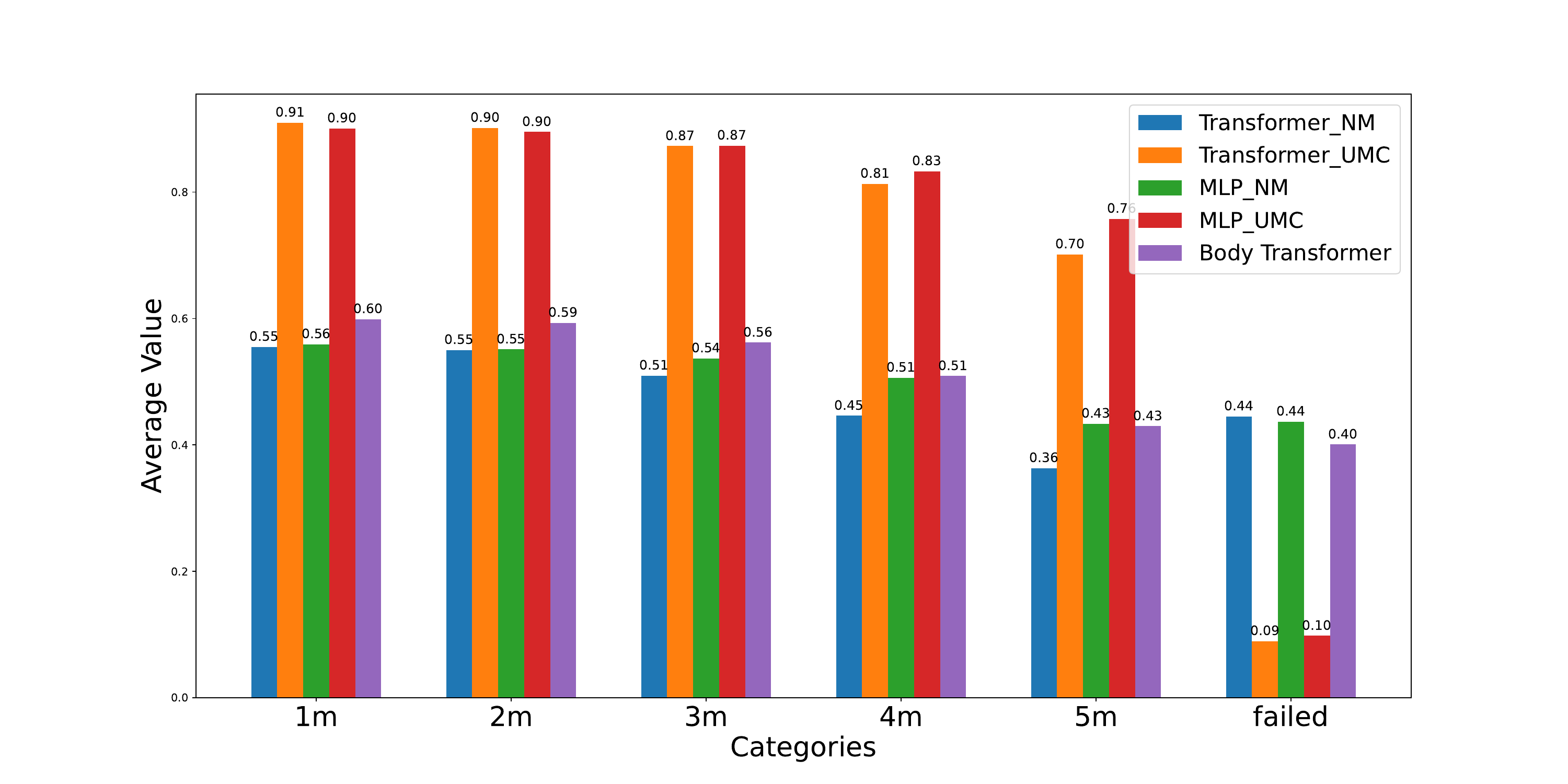}
    \caption{Undetected Velocity-Limit Condition}
    \label{fig:avg_vel_und}
  \end{subfigure}
  \hfill
  \caption{Average Performance of Five Methods Under Different Damage Conditions Across Three Tasks.}
  \label{fig:avg_detail_results}
\end{figure*}

\begin{figure*}[ht]
  \centering
  \begin{subfigure}{0.49\linewidth}
    \centering
    \includegraphics[width=\linewidth]{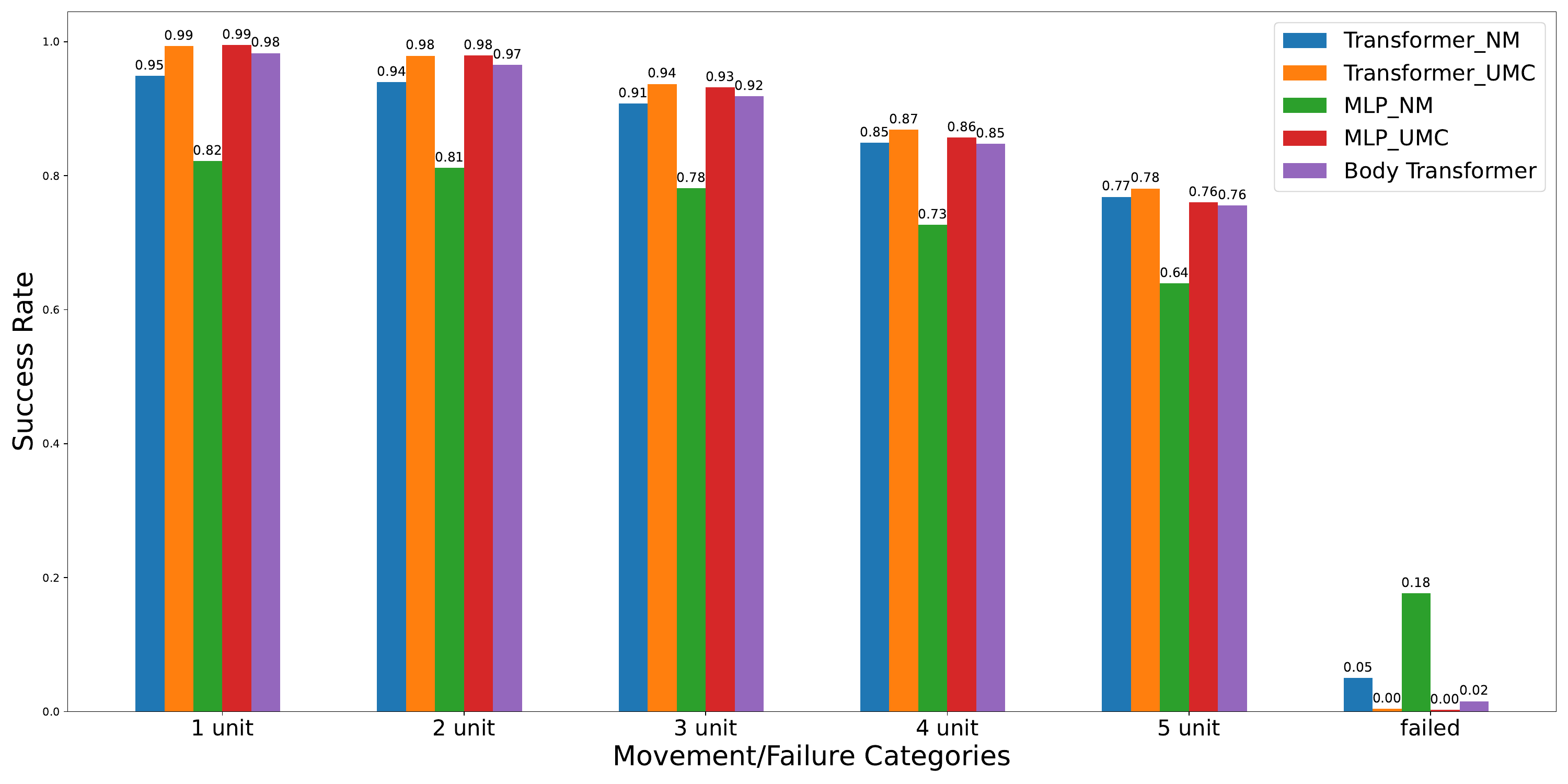}
    \caption{Normal Condition}
    \label{fig:solo8_norm}
  \end{subfigure}
  \hfill
  \begin{subfigure}{0.49\linewidth}
    \centering
    \includegraphics[width=\linewidth]{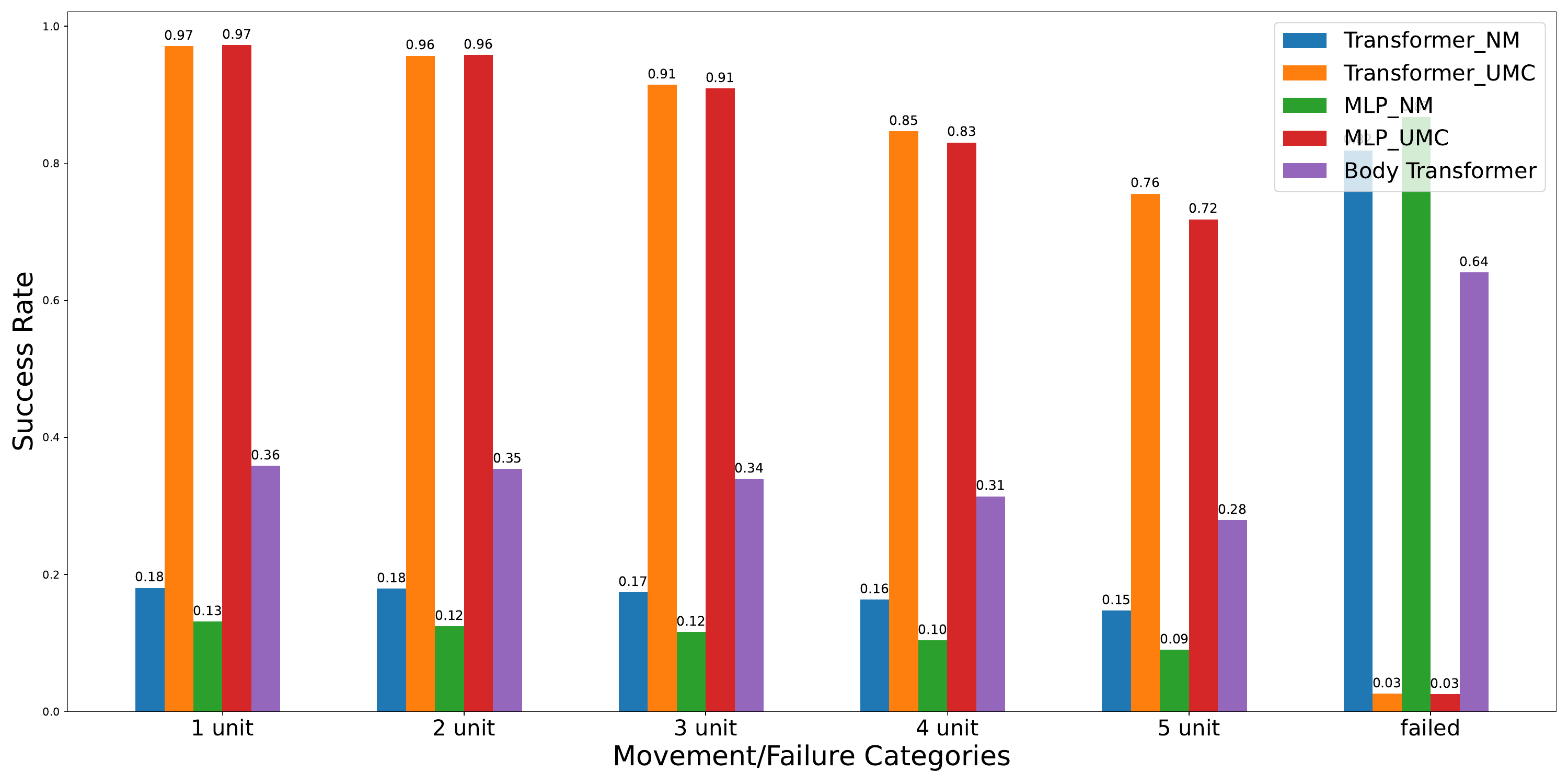}
    \caption{Sensor-Damaged Condition}
    \label{fig:solo8_obslimit}
  \end{subfigure}
  \hfill
  \begin{subfigure}{0.49\linewidth}
    \centering
    \includegraphics[width=\linewidth]{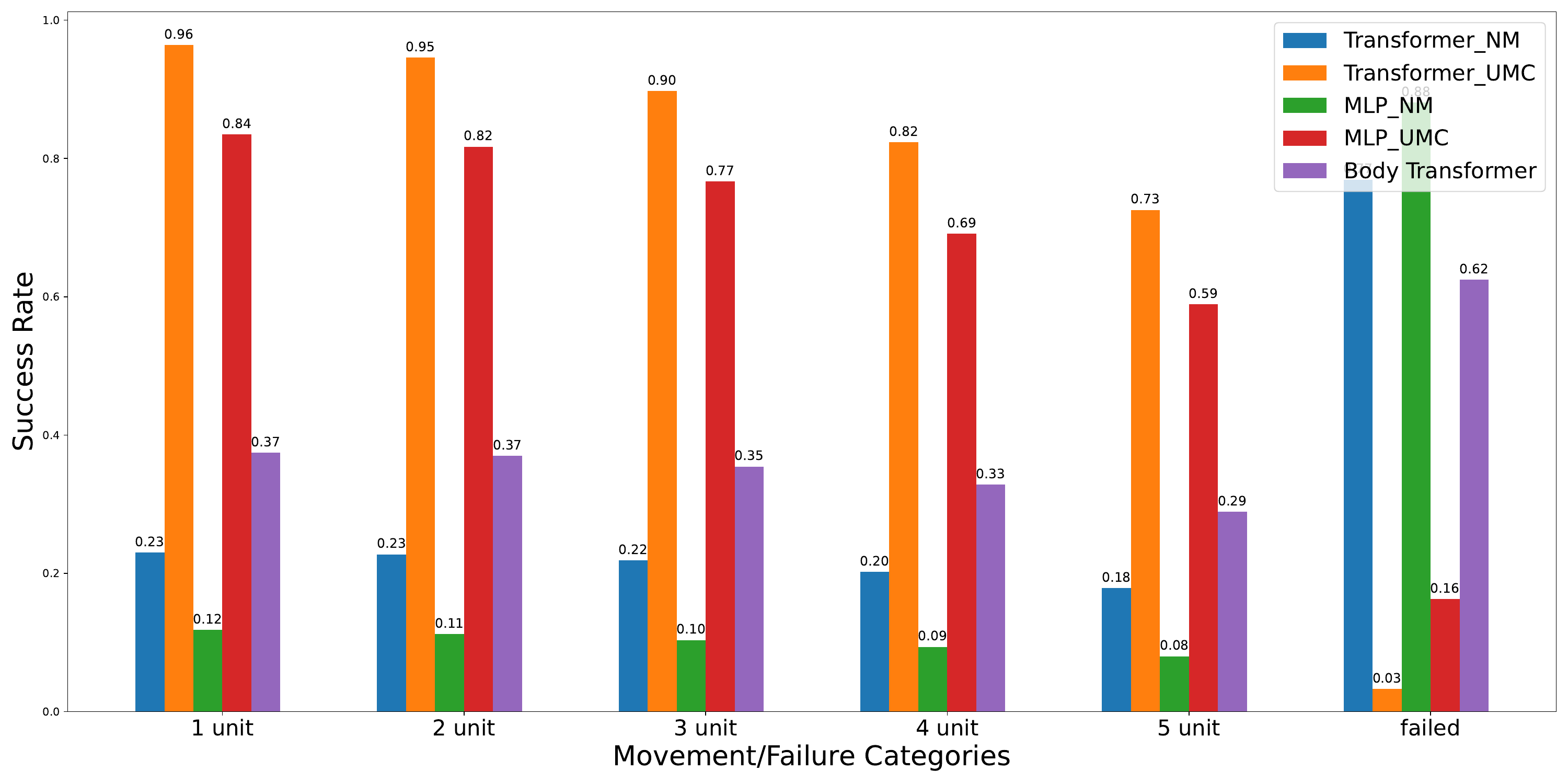}
    \caption{Detected ROM-Limit Condition}
    \label{fig:solo8_rom_det}
  \end{subfigure}
  \hfill
  \begin{subfigure}{0.49\linewidth}
    \centering
    \includegraphics[width=\linewidth]{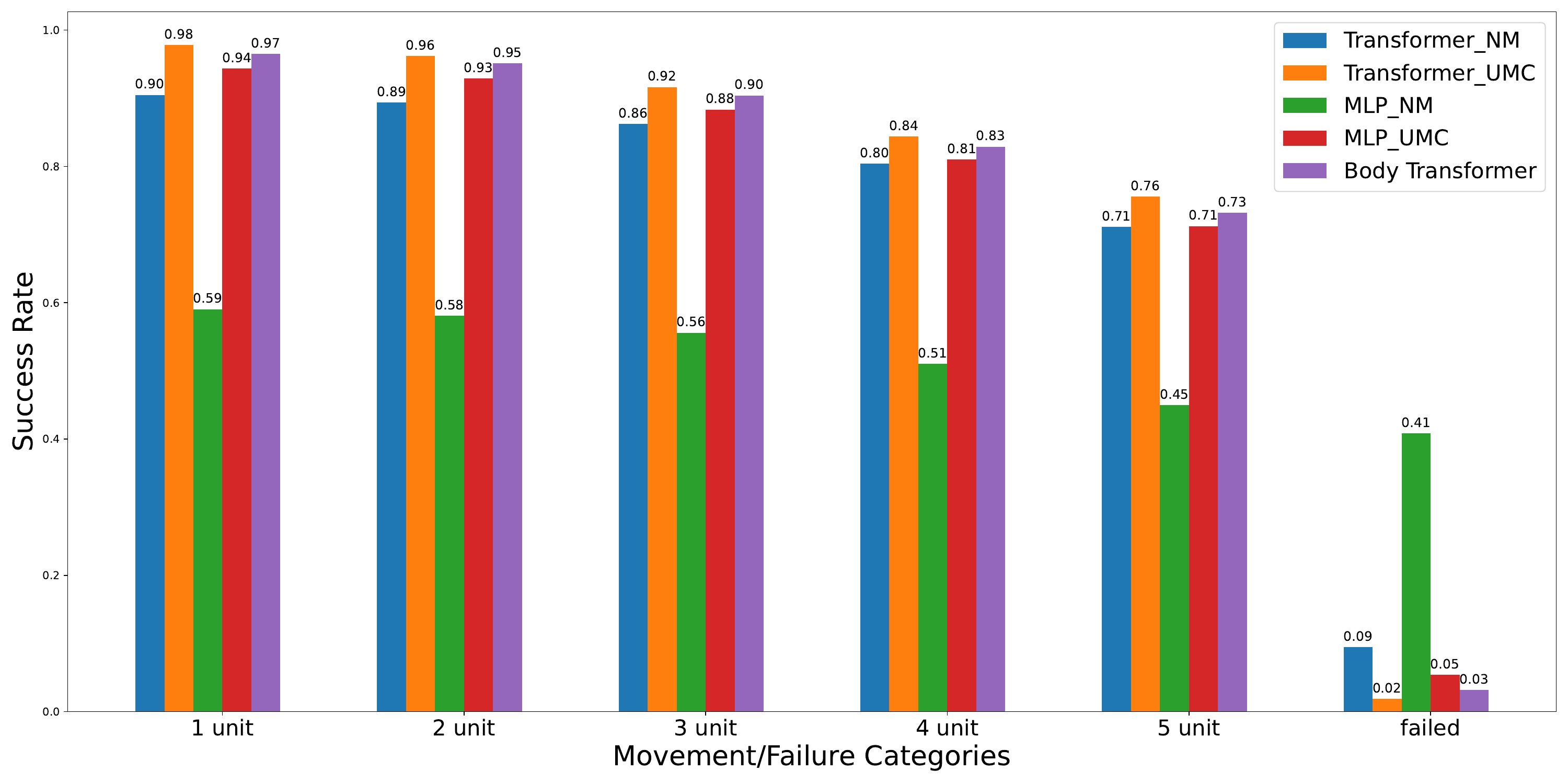}
    \caption{Undetected ROM-Limit Condition}
    \label{fig:solo8_rom_und}
  \end{subfigure}
  \hfill
  \begin{subfigure}{0.49\linewidth}
    \centering
    \includegraphics[width=\linewidth]{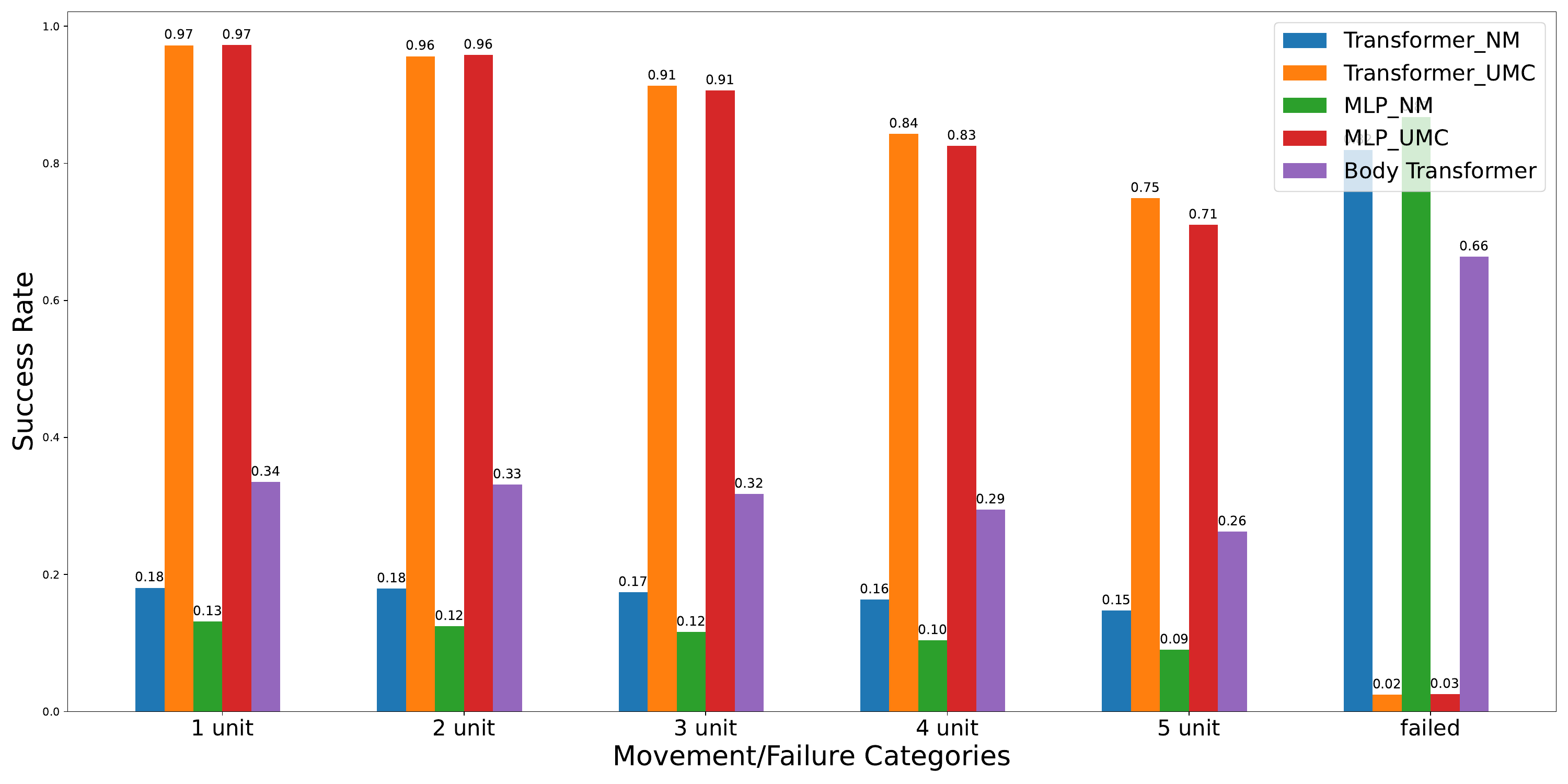}
    \caption{Detected Motor-Limit Condition}
    \label{fig:solo8_mot_det}
  \end{subfigure}
  \hfill
  \begin{subfigure}{0.49\linewidth}
    \centering
    \includegraphics[width=\linewidth]{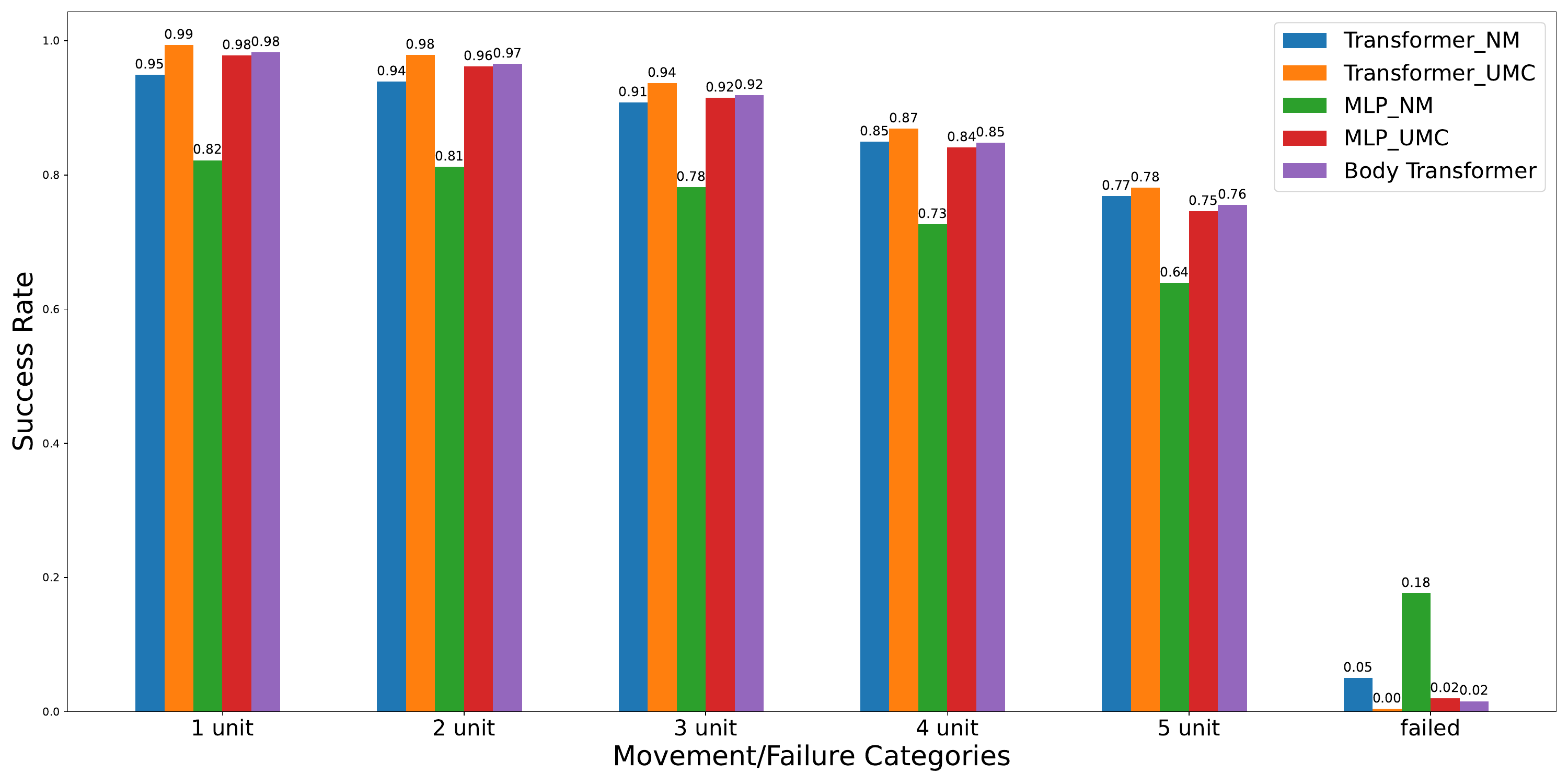}
    \caption{Undetected Motor-Limit Condition}
    \label{fig:solo8_mot_und}
  \end{subfigure}
  \hfill
  \begin{subfigure}{0.49\linewidth}
    \centering
    \includegraphics[width=\linewidth]{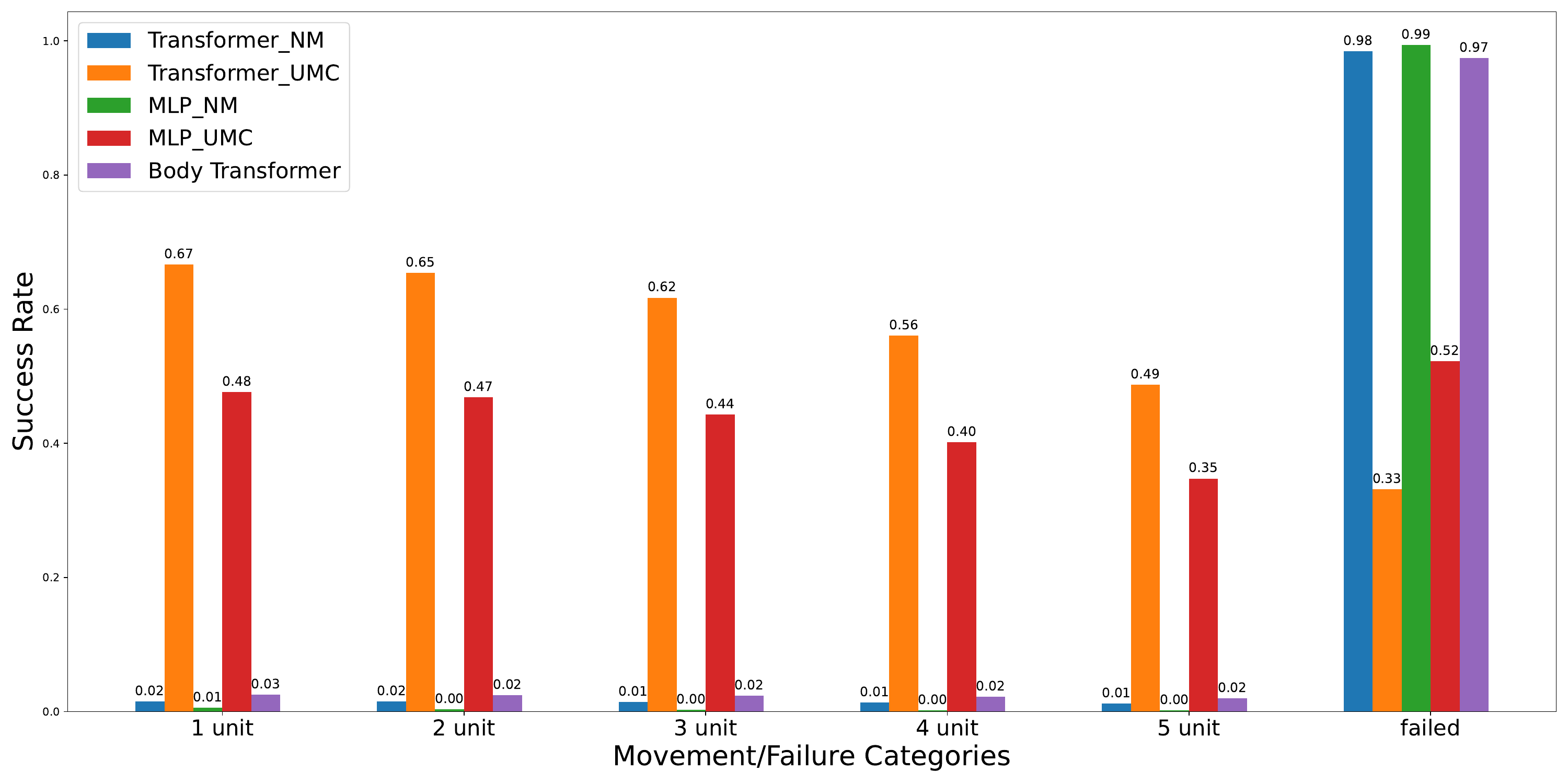}
    \caption{Detected Velocity-Limit Condition}
    \label{fig:solo8_vel_det}
  \end{subfigure}
  \hfill
  \begin{subfigure}{0.49\linewidth}
    \centering
    \includegraphics[width=\linewidth]{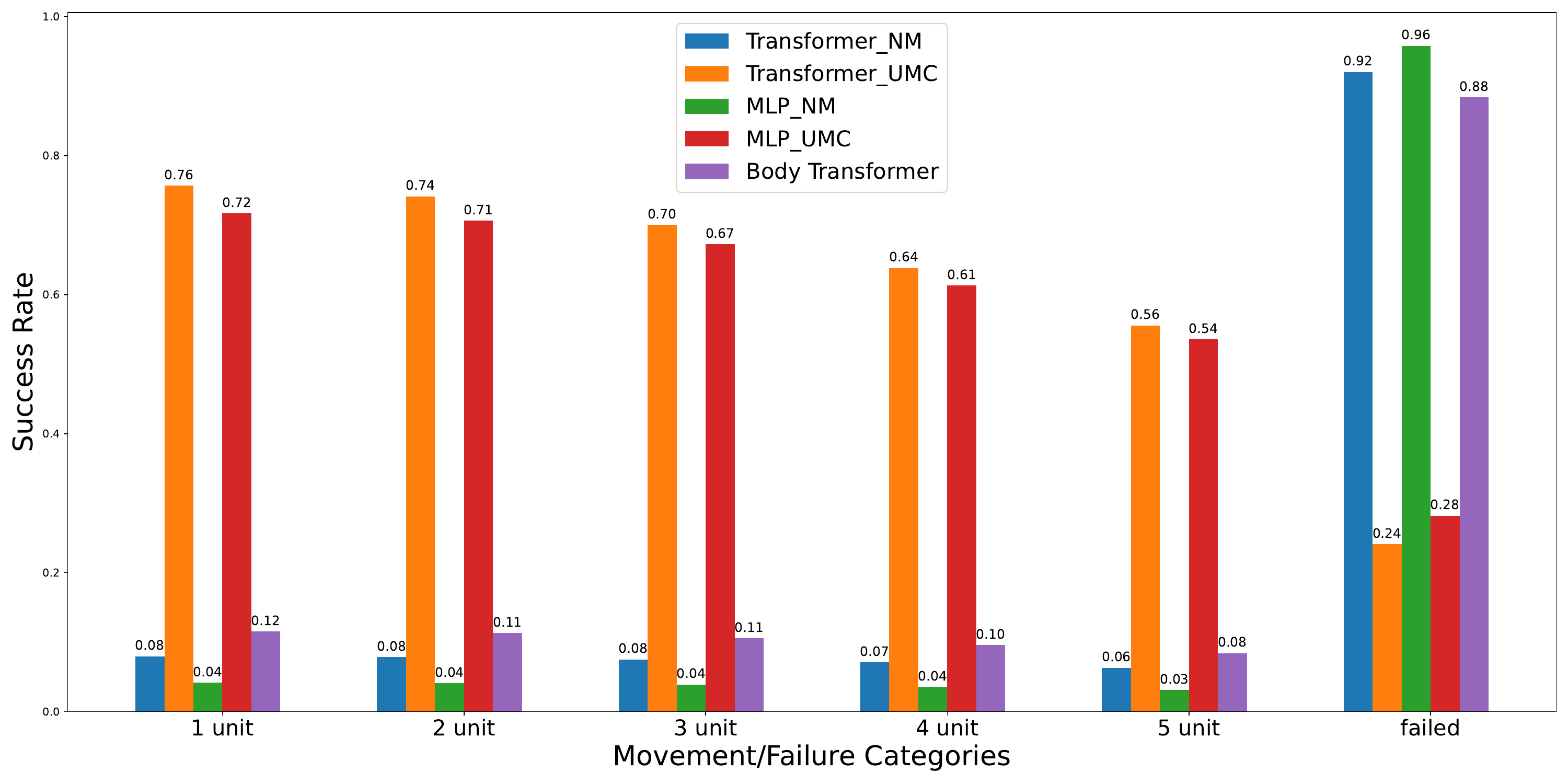}
    \caption{Undetected Velocity-Limit Condition}
    \label{fig:solo8_vel_und}
  \end{subfigure}
  \hfill
  \caption{Performance Between UMC and `MT-FTC' Under Different Damage Conditions in the Solo8 Task.}
  \label{fig:solo8_detail_results}
\end{figure*}

In this section, we present all experimental results to highlight the overall superiority of our UMC framework across various damage scenarios under different tasks.

In the body of the paper, we have presented the overall performance of all methods across all tasks and damage conditions, so we will not repeat such details here. Instead, we display their performance under each task's eight damage scenarios. Specifically, \cref{fig:a1_detail_results} is for the A1-Walk task, \cref{fig:h1_detail_results} is for the Unitree-H1 task, \cref{fig:g1_detail_results} is for the Unitree-G1 task, and \cref{fig:solo8_detail_results} is for the Solo8 task (SOTA comparison). Additionally, we calculated the average performance of the three baselines and our two UMC methods across three tasks for each damage condition and show the results in \cref{fig:avg_detail_results}. These statistics demonstrate the superior performance of our UMC framework.

Also, as addressed in the main text, \cref{fig:a1_norm}, \cref{fig:h1_norm} and \cref{fig:g1_norm} illustrate that our UMC framework does not compromise the robot's mobility under normal conditions. On the contrary, as shown in \cref{fig:g1_norm}, UMC even reduces the failure rate of the MLP model by 18\% while also enhancing its task-completion performance in the Unitree-G1 task under normal scenarios.

In sum, by synthesizing the insights from these figures, we provide a comprehensive comparison of the strengths and weaknesses of each method, highlighting the advantages of the UMC framework.

\section{More Ablation Results}

\begin{table}[ht]
    \caption{Inference Parameters for Ablations in \cref{sec:abl_maskvalue} (except for the stage-count ablation).}
  \centering
    \begin{tabular}{@{}l|c@{}}
      \toprule
      Parameter & Values \\
      \midrule
      Task & A1 \\
      Num Envs & 4000  \\
      Random Damage Range & [4,5]  \\
      ROM Limit & Random 10\% \\
      Motor Limit & 8 \\
      Velocity Limit & 3 \\
      Track Width & 6.0 \\
      Track Block Length & 6.0 \\
      Border Size & 4 \\
      Perlin Noise Seed & 100 \\
      Random Malfunction Seed & 1 \\
      Malfunction Timing & 75 \\
 
      \bottomrule
    \end{tabular}

  \label{tab:abl_setting}
\end{table}

\begin{figure*}[ht]
  \centering
  \begin{subfigure}{0.49\linewidth}
    \centering
    \includegraphics[width=\linewidth]{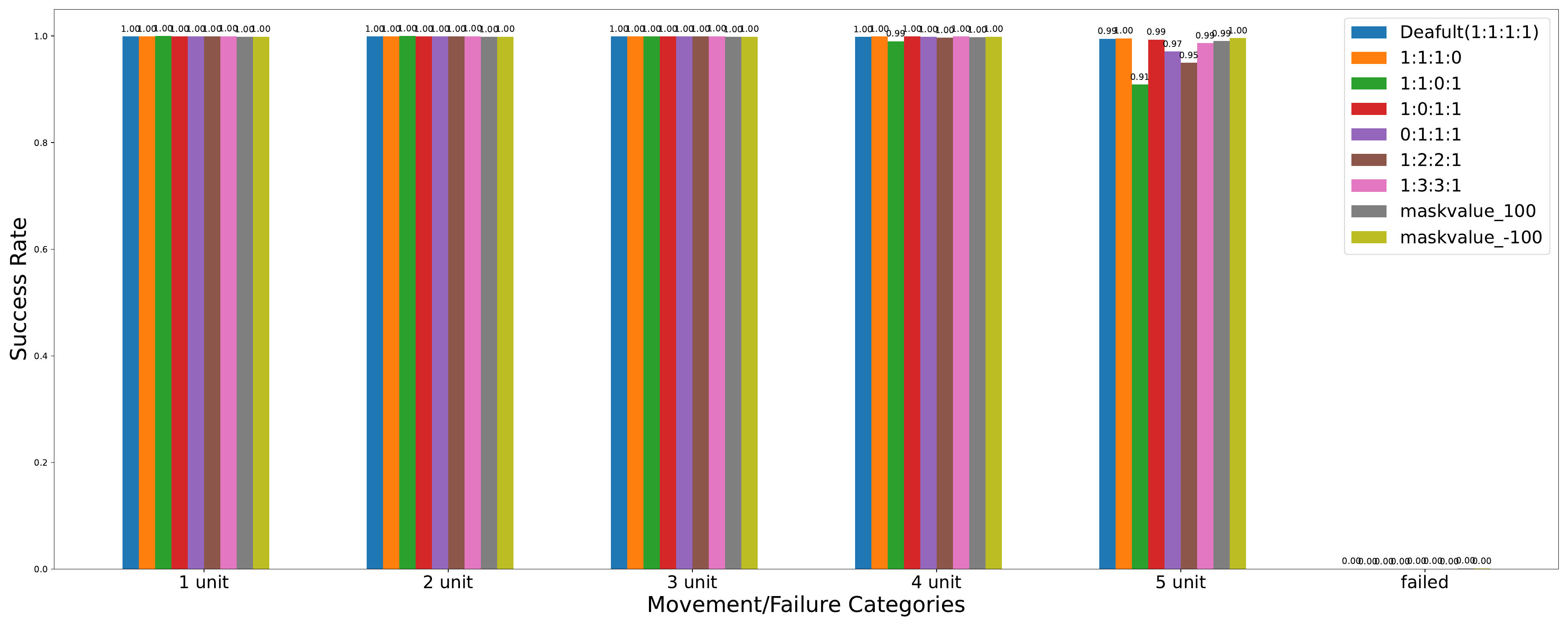}
    \caption{Normal Condition}
  \end{subfigure}
  \hfill
  \begin{subfigure}{0.49\linewidth}
    \centering
    \includegraphics[width=\linewidth]{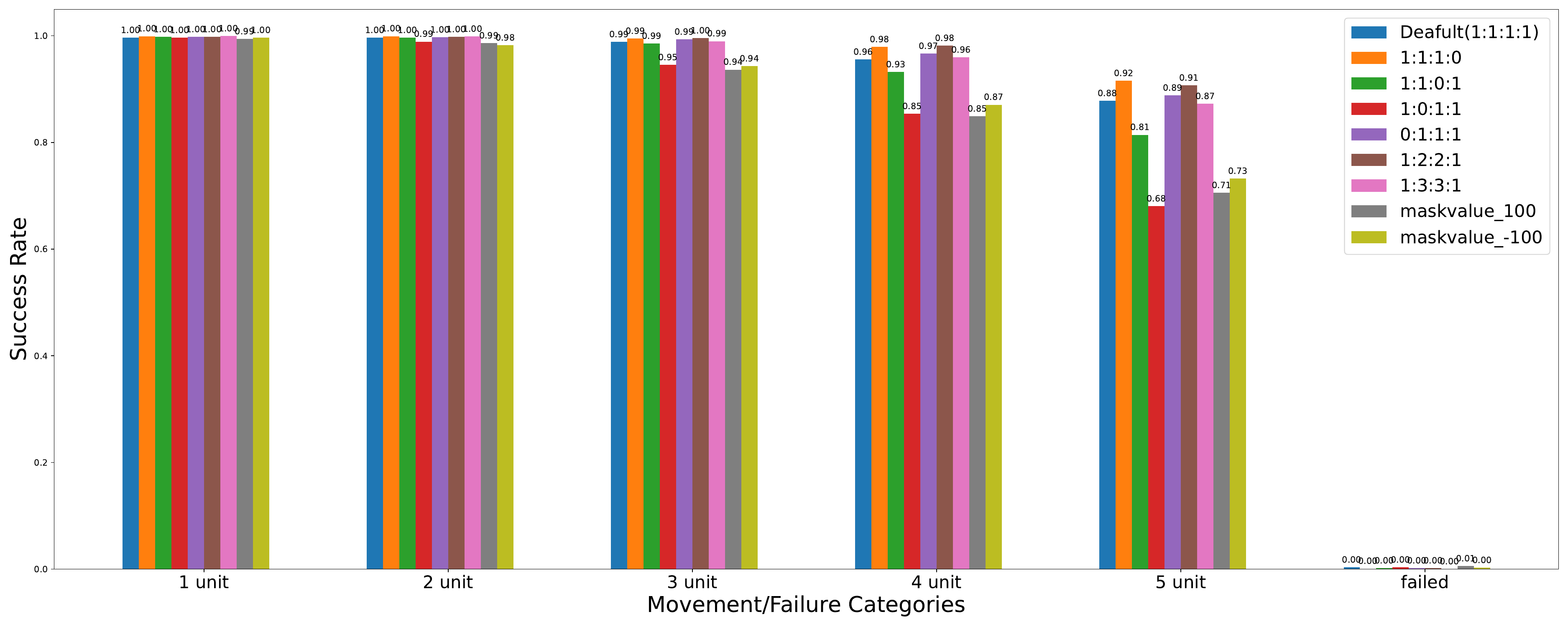}
    \caption{Sensor-Damaged Condition}
  \end{subfigure}
  \hfill
  \begin{subfigure}{0.49\linewidth}
    \centering
    \includegraphics[width=\linewidth]{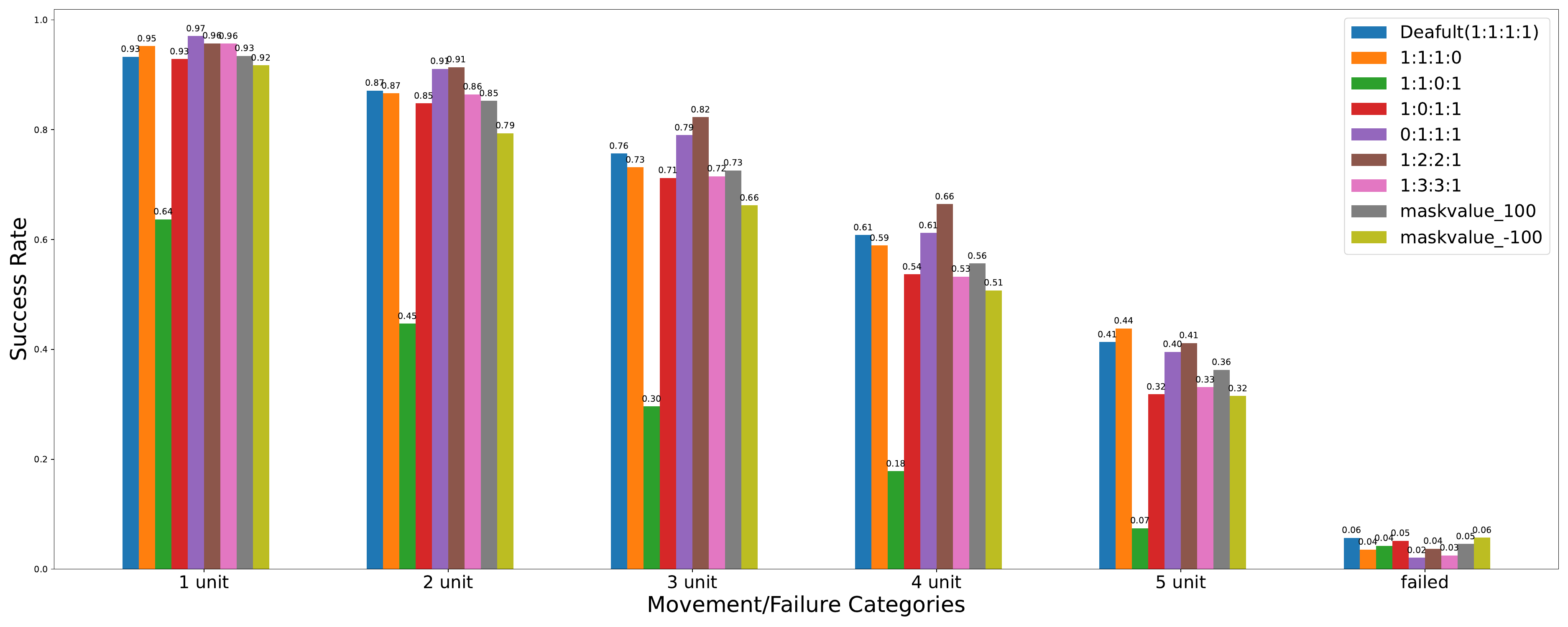}
    \caption{Detected ROM-Limit Condition}
  \end{subfigure}
  \hfill
  \begin{subfigure}{0.49\linewidth}
    \centering
    \includegraphics[width=\linewidth]{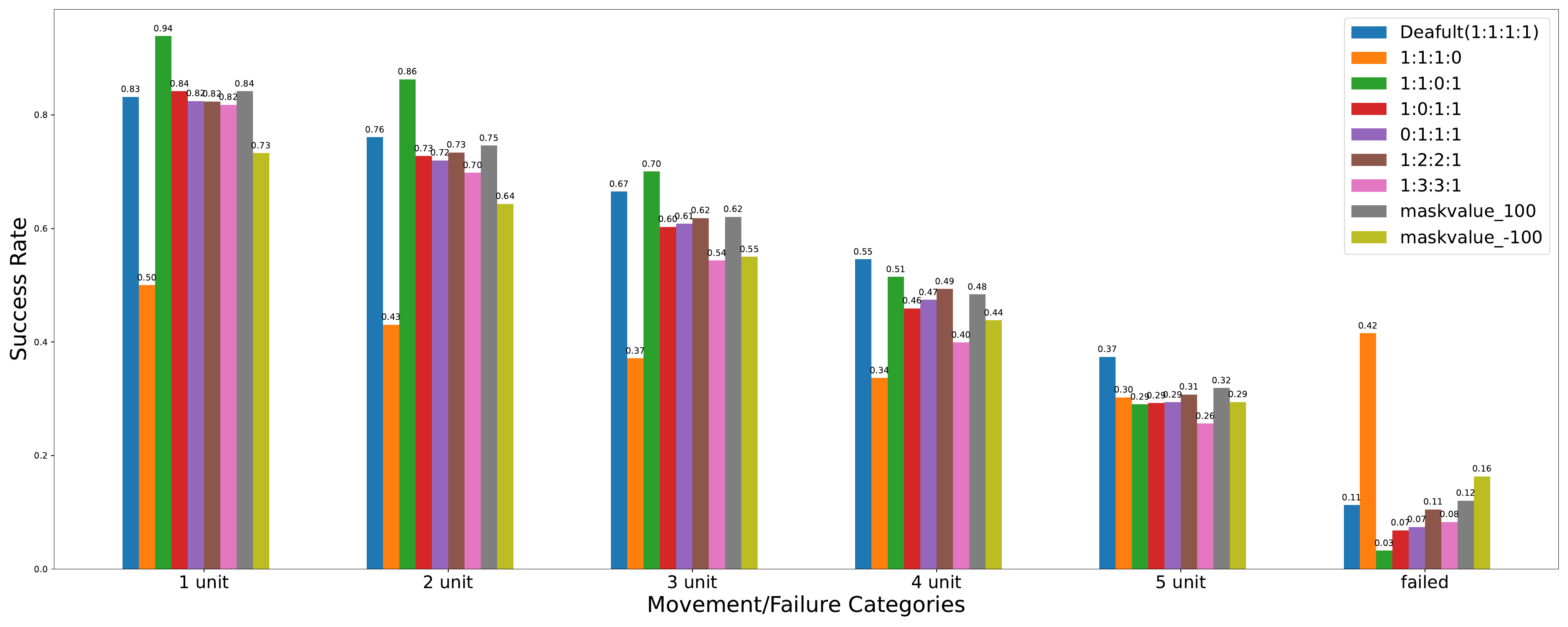}
    \caption{Undetected ROM-Limit Condition}
  \end{subfigure}
  \hfill
  \begin{subfigure}{0.49\linewidth}
    \centering
    \includegraphics[width=\linewidth]{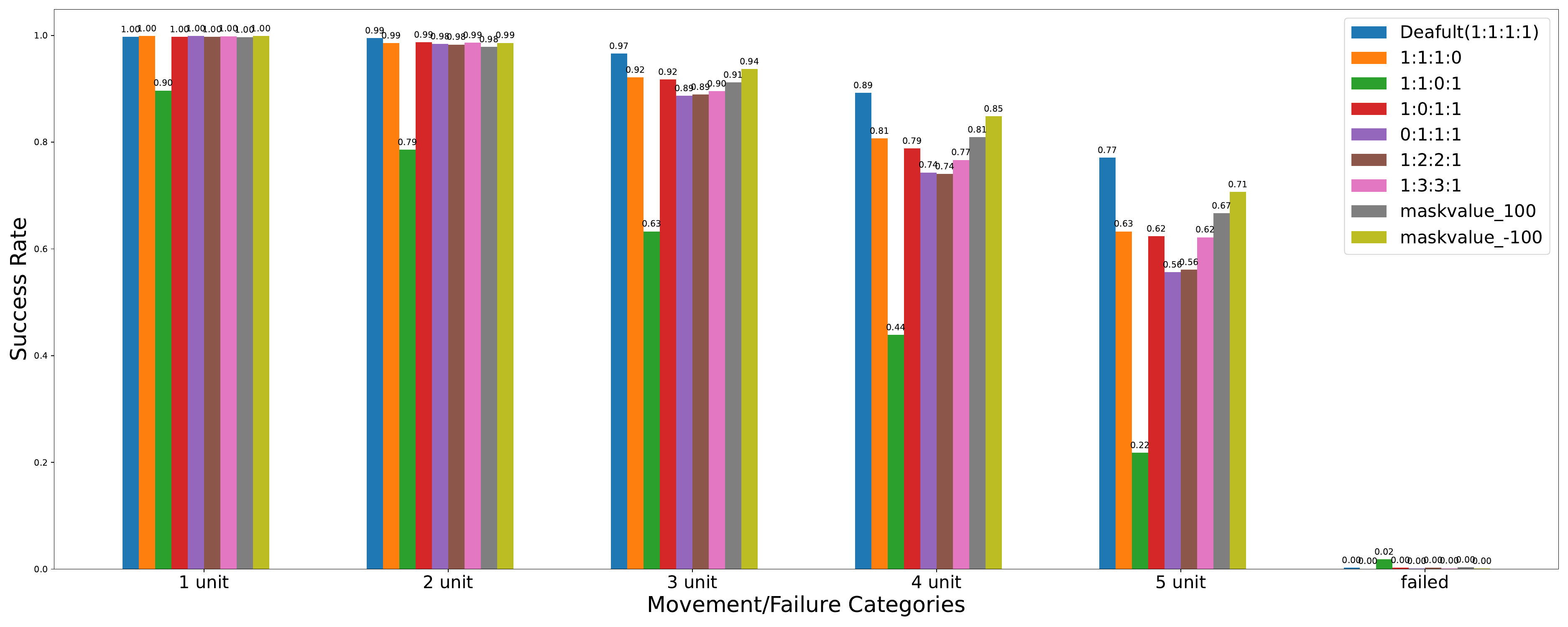}
    \caption{Detected Motor-Limit Condition}
  \end{subfigure}
  \hfill
  \begin{subfigure}{0.49\linewidth}
    \centering
    \includegraphics[width=\linewidth]{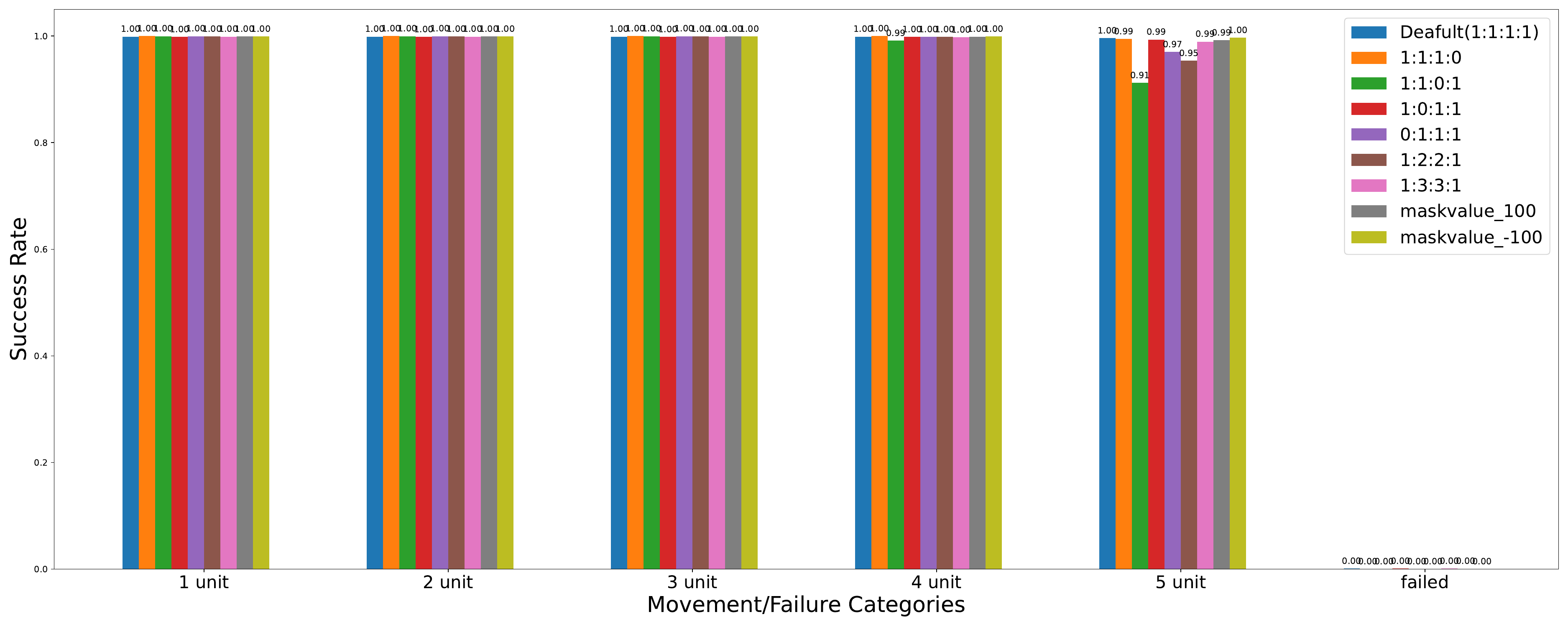}
    \caption{Undetected Motor-Limit Condition}
  \end{subfigure}
  \hfill
  \begin{subfigure}{0.49\linewidth}
    \centering
    \includegraphics[width=\linewidth]{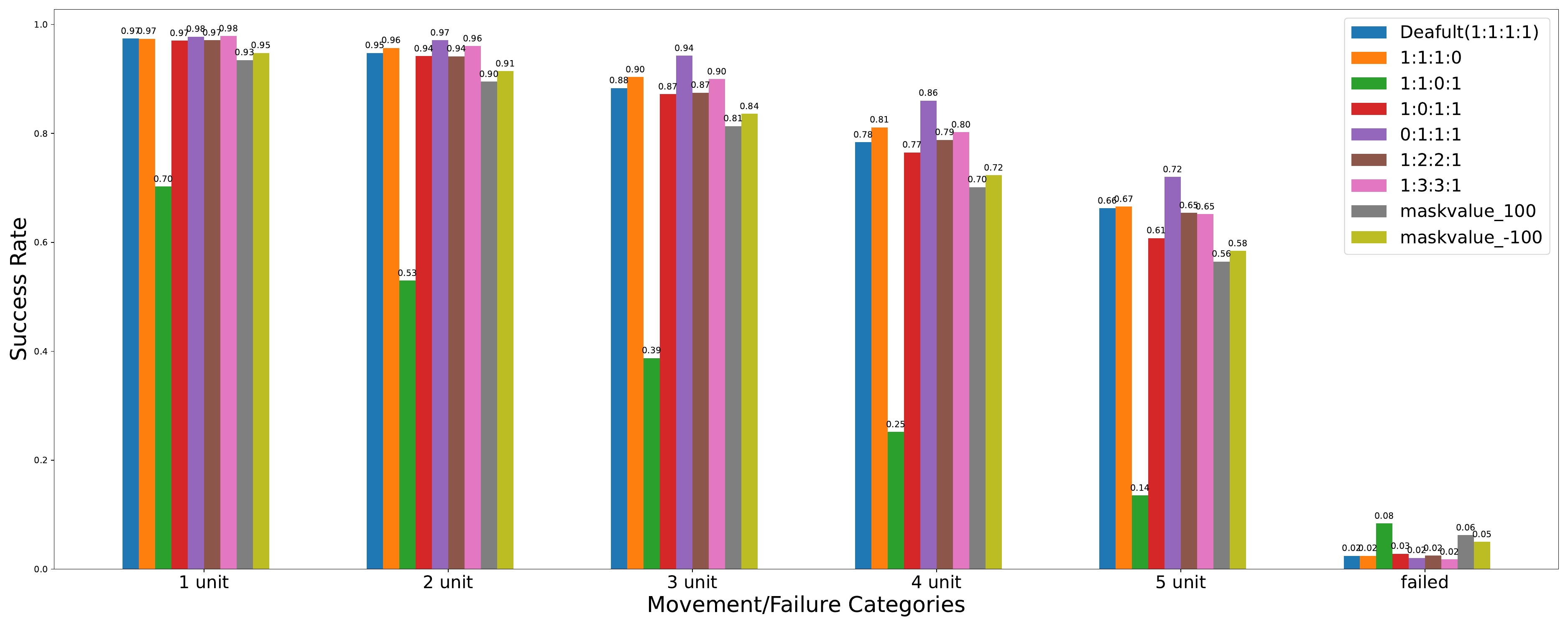}
    \caption{Detected Velocity-Limit Condition}
  \end{subfigure}
  \hfill
  \begin{subfigure}{0.49\linewidth}
    \centering
    \includegraphics[width=\linewidth]{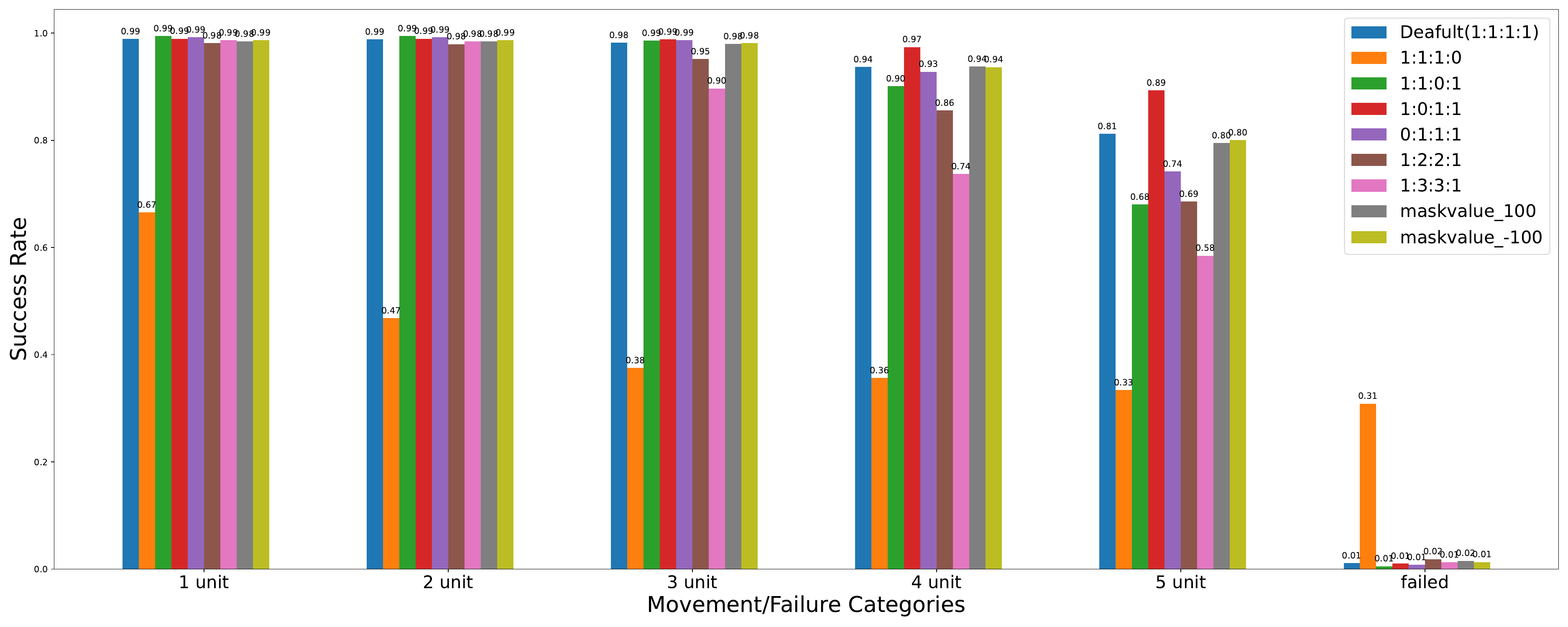}
    \caption{Undetected Velocity-Limit Condition}
  \end{subfigure}
  \hfill
  \caption{Full Comparison for the Ratio and Masking-Value Ablations Under Different Damage Conditions.}
  \label{fig:full_ablation_comparison}
\end{figure*}

All ablations, except for the training-stage one, are conducted under the A1 task with one inference damage setting under transformer-based UMC. \cref{tab:abl_setting} shows the parameter setting during those ablations. \cref{fig:full_ablation_comparison} demonstrates the performance of the ratio and masking-value ablation under eight specific damaged scenarios.

\section{Loss}
The total loss function in PPO adopted in our work is defined as:
\begin{equation}
\mathcal{L} = \mathcal{L}_{\text{surrogate}} + \lambda_{1} \cdot \mathcal{L}_{\text{value}} + \lambda_{2} \cdot \mathcal{L}_{\text{entropy}},
\end{equation}
where $\lambda_{1}$ and $\lambda_{2}$ denote weight parameters. \(\mathcal{L}_{\text{surrogate}}\) is illustrated in \cref{eq:surrogate}, \(\mathcal{L}_{\text{value}}\) is illustrated in \cref{eq:value}, and \(\mathcal{L}_{\text{entropy}}\) is an entropy regularization to encourage exploration. Gradient clipping is also applied to ensure stability during training.

The actor model in PPO is trained using a clipped surrogate loss to ensure that the updated policy does not deviate excessively from the previous policy. This clipping mechanism helps balance between exploring new actions and maintaining stability in learning. The loss function is defined as:
\begin{equation}
\mathcal{L}_{\text{surro}} = -\mathbb{E}_t \left[\min\left(r_t(\theta) \cdot A_t, \text{clip}(r_t(\theta), 1-\epsilon, 1+\epsilon) \cdot A_t \right)\right],
\label{eq:surrogate}
\end{equation}
where \(t\) represents the timestep index within a trajectory. The ratio \( r_t(\theta) = \frac{\pi_\theta(a_t | s_t)}{\pi_{\theta_{\text{old}}}(a_t | s_t)} \) is the probability ratio between the new policy \(\pi_\theta(a_t | s_t)\) and the old policy \(\pi_{\theta_{\text{old}}}(a_t | s_t)\), reflecting how much the new policy has changed. The term \(A_t\) represents the advantage estimate at timestep \(t\), which indicates how much better or worse the action \(a_t\) is compared to the average action under the current policy. The advantage \(A_t\) is computed as:
\begin{equation}
A_t = \sum_{k=0}^{\infty} (\gamma \lambda)^k \delta_{t+k},
\label{eq:advantage}
\end{equation}
where \(\delta_{t+k}\) is the temporal difference residual at timestep \( t+k \), representing the error in predicting the future value of the state based on the current value function. \( \gamma \) is the discount factor, determining the importance of future rewards, and \( \lambda \) balances the bias-variance tradeoff in advantage estimation.

The critic model in UMC shares the same architecture as the actor model, except without the damage detection module, and it is trained using the value function loss to minimize the error between the predicted value \(V(s_t)\) and the target return \(R_t\):
\begin{equation}
\begin{split}
\mathcal{L}_{\text{value}} = \mathbb{E}_t \big[ & \max\big((V(s_t) - R_t)^2, \\
& \text{clip}(V(s_t), V_{\text{old}}(s_t) - \epsilon, V_{\text{old}}(s_t) + \epsilon) - R_t)^2\big)\big],
\end{split}
\label{eq:value}
\end{equation}
where \(V_{\text{old}}(s_t)\) is the previous value function estimate, and \(\epsilon\) is the clipping threshold to ensure stability during training.


\end{document}